\PassOptionsToPackage{dvipsnames}{xcolor} 
\documentclass{article}

\usepackage{microtype}
\usepackage{graphicx}
\usepackage{adjustbox} 
\usepackage{booktabs} 
\usepackage{subcaption} 
\usepackage{enumitem}
\usepackage{caption} 

\usepackage[colorlinks]{hyperref}

\usepackage[accepted]{icml2025} 

\usepackage{xcolor}

\usepackage{amsmath}
\usepackage{amssymb}
\usepackage{mathtools}
\usepackage{amsthm}
\usepackage{natbib}

\usepackage{algorithm}
\usepackage{algorithmic}

\usepackage[capitalize,noabbrev]{cleveref}

\usepackage{array}
\usepackage{colortbl}  
\usepackage{multirow}

\usepackage{float}
\usepackage{placeins}
\usepackage{multicol}

\usepackage[textsize=tiny]{todonotes}

\theoremstyle{plain}

\theoremstyle{definition}

\theoremstyle{remark}

\icmltitlerunning{Diffusion Instruction Tuning}

\begin{document}

\twocolumn[
\icmltitle{Diffusion Instruction Tuning}

\begin{icmlauthorlist}
\icmlauthor{Chen Jin}{az}
\icmlauthor{Ryutaro Tanno}{deepmind}
\icmlauthor{Amrutha Saseendran}{az}
\icmlauthor{Tom Diethe}{az}
\icmlauthor{Philip Teare}{az}
\end{icmlauthorlist}

\icmlaffiliation{az}{Centre for AI, AstraZeneca, Cambridge, UK}
\icmlaffiliation{deepmind}{Google DeepMind, UK}

\icmlcorrespondingauthor{}{chen.jin@astrazeneca.com}

\begin{@twocolumnfalse}
\vspace{0.2cm}
  {
    \centering
    \includegraphics[width=1\linewidth]{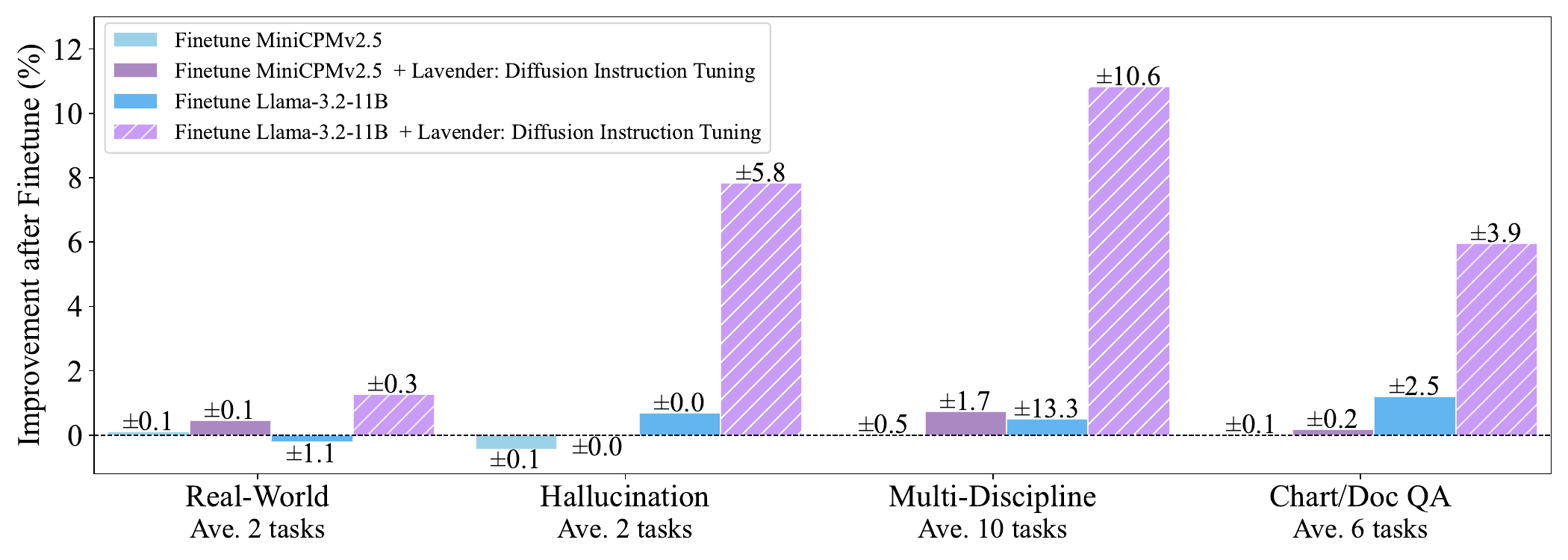}
    \vspace{-8mm}
    \captionof{figure}{Average Performance on 20 Vision-Language Reasoning Benchmarks (Grouped into 4 Categories).
    }
    \vspace{-2mm}
    \label{fig:opening}
  }
\end{@twocolumnfalse}

\icmlkeywords{Machine Learning, ICML}

\vskip 0.3in
]



\printAffiliationsAndNotice{}  

\begin{abstract}
We introduce \textit{Lavender}, a simple supervised fine-tuning (SFT) method that boosts the performance of advanced vision-language models (VLMs) by leveraging state-of-the-art image generation models such as Stable Diffusion. 
Specifically, Lavender aligns the text-vision attention in the VLM transformer with the equivalent used by Stable Diffusion during SFT, instead of adapting separate encoders. This alignment enriches the model’s visual understanding and significantly boosts performance across in- and out-of-distribution tasks.
Lavender requires just 0.13 million training examples---2.5\% of typical large-scale SFT datasets---and fine-tunes on standard hardware (8 GPUs) in a single day. It consistently improves state-of-the-art open-source multimodal LLMs (e.g., Llama-3.2-11B, MiniCPM-Llama3-v2.5), achieving up to 30\% gains and a 68\% boost on challenging out-of-distribution medical QA tasks. 
By efficiently transferring the visual expertise of image generators with minimal supervision, Lavender offers a scalable solution for more accurate vision-language systems. Code, training data, and models are available on the \href{https://astrazeneca.github.io/vlm/}{project page}.
\end{abstract}

\vspace{-4mm}
\begin{figure}[ht]
    \centering
    \includegraphics[width=1\linewidth]{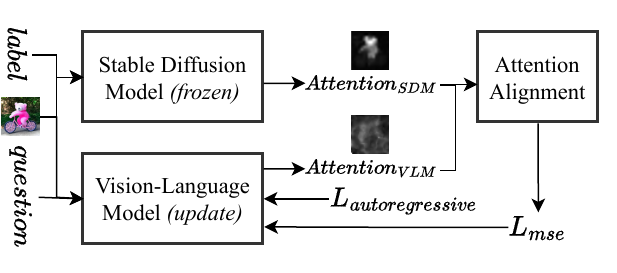}
    \vspace{-4mm}
    \caption{
    \textbf{\textit{Lavender}: Diffusion Instruction Tuning.}
    Lavender uses the text-vision attention maps of a Stable Diffusion Model, \(Attention_{SDM}\), as a guiding objective for the attention of the target vision-language model (VLM), \(Attention_{VLM}\). The \textit{Attention Alignment} module employs a 3-Layer ConvNet to transform \(Attention_{VLM}\) to match \(Attention_{SDM}\) via an MSE loss, acting as a regularisation term during supervised fine-tuning.
    }
    \label{fig:opening_lavender}
\end{figure}

\section{Introduction}
\label{introduction}

Training frontier foundation models from scratch costs millions of dollars at minimum, requiring hundreds of GPUs and millions to billions of data \cite{deepseekai2024deepseekv3technicalreport}. This challenge is even more pronounced in multimodal settings: vision-language models (VLMs) often face data scarcity because collecting paired image-text datasets is expensive \cite{zhu2024multimodal}. A common workaround is to apply supervised fine-tuning (SFT) on a pretrained large language model (LLM), leveraging its abundant text-only pretraining and adjusting bridging layers or additional encoders with limited image-text pairs \cite{liu2024visual, covert2024locality, jiang2023clip}. However, these methods typically overlook the importance of transformer-level attention alignment within the LLM core---a key component for effectively expanding text-based models into the visual domain.

\begin{figure}
    \centering
    \includegraphics[width=1\linewidth]{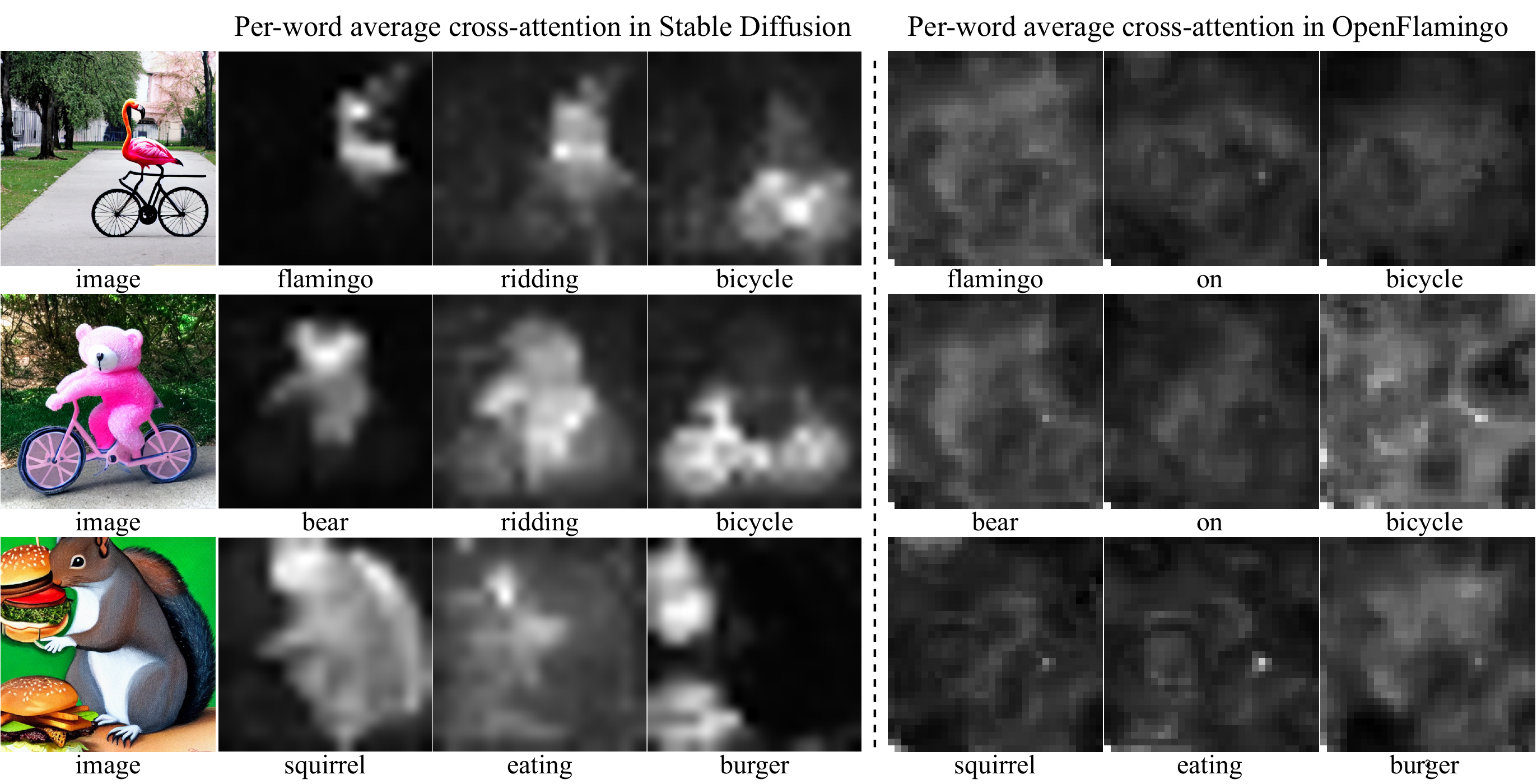}
    \vspace{-6mm}
    \captionsetup{skip=0pt} 
    \caption{\textbf{Image generation models (Stable Diffusion on the left) exhibit stronger word-to-region attention alignment than VLMs (Open-Flamingo on the right).} Per-word average attention maps suggest that diffusion models may be closer to an ideal distribution correlating image regions with textual tokens.}
    \label{fig:empirical}
\end{figure}

Precise visual-text alignment is crucial for advanced multimodal reasoning. While both VLMs and diffusion models (DMs) process text and images, they diverge in their generation objectives. We observe that DMs, such as Stable Diffusion \cite{rombach2021highresolution}, which reconstructs images at the pixel level, appear to learn more precise text-vision attention maps than VLMs that are optimised solely for text token generation (\cref{fig:empirical}).

In this work, we demonstrate that the high-quality cross-attention maps from these DMs indeed offer a useful target for guiding the text-vision attention in VLMs during SFT, thus improving word-to-region alignment and the overall performance. We introduce \textbf{Lavender} (\textbf{L}anguage-\textbf{a}nd-\textbf{V}ision fin\textbf{e}-tu\textbf{n}ing with \textbf{D}iffusion Align\textbf{er}), \emph{the first framework to directly align VLM transformer attention layers with those of Stable Diffusion} (\cref{fig:opening}(c)). Specifically, during SFT, Lavender transfers diffusion-based attention distributions to VLMs, enhancing core visual-textual interactions. To mitigate catastrophic forgetting, we additionally propose several attention aggregation methods and training strategies that preserve existing VLM competencies.

We begin by verifying Lavender on a small OpenFlamingo model: entropy and visual analyses show Lavender aligns VLM attention with DM attention. Leveraging Stable Diffusion to \emph{offline} extract per-word attention on 130k label-image pairs-\emph{no extra training cost}-Lavender yields notable gains over autoregressive finetuning on 20 diverse benchmarks, including up to 70\% improvement on OpenFlamingo across seven benchmarks. For Llama 3.2-11B, fine-tuned on in- and out-of-distribution data, performance improves by up to 30\% on 19 benchmarks, surpassing comparable small open-source models by 50\%. On self-attention-only MiniCPMv2.5, it achieves up to 4\% gains.

This advantage extends to severely OOD domains, evidenced by a 68\% boost on the WorldMedQA medical benchmark for Llama 3.2-11B. Further analyses reveal that larger fine-tuning sets help Lavender resist overfitting more effectively than autoregressive baselines, and the aligned attention maps yield finer-grained visual understanding. Together with qualitative evidence of improved VLM attention, these results confirm Lavender’s premise: \emph{diffusion-based attention distributions effectively align visual and textual representations}, fostering more robust, data-efficient VLMs.

Ablation studies reveal that the method of attention aggregation and the choice of layers for fine-tuning are critical to performance. Learned aggregation strategies outperform manual ones and lightweight pretraining of an Aligner Network helps prevent catastrophic forgetting on small datasets. LoRA fine-tuning delivers faster improvements, full fine-tuning proves more effective for handling complex tasks. Aligning all cross-attention layers proves most effective, highlighting the importance of precise attention alignment.

\vspace{-2mm}
\paragraph{In summary,} we introduce \textit{Lavender}, a novel framework that transfers ``visual expertise'' from text-to-image diffusion models to vision-language models \emph{without additional annotations}. By aligning attention distributions, Lavender enhances word-to-region grounding, improves fine-tuning efficiency, and boosts model robustness, particularly in out-of-distribution settings.
Moreover, our architecture-agnostic attention alignment loss is compatible with RL post-training, offering scalable diffusion-guided feedback instead of costly, subjective human vision feedback.
Beyond addressing data scarcity, Lavender demonstrates that pretrained generative models can guide multimodal learning in a scalable and compute-efficient manner. This approach bridges two expert paradigms—language and vision generation—into a more unified, capable system. Our findings suggest broader applications in multimodal AI, offering a modular and privacy-friendly alternative to closed-source models. We open-source our work to encourage further exploration of diffusion-guided alignment, unlocking new possibilities in vision-language reasoning.

\section{Related Work}
\label{sec:related_work}

\emph{We find that the gap in VLM alignment partly stems from the technological trajectories pursued over the past half-decade.}  
One key VLM milestone was Flamingo \cite{alayrac2022flamingo}, which laid the foundation for modern VLMs. In its design, images and text are processed by separate encoders, unified through a perceiver resampler \cite{jaegle2021perceiver}, and passed through deep transformer layers combining cross-attention and self-attention. Flamingo's elegant architecture established a new standard, influencing a range of subsequent models \cite{li2022blip, li2023blip, you2023ferret}.  
The importance of aligning vision-text correlations is evident in the design of Llama 3.2 \cite{dubey2024llama}, released two years later, which adopts Flamingo’s strategy of using a dedicated cross-attention module for effective interaction handling. However, training a VLM with a dedicated cross-attention module end-to-end requires substantial data and computational resources. These models are typically pre-trained on millions or even billions of image-text pairs and interleaved image-text datasets \cite{zhu2024multimodal}. 
Similar challenges apply to broader multimodal models beyond vision and language \cite{lu2024unified}.

Unlike VLMs, single-modality large language models (LLMs) have scaled more rapidly \cite{ouyang2022training, brown2020language, chowdhery2023palm}, often consuming over 100 million examples spanning 1,800 tasks \cite{longpre2023flan}. VLMs, however, face a training data gap due to the high cost of acquiring paired image-text datasets.  
To address this, researchers proposed leveraging scaled LLMs by instruction fine-tuning them on as little as 150k paired visual question answering (VQA) data using an autoregressive loss. This approach, pioneered by \citet{zhu2023minigpt, dai2023instructblipgeneralpurposevisionlanguagemodels, liu2024visual}, aligns text and image tokens through fine-tuning connectors such as MLPs, encoders, or decoders connecting to the LLM, providing an efficient pathway to integrate vision with language models for diverse tasks \cite{wang2024qwen2, li2024llava, koh2024generating, chen2025sharegpt4v, wang2024visionllm, chen2023tem, huang2024segment, liu2024universal, gao2023llama, cha2024honeybee}.

However, the community soon recognised that the vision capabilities integrated through these small adapter layers outside the LLM (in LLaVA-like approaches) remain insufficient \cite{tong2024eyes}. To address this gap, \citet{covert2024locality, karamcheti2024prismatic} refined vision encoders to align more closely with pretrained vision models.  
\citet{jiang2023clip, kar2025brave} proposed merging multiple visual encoders with projection layers before feeding them into the LLM.  
Separately, \citet{tong2024cambrian, shi2024eagle, zong2024mova} explored merging a larger number of diverse vision expert models with fine-tuned projection layers, integrating the combined vision features either before the LLM or within the LLM transformer layers, respectively.  

Diffusion Models (DMs), as vision experts, have gained recent attention. \citet{wang2024diffusion} use DM’s image generation loss to enhance the visual encoder. Other approaches merge VLMs with DM’s image generation, either by fine-tuning DMs with VLM outputs \cite{tong2024metamorph, hernandez2024generative}, sharing a central transformer \cite{shi2025lmfusionadaptingpretrainedlanguage, chen2025janusprounifiedmultimodalunderstanding}, or integrating DMs with LLM transformers \cite{zhou2024transfusion}.

Despite attempts to integrate DMs and VLMs with minimal modifications to their internal architectures, \emph{one overlooked aspect in prior work is the role of self- and cross-attention layers within DM and LLM Transformers}. These layers govern the interplay between multi-modal tokens, yet have not been closely examined. Although DMs and the LLM component in VLMs share the same foundational Transformer architecture \cite{vaswani2017attention, dosovitskiy2020image}, they exhibit markedly different text-to-region alignment due to their distinct optimization objectives. Notably, DMs demonstrate stronger alignment than VLMs, as shown in \cref{fig:empirical}.

\section{Diffusion Instruction Tuning}
\label{method}

\begin{figure*}[ht!]
    \centering
    \begin{minipage}[t]{0.58\linewidth}
        \centering
        \includegraphics[width=\linewidth]{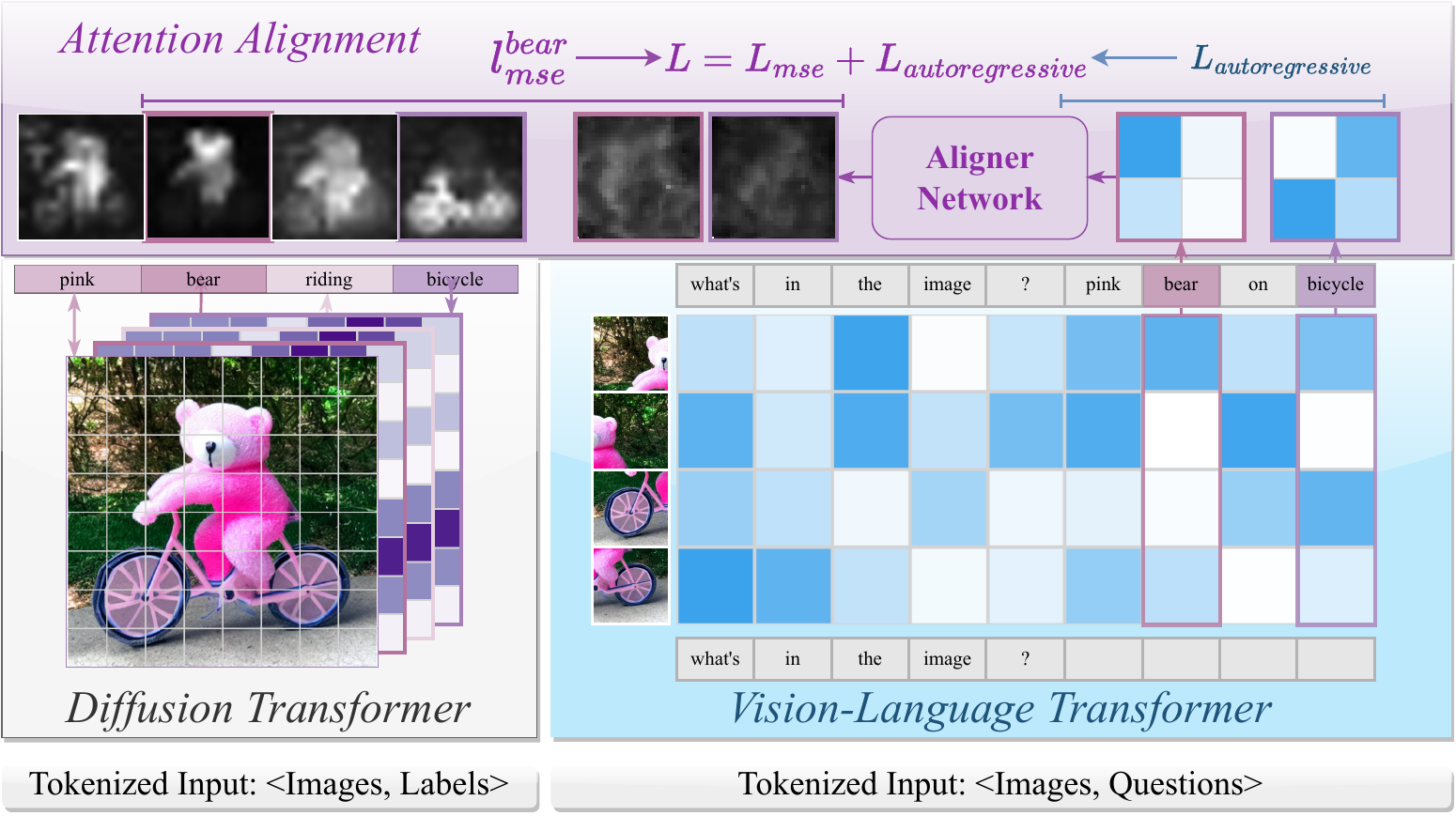}
        \vspace{-20pt}
        \caption{Sketch of Diffusion Instruction Tuning (left) and a short pseudo code (right), whose full version is available in Appendix \ref{sec:full_pseudo_code}.}
        \label{fig:diffag_overview}
    \end{minipage}%
    \hspace{0.02\linewidth}
    \begin{minipage}[t]{0.38\linewidth}
        \vspace{-168pt}
        \begin{algorithm}[H]
        \scriptsize
        \caption{\small Diffusion Instruction Tuning}
        \label{alg:diffusion_instruction_tuning_short}
        \begin{algorithmic}
        \REQUIRE $D=\{(x^{(i)}, y^{(i)})\}$, $\theta_D$, $\theta$, scale $\lambda$
        \ENSURE Fine-tuned VLM parameters $\theta$

        \textbf{Stage 1: Precompute DM Attention (run once)}\\
        \FOR{each $(x^{(i)}, y^{(i)}) \in D$}
        \STATE $A_{\text{DM}}^{(i)} \leftarrow p_{\text{DM}}(a \mid x^{(i)}, y^{(i)}; \theta_D)$
        \ENDFOR

        \textbf{Stage 2: Fine-Tune VLM}\\
        \REPEAT
            \STATE Sample batch $B \subseteq D$, set $L_{\text{VLM}}(\theta)=0, L_{\text{att}}(\theta)=0$
            \FOR{each $(x^{(i)}, y^{(i)}) \in B$}
                \STATE Compute $p_{\text{VLM}}(a \mid x^{(i)}, y^{(i)}; \theta)$
                \STATE $\delta^{(i)}(\theta) \leftarrow \text{Aligner}\bigl(p_{\text{VLM}}(a)\bigr)\;-\;A_{\text{DM}}^{(i)}$
                \STATE $L_{\text{VLM}}(\theta) {+}{=}-\log p(y_l^{(i)} \mid x^{(i)}, y_q^{(i)}; \theta)$
                \STATE $L_{\text{att}}(\theta) {+}{=}\|\delta^{(i)}(\theta)\|^2$
            \ENDFOR
            \STATE $L_{\text{total}}(\theta) \leftarrow L_{\text{VLM}}(\theta) + \lambda L_{\text{att}}(\theta)$
            \STATE Update $\theta \leftarrow \theta - \eta \nabla_\theta L_{\text{total}}(\theta)$
        \UNTIL{convergence}
        \label{alg:algorithm_1}
        \end{algorithmic}
        \end{algorithm}
    \end{minipage}
\end{figure*}

We aim to enhance a pretrained Vision-Language Model (VLM) by leveraging attention distributions from a pretrained Diffusion Model (DM). We assume there is an \emph{ideal} attention distribution that maximises VLM performance and that the DM's attention is closer to this ideal distribution.

\subsection{Models and Notation}

\noindent\textbf{Vision-Language Model (VLM).}
Let \(\theta\) be the VLM parameters, pretrained for tasks such as image captioning or question-answering. It models:
\begin{equation}
p(y_l | x, y_q; \theta),
\end{equation}
where $x, y_q, y_l$, as image, question, and label answer.

\noindent\textbf{Diffusion Model (DM).}
Let \(\theta_D\) be the DM parameters, which remain fixed during our procedure. It models:
\begin{equation}
p(x | y; \theta_D).
\end{equation}

\noindent\textbf{Attention Distributions.}
We write:
\begin{equation}
p_{\text{VLM}}(a | x, y; \theta), \quad p_{\text{DM}}(a | x, y; \theta_D),
\end{equation}
We hypothesise that $p_{\text{DM}}(a | x, y; \theta_D)$ is closer to the optimal posterior attention distribution $p^*(a | x, y)$ than $p_{\text{VLM}}(a | x, y; \theta)$, and the two can be aligned by projecting \( p_{\text{VLM}} \) into a comparable space using small learnable layers.

\subsection{Assumptions}
\label{sec:assumption}
\emph{Ideal attention in Vision-Centric Tasks:} An attention distribution \(p^*(a | x, y)\) minimises the next-token prediction loss of VLM, \(L_{\textit{VLM}}\);
\emph{DM Attention Proximity:} Empirically, the DM’s attention is more concentrated (lower entropy) and hence closer to \(p^*\) than the VLM’s, supported by \cref{fig:empirical}, experiments in \cref{sec:empirical_verification} and detailed justifications in \cref{sec:dm_attn_proximity_app};
\emph{Shared Dataset:} Both models use the same image-text set \(\{(x^{(i)}, y^{(i)})\}\);
\emph{Fixed DM Parameters:} \(\theta_D\) is kept fixed; only \(\theta\) is updated;
\emph{Pretrained VLM Parameters:} \(\theta\) is further fine-tuned with an attention alignment loss.

\subsection{Bayesian Derivation}
\label{sec:bayesian_main}
Our objective is to update the VLM parameters $\theta$ such that the model not only performs well on its primary task but also aligns its attention mechanism with that of the DM. We formalise this objective within a Bayesian framework.

\textbf{Posterior Distribution:}
We aim to find the posterior distribution of the VLM parameters given the data $D$ and the DM's attention distributions:
\begin{equation}
p(\theta | D, A_{\text{DM}}) \propto p(D | \theta) \, p(A_{\text{DM}} | \theta) \, p(\theta),
\label{eq:posterior}
\end{equation}
where $A_{\text{DM}} = \{ p_{\text{DM}}(a | x^{(i)}, y^{(i)}; \theta_D) \}$ is the collection of attention outputs derived from the DM's conditional distribution, and $p(\theta)$ is the prior over the VLM parameters.

\textbf{Likelihood of the Data:}
The likelihood of the data given $\theta$ is:
\begin{equation}
p(D | \theta) = \prod_{i} p(y_l^{(i)} | x^{(i)}, y_q^{(i)}; \theta).
\end{equation}
The negative log-likelihood corresponds to the standard loss function $L_{\text{VLM}}(\theta)$ used to fine-tune the VLM:
\begin{equation}
L_{\text{VLM}}(\theta) = -\sum_{i} \log p(y_l^{(i)} | x^{(i)}, y_q^{(i)}; \theta).
\end{equation}

\textbf{Likelihood of the DM's Attention:}
We model the likelihood of observing the DM's attention given the VLM's parameters, denoted as $p(A_{\text{DM}} | \theta)$. To simplify the notation and make the equations more concise, we introduce:
\begin{equation}
\footnotesize
\delta^{(i)}(\theta) = p_{\text{VLM}}(a \,|\, x^{(i)}, y^{(i)}; \theta) - p_{\text{DM}}(a \,|\, x^{(i)}, y^{(i)}; \theta_D).
\end{equation}
This represents the pointwise difference between the VLM's and DM's attention distributions for the \( i \)-th data point, serving as a measure of divergence at each attention location \( a \).
Assuming that these differences are Gaussian-distributed with equal variance, the likelihood can be expressed as:
\begin{equation}
p(A_{\text{DM}} | \theta) \propto \exp\left( -\frac{\lambda}{2} \sum_{i} \left\| \delta^{(i)}(\theta) \right\|^2 \right).
\end{equation}

This corresponds to the attention alignment loss $L_{\text{att}}(\theta)$:
\begin{equation}
L_{\text{att}}(\theta) = \sum_{i} \left\| \delta^{(i)}(\theta) \right\|^2.
\end{equation}
By minimizing \( L_{\text{att}}(\theta) \), i.e., the MSE loss, we encourage the VLM's attention to align with that of the DM, guiding it toward the optimal posterior attention distribution \( p^*(a | x, y) \).

For simplicity, we assume a non-informative prior \(p(\theta)\). Consequently, the posterior distribution in the previous equation~\cref{eq:posterior} is governed primarily by \(p(D|\theta)\) and \(p(A_{\text{DM}}|\theta)\). If regularisation is needed, a more informative prior can be seamlessly incorporated.
Combining the terms, the negative log-posterior (up to a constant) becomes:
\begin{equation}
L_{\text{total}}(\theta) = L_{\text{VLM}}(\theta) + \lambda L_{\text{att}}(\theta).
\label{eq:total_loss}
\end{equation}
Here, $\lambda$ balances the importance of aligning the attention distributions with the primary task. We fully justify the inclusion of the attention alignment loss in Appendix \ref{sec:justification_full}.

\subsection{Interpretation and Practical Implementation}
By minimizing \( L_{\text{total}}(\theta) \), we maximize \( p(\theta | D, A_{\text{DM}}) \), aligning the VLM's attention closer to the ideal posterior \( p^*(a | x, y) \). This improves the VLM’s ability to associate textual tokens with relevant visual regions, enhancing vision-text understanding.  
To implement this, we extract per-word attention distributions from the pretrained DM and VLM:
\begin{align}
p_{\text{DM}}(a | x^{(i)}, y^{(i)}; \theta_D), \quad p_{\text{VLM}}(a | x^{(i)}, y^{(i)}; \theta).
\end{align}
Fine-tuning minimizes:
\begin{equation}
\footnotesize
L_{\mathrm{total}}(\theta) \;=\; 
-\sum_i \log p\bigl(y_l^{(i)} | x^{(i)}, y_q^{(i)};\theta\bigr)
\;+\;\lambda \sum_i \|\delta^{(i)}(\theta)\|^2.
\end{equation}
This model-agnostic process, outlined in Algorithm \cref{alg:algorithm_1}, requires no additional data and works with various VLM architectures.

\section{Attention Alignment}
\label{sec:aggregations}

We discuss how to compute per-word attention in VLMs and DMs. Although both employ attention to capture vision-text interplay, their attention aggregation differs (\cref{fig:diffag_overview}). Understanding these distinctions is key to effective alignment.

\subsection{Attention Aggregation in Diffusion Models}
\label{sec:aggregation_dm}

Text-guided diffusion models generate images from textual input by iteratively denoising a random-noise image. During each denoising step, \emph{cross-attention} layers enable the model to focus on relevant textual tokens. Specifically, queries $Q$ are derived from the noisy image $x_t$, while keys $K$ and values $V$ come from the text embedding $v$:
\begin{align}
Q = f_Q(x_t), \quad K = f_K(v), \quad V = f_V(v),
\end{align}
where $f_Q$, $f_K$, and $f_V$ are pretrained projection matrices of DM. The attention map $M$ is then computed as:
\begin{equation}
M = \text{Softmax}\bigl((QK^\top)/\!\sqrt{d}\bigr).
\end{equation}
Prior work~\cite{hertz2022prompt} shows that averaging these maps across layers and time steps reveals meaningful correspondences between words and image regions. The resulting per-word attention distributions $p_{\text{DM}}(a | x, y; \theta_D)$ indicate salient image regions for each token, as shown in \cref{fig:empirical}. We leverage these maps as a proxy for the optimal posterior attention distribution $p^*(a | x, y)$, guiding the VLM’s alignment toward more focused vision-text interactions.

\subsection{Attention Aggregation in Vision-Language Models}
\label{sec:aggregation_vlm}

Vision-language models (VLMs) use transformer attention to connect text tokens (\(T_t\)) with image patch tokens (\(T_p\)) across multiple heads and layers, forming attention weights \(w_{(t,p)}^{hl}\), where $h \in H, l \in L, t \in T, p \in P$, with $H, L, T, P$ being heads, layers, tokens, and patches, respectively. These weights capture semantic and spatial relationships between tokens and patches.

\begin{figure}[htp]
    \raggedright
    \includegraphics[width=\linewidth,trim={1cm 0cm 0cm 0cm},clip]{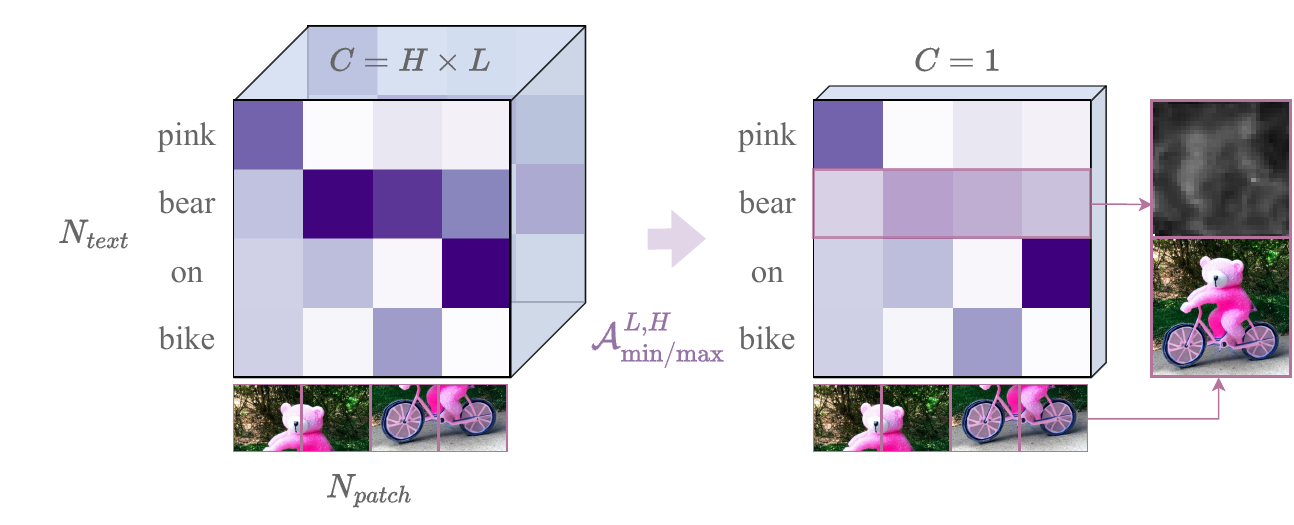}
    \vspace{-6mm}
    \caption{\textbf{Illustration of attention aggregation in VLMs.} Attention weights between text tokens and image patches are aggregated to form per-word saliency maps.}
    \label{fig:aggregate-reconstruct}
\end{figure}

To create per-word saliency maps, we first aggregate attention across all heads and layers, reducing the \((N_{\text{text}} \times N_{\text{patch}} \times H \times L)\) tensor to a \((N_{\text{text}} \times N_{\text{patch}})\) matrix. Next, we reshape this matrix into a \(\sqrt{N_{\text{patch}}} \times \sqrt{N_{\text{patch}}}\) grid, which approximately reconstructs the spatial layout of the original image (\cref{sec:reconstruction_matter}).
This process generates interpretable saliency maps, where each row highlights how a text token focuses on image patches. These maps allow us to align VLM attention with that of the Diffusion Model (DM), as shown in \cref{fig:aggregate-reconstruct}.
To explore the best attention aggregation mechanisms for VLMs, we employ various approaches.

\subsubsection{Simple aggregation functions}
\label{sec:attn_simple}

A straightforward approach is to pool attention weights $\mathbf{A}$ (i.e., $w_{(t,p)}^{hl}$) across heads $H$ and layers $L$ via mean or max. We consider four strategies:
\[
\mathcal{A}_{\text{mean/max}}^{(L, H)}(\mathbf{A}) \in \{\scriptsize{\text{max-max},\;\text{max-mean},\;\text{mean-max},\;\text{mean-mean}}\},
\]
where each denotes a combination of \(\{\text{Max}, \text{Mean}\}\) over $H$ and $L$. This yields a single per-word attention map, capturing a coarse measure of word-to-patch alignment.

\subsubsection{Attention Flow}
\label{sec:attn_flow}

Proposed by \citet{abnar2020quantifying}, \emph{attention flow} accumulates attention across multiple layers to track how information propagates through the network. Unlike simple pooling, this method captures deeper dependencies by considering all layers together.

To capture interactions across layers, we recursively update the attention map. Starting with the first layer’s attention \( A^{(1)} \), we merge each subsequent layer \( A^{(l)} \) using element-wise multiplication or addition:
\[
\bar{A} \,\leftarrow\, \bar{A} \circ A^{(l)} \quad \text{or} \quad \bar{A} \,\leftarrow\, \bar{A} + A^{(l)}.
\]
This method, previously applied by \citet{lin2024training} at the \textit{sentence level}, is extended here to finer-grained \textit{word-level attention maps}. By aggregating attention across layers, this approach may highlight semantic relationships that simpler methods overlook. Further details are in Appendix~\ref{sec:attention_flow}.

\subsubsection{Learning the Attention Aggregations}
\label{sec:attn_learn}

Standard approaches aggregate attention using predefined pooling methods, which may lose fine-grained relationships. Instead, we introduce \emph{parallel} cross-attention parameters ($W_{Q_d},W_{K_d}$) alongside the pretrained projections $(W_Q,W_K)$. This enables us to \emph{capture richer semantic correlations without overwriting existing weights}.

\begin{figure}
    \centering
    \includegraphics[width=1\linewidth]{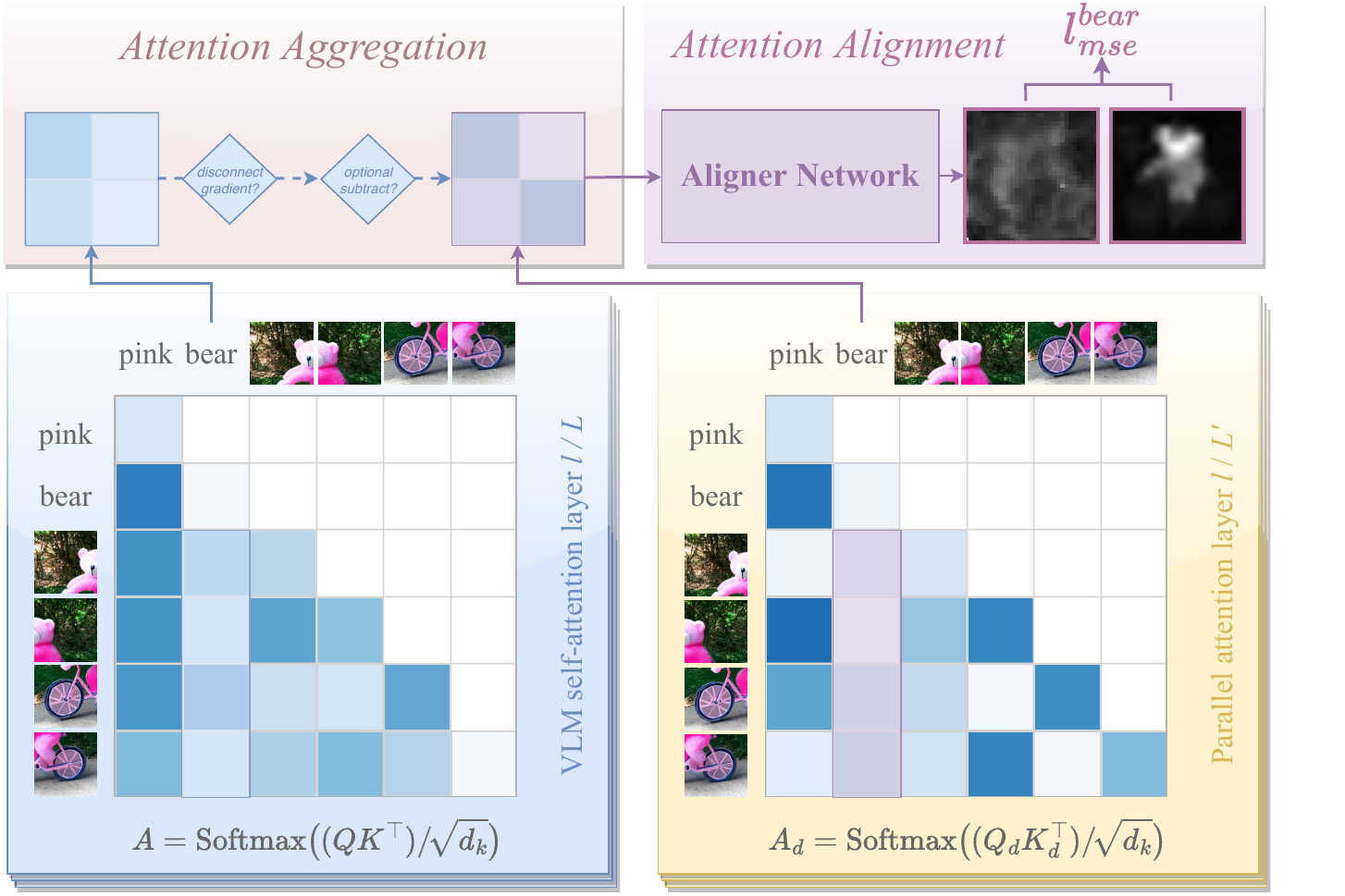}
    \caption{\textbf{Learning to aggregate with parallel attention.} The demonstration is based on a self-attention VLM. The parallel attention constitutes about $1/5^{th}$ of the total VLM layers ($L' \in L$).}
    \label{fig:learn_attn_self_attn}
\end{figure}

To retain the benefits of the pretrained attention while incorporating new learned patterns, we compute both the original attention $A$ and a parallel attention $A_d$ during each forward pass. The parallel attention $A_d$ is then used to align with the DM:
\begin{align}
A \;=\;\text{Softmax}\!\bigl((QK^\top)/\sqrt{d_k}\bigr), \\
A_d \;=\;\text{Softmax}\!\bigl((Q_dK_d^\top)/\sqrt{d_k}\bigr).
\end{align}
Empirically, we find that learning parallel cross-attention in just $1/5^\text{th}$ of the layers (see \cref{fig:learn_attn_self_attn}) is sufficient for effective alignment. This approach balances efficiency with accuracy, preserving the VLM’s core knowledge while benefiting from the DM’s attention distributions.

\subsection{Aligner Network}
\label{sec:aligner_network}

To improve attention alignment between the Vision-Language Model (VLM) and the Diffusion Model (DM), we introduce a lightweight Aligner network. This network refines the parallel (or aggregated) attention \( A_d \) into a single-channel map, making it directly comparable to the DM's attention \( p_{\text{DM}}(a | x,y;\theta_D) \). Inspired by Squeeze-and-Excitation networks~\cite{hu2018squeeze}, it efficiently transforms attention representations while preserving key semantic information.
The Aligner network consists of several small (3-5) layers, using either MLPs or convolutions. First, it expands the representation to capture richer features, applies non-linear transformations, and then compresses it back into a single-channel attention map. 
Empirically, we found \emph{convolutional layers better capture local spatial cues than MLP}, detailed comparisons are provided in Appendix~\ref{sec:extra_results_ofm}. During fine-tuning, the Aligner output is compared to the DM’s attention via:
\begin{equation}
L_{\text{att}}(\theta') = \sum_{i} \bigl\|\mathrm{Aligner}\bigl(A_d^{(i)}\bigr)\;-\;p_{\text{DM}}(a | x^{(i)},y^{(i)};\theta_D)\bigr\|^2,
\end{equation}
guiding the VLM’s attention toward the DM’s more focused distribution, capturing complex semantic correlations while preserving the original pretrained parameters.

\subsection{Lavender Integration}
\label{sec:labender_integration}

\paragraph{Cross-Attention.}
For VLMs with dedicated cross-attention layers, each head produces word-to-patch weights \(w_{(t,p)}^{hl}\) mapping text tokens \(T_t\) to image patches \(T_p\). We can reshape these weights into spatial grids and aggregate across heads/layers, then apply the Aligner network to yield final per-word saliency maps comparable to DM attention.
To ensure consistency, we process the extracted attention weights as follows:
1) Interpolate the attention weights to form a roughly square matrix.
2) Arrange the tiles (if the image is represented as multiple tiles) into a coherent spatial layout (see \cref{fig:attention_reconstruction_tiles}).
3) Resize the maps to a standard resolution (e.g., \(32 \times 32\)) for downstream processing.

\paragraph{Self-Attention Only.}
When both text and image patches are interleaved in a single sequence, tokens attend to each other in a bidirectional or causal manner. To extract word-to-patch correlations, we must:
1) Identify which subset of tokens corresponds to text and which correspond to image patches.
2) Apply a causal or bidirectional mask to exclude irrelevant attention connections.
3) Reshape and interpolate the extracted attention weights to reconstruct a meaningful spatial representation.
This process involves selecting the appropriate text and vision token indices, interpolating attention maps into a square grid, resizing to a fixed resolution (e.g., \(32 \times 32\)), and optionally incorporating the Aligner network output \( A_d \) or merged attention maps. This process is demonstrated in \cref{fig:learn_attn_self_attn}.
By following this procedure, we extract a meaningful per-word attention map aligned with the DM attention, enabling Lavender to improve vision-text alignment even in models that rely exclusively on self-attention.

\section{Implementation}
\label{sec:implementation}

We integrate Lavender with three VLMs—cross-attention VLMs (OpenFlamingo, Llama 3.2-11B-Vision Instruct) and self-attention VLMs (MiniCPM-Llama 3-v2.5)—using Stable Diffusion v1.4 \cite{rombach2021highresolution} to provide per-word attention targets. 
We introduce a lightweight Aligner network and an attention alignment loss to guide the VLM toward the DM’s more focused distributions.

1) \emph{DM Attention Extraction.}
We extract per-word attention maps from Stable Diffusion v1.4 by applying a shortened inversion process~\cite{mokady2022nulltextinversioneditingreal, jin2023image} using the paired image-label data. 
Attention maps are collected at each denoising step and reshaped into a fixed \(32 \times 32\) resolution. 
For efficiency, we limit inversion to 5 steps and diffusion to 10 steps, enabling each image to be processed in ~20 seconds on a single V100 GPU.
2) \emph{Cross-Attention VLMs.}
For OpenFlamingo and Llama 3.2-11B-Vision Instruct, which use cross-attention layers, we integrate Lavender by wrapping the attention layers to extract per-word attention maps. These are then aligned with DM-derived distributions using the Aligner network and attention alignment loss.
3) \emph{Self-Attention VLMs.}
For MiniCPM-Llama 3-v2.5, which relies solely on self-attention, we extract word-to-patch correlations by identifying text and image tokens in the self-attention layers. 
We reshape and interpolate the extracted attention maps into a structured spatial layout, aligning them with DM attention in the same way as cross-attention models.
4) \emph{Computing Environment.}
Experiments are conducted on NVIDIA GPUs (V100, A10G, A100) using PyTorch. 
For efficient training, we use DeepSpeed with MiniCPM-Llama 3-v2.5 and Fully Sharded Data Parallel (FSDP) for Llama 3.2-11B-Vision Instruct. 
All models are fine-tuned with Lavender while maintaining their original training procedures and hyperparameters.

\section{Training Strategies and Dataset Preparation}
\label{sec:train_recipe}

\paragraph{Training Strategies.}
We adopt several strategies to stabilise alignment objectives and preserve a VLM’s pretrained capabilities:
1) \emph{Pretraining the Aligner Network.} Before updating all parameters, we optionally \emph{pretrain only the Aligner network}, freezing the rest of the model to absorb DM-based attention signals without disrupting existing representations. This stabilises early training and prevents immediate overwriting of pretrained features.
2) \emph{Attention Aggregation and Normalisation.} We experiment with different attention aggregation techniques (Section~\ref{sec:aggregations}) and apply instance or batch normalisation within the Aligner network to control variance, making the final attention distributions more stable and interpretable.
3) \emph{Configurable Aligner Network.} The Aligner is configurable in depth, allowing trade-offs between efficiency and accuracy. We experiment with “light” (single-layer), “sim” (two-layer), and “deep” (four-layer) variants to balance computational complexity and performance.
4) \emph{Parameter-Efficient Fine-Tuning (PEFT).} We incorporate PEFT (e.g., LoRA) to prevent catastrophic forgetting by restricting updates to a small, controlled parameter subset. This improves stability and preserves pretrained VLM knowledge.
5) \emph{Sampling Strategies.} We define two approaches for selecting predicted words for alignment: 
  a) \emph{Root word match} (allows lemmatized matches, reducing over-strict alignment constraints).  
  b) \emph{Exact word match} (requires exact token match for stricter supervision).  

\paragraph{Dataset Preparation.}
We process images with Stable Diffusion to obtain per-word attention targets across four datasets:
1) \emph{Flickr30k} \cite{young2014image}: 31,783 images, 158,915 captions, focused on everyday human activities.
2) \emph{Laion50k} \cite{schuhmann2022laion}: A 50k subset of Laion-5B (5.85B image-text pairs), used for additional method verification.
3) \emph{RLAIF-V 83k} \cite{yu2024rlaifv}: A large-scale multimodal dataset provides 83,132 preference pairs covering multiple sources (MSCOCO, ShareGPT-4V, MovieNet, VQA variants).
4) \emph{OCRVQA30k}: A 30k subset of OCR-VQA, containing text-rich images with associated question-answer pairs.
All datasets are processed with DMs to extract attention maps and scaled across different VLMs, with Laion50k used primarily for OpenFlamingo validation.

\section{Experiments and Results}
\label{sec:exps_main}

We first validate Lavender on a small-scale setup with OpenFlamingo (\cref{sec:empirical_verification}) before scaling it to MiniCPMv2.5~\cite{yao2024minicpmv} and Llama 3.2-11B-Vision Instruct~\cite{dubey2024llama}. We evaluate these models on 20 standard VLM benchmarks, comparing them against 23 baseline models (\cref{sec:scaled_results}).  
To further analyse performance, we conduct data overlap analysis (\cref{sec:correlation}), investigate scaling behaviour and training efficiency (\cref{sec:scaling}), and examine out-of-distribution generalisation in medical QA benchmarks (\cref{sec:ood_benchmark}).  
Finally, we present qualitative results, including aligned attention visualisations and VQA examples, using Llama 3.2-11B (\cref{sec:qualitative}).

\subsection{Empirical Verification}
\label{sec:empirical_verification}

We begin by validating the key hypothesis that \emph{DM attention distributions are more concentrated and closely approximate an ideal posterior attention for VLMs} (\cref{sec:assumption}).  

\begin{figure}[ht]
    \centering
    \includegraphics[width=.8\linewidth,trim={0.5cm 0.5cm 0cm 0cm},clip]{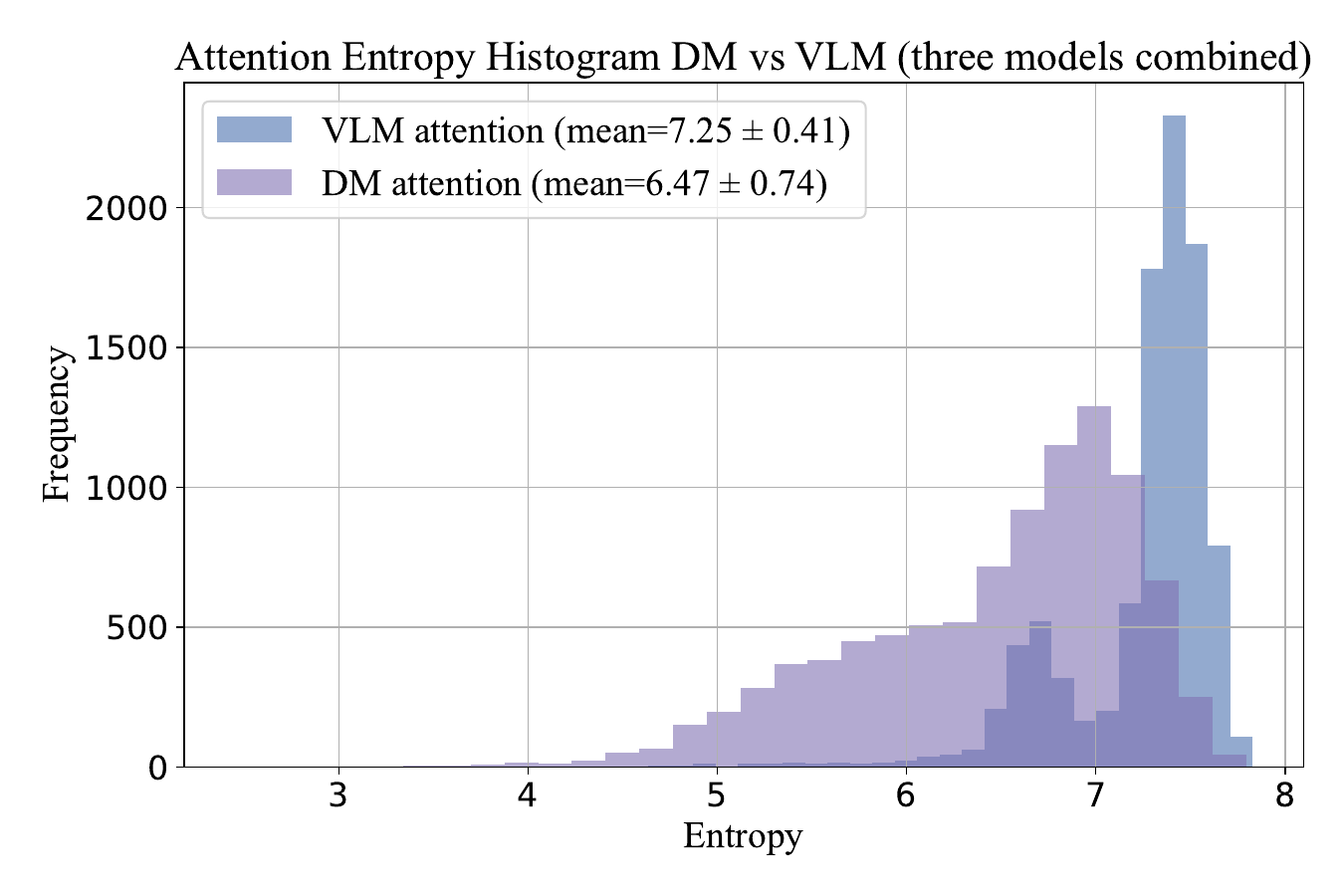}
    \vspace{-4mm}
    \caption{
    \textbf{DM attention is more concentrated than VLM attention, aligning closer to the optimal posterior distribution $p^*(a | x, y)$.} Measured across OpenFlamingo, MiniCPM-v2.5, and Llama 3.2-11B. Full results in Appendix~\ref{sec:extra_results_ofm}.
    }
    \label{fig:entropy_histogram_combined}
\end{figure}

\noindent\emph{1. DM attention has lower entropy than VLM attention.}  
We compare attention entropy across Flickr30k, RLAIF-V83k, and OCRVQA30k, using three VLMs (OpenFlamingo, MiniCPM-v2.5, and Llama 3.2-11B). As shown in \cref{fig:entropy_histogram_combined}, the DM’s attention entropy is consistently lower, supporting its proximity to the optimal posterior attention \( p^*(a | x, y) \) in vision-centric tasks.

\begin{figure}[ht]
    \centering
    \includegraphics[width=0.8\linewidth]{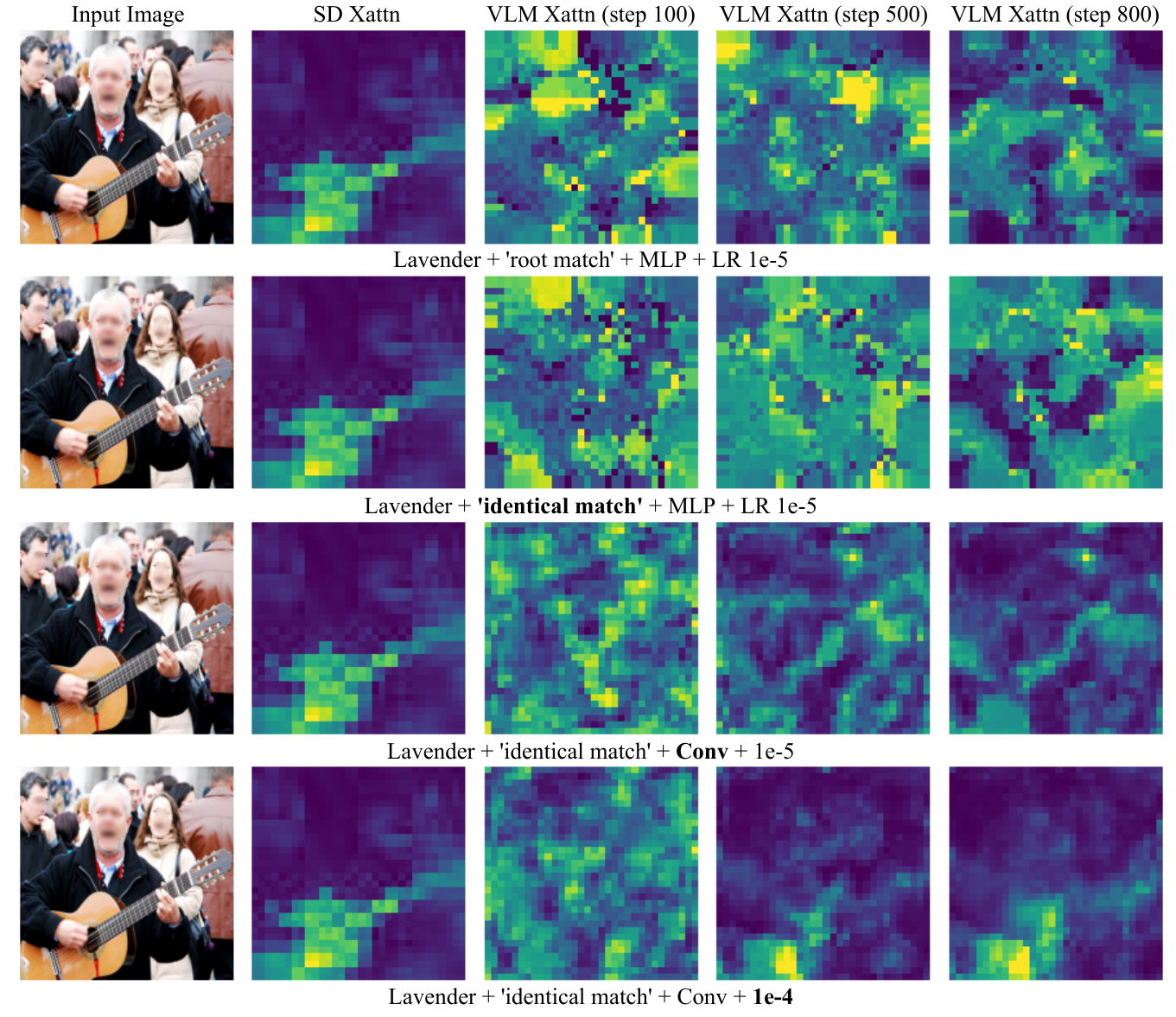}
    \vspace{-2mm}
    \caption{
    \textbf{Aligning VLM attention with DM for ``guitar.''} Each row adds a training technique. More examples in Appendix~\ref{sec:extra_results_ofm}.
    }
    \label{fig:visual_guitar}
\end{figure}

\vspace{-2mm}
\noindent\emph{2. Lavender aligns VLM attention with DM attention.}  
We fine-tune OpenFlamingo with Lavender on Flickr30k and visualise attention alignment over multiple training steps (\cref{fig:visual_guitar}). The best alignment is achieved using:  
1) \textit{Exact word match sampling} instead of root word match.  
2) \textit{Convolutional Aligner networks} instead of MLP.  
3) \textit{Moderate learning rates} (e.g., \(1\text{e}{-4}\) instead of \(1\text{e}{-5}\)).  

\begin{figure}[ht]
    \centering
    \includegraphics[width=.8\linewidth]{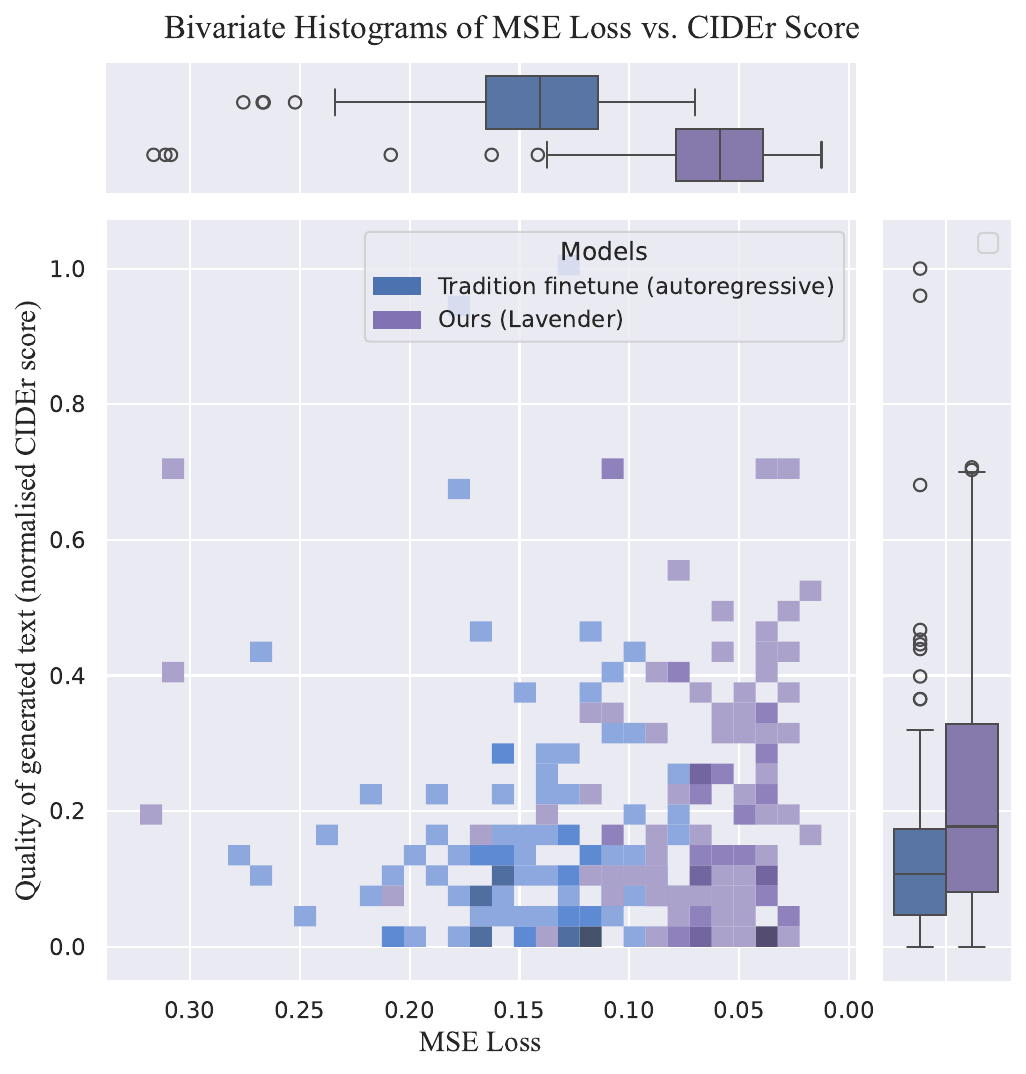}
    \vspace{-4mm}
    \caption{\textbf{Lower MSE aligns with higher caption quality.} Results on COCO, fine-tuning OpenFlamingo with Flickr30k.}
    \label{fig:calibration_mse}
\end{figure}

\noindent\emph{3. Jointly minimising \( L_{\text{VLM}} \) and \( L_{\text{att}} \) improves text generation.}  
To test whether Lavender enhances text generation quality, we train OpenFlamingo with Lavender on Flickr30k and Laion50k and evaluate COCO captioning scores. As shown in \cref{fig:calibration_mse}, \textit{MSE loss reduction correlates with higher text generation quality}, confirming that aligning attention improves downstream performance.

\begin{figure}[h]
    \centering
    \includegraphics[width=.8\linewidth]{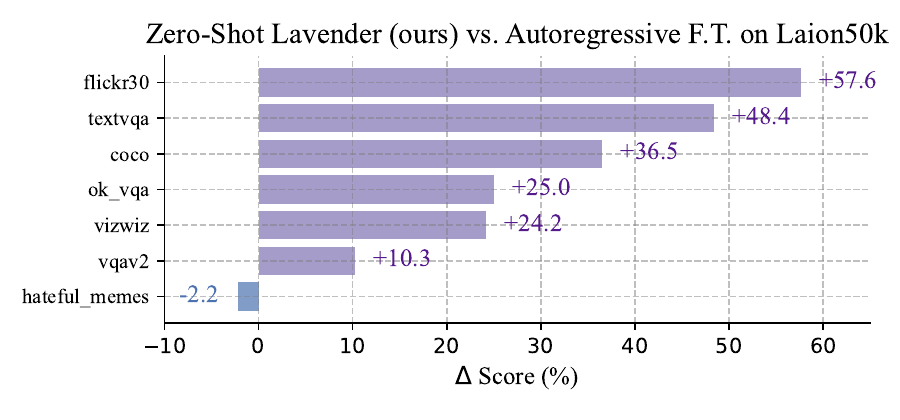}
    \includegraphics[width=.8\linewidth]{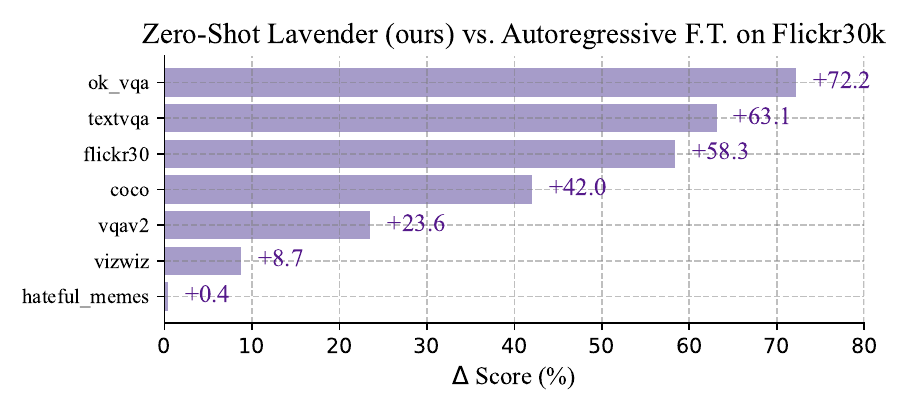}
    \vspace{-6mm}
    \caption{
    \textbf{Lavender surpasses autoregressive fine-tuning by up to 72\% across seven zero-shot benchmarks.} Models are trained on either the Laion50k or Flickr30k dataset.
    }
    \label{fig:openflamingo_benchmark}
\end{figure}

\noindent\emph{4. Lavender boosts zero-shot performance across multiple benchmarks.}  
We compare OpenFlamingo fine-tuned with Lavender against standard autoregressive fine-tuning on seven VLM benchmarks, covering captioning (COCO~\cite{chen2015microsoft} and Flickr30K~\cite{young2014image}), VQA (VQAv2~\cite{antol2015vqa}, OK-VQA~\cite{marino2019ok}, TextVQA~\cite{singh2019towards}, and VizWiz~\cite{gurari2018vizwiz}), and ranking (HatefulMemes~\cite{kiela2020hateful}). Results in \cref{fig:openflamingo_benchmark} show \textit{up to 72\% improvement in zero-shot accuracy}, demonstrating Lavender’s effectiveness in vision-text alignment.

\subsection{Scaled Results with Lavender}
\label{sec:scaled_results}
After validating Lavender on small-scale fine-tuning with OpenFlamingo, we scale experiments using the Llama 3-based MiniCPMv2.5 and Llama 3.2-11B-Vision-Instruct implementations.  
Fine-tuning is conducted on a mixture of RV83k, Flk30k, and OV30k datasets, using both autoregressive and Lavender methods, with LoRA or full fine-tuning strategies.  
We evaluate these models on \textit{20 multimodal benchmarks}, comparing against \textit{23 baseline VLMs}.

\textbf{Evaluation on Multimodal Benchmarks}
Lavender is evaluated across \textit{20 VLM benchmarks}, grouped into four categories:
1) \textit{Chart, Diagram, and Document Understanding:}  
AI2D~\cite{kembhavi2016ai2d}, ChartQA~\cite{masry2022chartqa}, OCRBench~\cite{liu2023hidden}, OCRVQA~\cite{mishra2019ocr}, TextVQA~\cite{singh2019towards}, DocVQA~\cite{mathew2021docvqa}, and InfoVQA~\cite{mathew2022infographicvqa}.
2) \textit{Perception and Multi-Discipline Reasoning:} 
MME~\cite{fu2023mme}, MMBench (four subsets)~\cite{MMBench}, ScienceQA~\cite{saikh2022scienceqa}, MMStar~\cite{chen2024we}, and MMMU~\cite{yue2024mmmu}.
3) \textit{Real-World Visual Understanding:}  
RealworldQA~\cite{grok15v}, SEED~\cite{li2023seed}.
4) \textit{Hallucination Detection:}  
HallucinationBench~\cite{liu2023hallusionbench} and POPE~\cite{li2023evaluating}.
All benchmarks are evaluated using their default metrics.

\textbf{Baseline Models.}  
In addition to MiniCPMv2.5~\cite{yao2024minicpmv} and Llama 3.2-11B~\cite{dubey2024llama} and their fine-tuned variants as baselines, we include the following groups of VLMs and their performance on the above benchmarks as reference:  
1) \textit{Small Budget-Constrained Models.}  
This group consists of open-source VLMs with sizes smaller than 10B, using backbones from Vicuna-7B~\cite{zheng2023judging}, Qwen-7B~\cite{bai2023qwentechnicalreport}, or Llama3-8B~\cite{dubey2024llama}. Models include: LLaVA-1.5-7B~\cite{liu2024improved}, LLaVA-Next-7B~\cite{liu2024llavanext}, Mini-Gemini-7B~\cite{team2024gemini}, Eagle-X5-8B~\cite{shi2024eagle}, and Cambrian-1-8B~\cite{tong2024cambrian}.  
2) \textit{Small Data-Heavy SOTA Models.}  
This group includes open-source VLMs with sizes smaller than 20B, typically fine-tuned on datasets containing more than 5M samples. Models in this category are: LLaVA-OneVision-7B~\cite{li2024llava}, InternVL2-8B~\cite{chen2024internvl}, Qwen2-VL-7B~\cite{wang2024qwen2}, Molmo-7B~\cite{deitke2024molmo}, and Pixtral-12B~\cite{agrawal2024pixtral12b}.  
3) \textit{Large SOTA Models.}  
This group comprises large VLMs with sizes greater than 20B, including both open- and closed-source models, typically fine-tuned on extensive datasets (minimum 5M samples). Models include: Cambrian-1-34B~\cite{tong2024cambrian}, LLaVA-OneVision-72B~\cite{li2024llava}, Qwen2-VL-72B~\cite{wang2024qwen2}, Molmo-72B~\cite{deitke2024molmo}, Claude-3 Haiku~\cite{anthropic2024claude}, Claude-3.5 Sonnet~\cite{anthropic2024claude}, GPT-4V~\cite{gpt4}, GPT-4o~\cite{gpt4omini}, Gemini 1.5 Pro~\cite{team2024gemini}, and Llama-3.2-90B~\cite{li2024llava}.

\paragraph{Main Results with MiniCPM-V-2.5.}  
\cref{fig:performance_minicpm_lm32} (a) presents results for MiniCPM-V-2.5, a self-attention-only model based on Llama3. Compared to autoregressive fine-tuning, Lavender improves performance on 16 out of 20 tasks by up to 4\%, while limiting drops to a maximum of -1\%. The modest overall gains are expected due to dataset reuse (RV83k) and the inherent difficulty of applying text-vision alignment to a purely self-attention model. Nonetheless, \emph{Lavender consistently enhances performance even on saturated datasets, demonstrating its capacity to refine modern VLMs without additional data.}
\vspace{-4mm}
\paragraph{Main Results with Llama 3.2-11B-Vision-Instruct.}  
\cref{fig:performance_minicpm_lm32} (b) presents results for Llama 3.2-11B-Vision-Instruct, a cross-attention-equipped model fine-tuned with both LoRA and full fine-tuning. Lavender outperforms autoregressive fine-tuning by up to 30\% on 19 out of 20 benchmarks with LoRA and up to 25\% on 17 out of 20 benchmarks with full fine-tuning.  
These results highlight Lavender’s stronger improvements on cross-attention-based VLMs (Llama 3.2 and OpenFlamingo) compared to the self-attention-only MiniCPM-V-2.5, reinforcing that \emph{the principle of text-vision alignment underlying Lavender is fundamentally rooted in cross-attention mechanisms}

\begin{figure*}[h]
    \centering
    \includegraphics[width=1\linewidth]{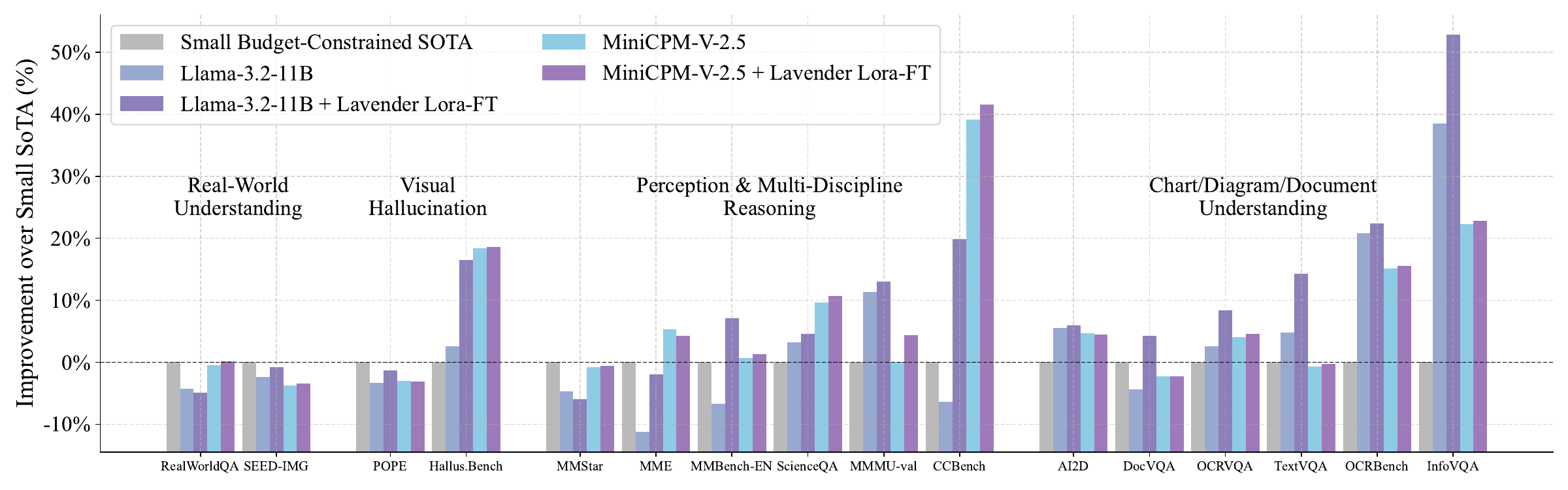}
    \caption{\textbf{Lavender improves MiniCPM-V-2.5 and Llama-3.2-11B, surpassing Small Budget-Constrained SOTA by up to 50\%.} This plot highlights the key zero-shot accuracy results across 16 VLM benchmarks from \cref{tab:main_table_full}, focusing on Small Budget-Constrained Models. Lavender shows its greatest improvements in \textit{Chart, Diagram, and Document Understanding}, relying on precise text-visual alignment in OCR tasks. Moderate gains are observed in \textit{Perception and Multi-Discipline Reasoning} and \textit{Hallucination}, while the weakest improvements occur in \textit{Real-World Visual Understanding}, which requires broader knowledge.}
    \label{fig:full_benchmark_highlight}
\end{figure*}

\paragraph{Benchmarking Lavender Against External Baselines.}  
Lavender enhances VLM performance beyond standard next-token prediction fine-tuning. To contextualise its improvements, we compare it against a broad set of baseline models across 16 benchmarks. \cref{fig:full_benchmark_highlight} presents key insights, while \cref{tab:main_table_full} provides detailed comparisons.

\begin{figure}[h]
    \centering
    \includegraphics[width=1\linewidth]{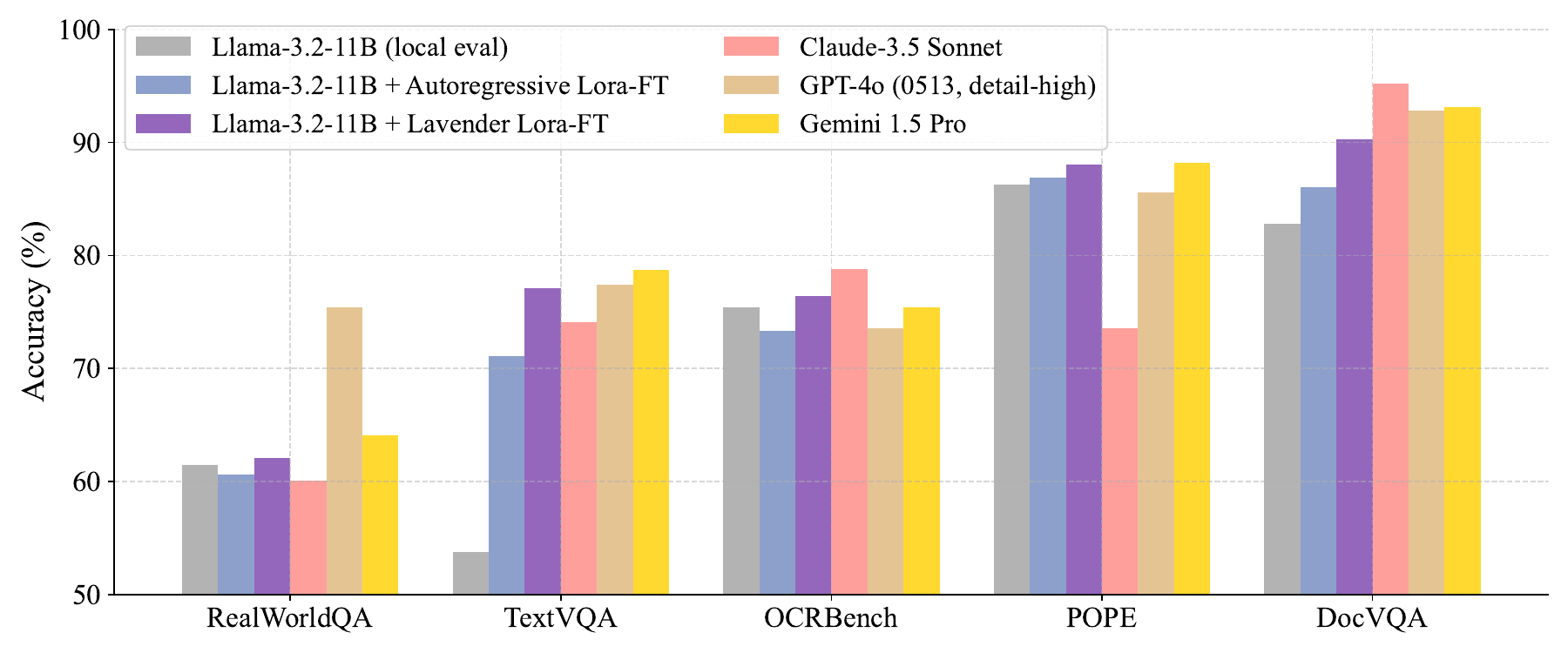}
    \vspace{-8mm}
    \caption{
     \textbf{Lavender demonstrates comparable performance to certain High-Resource State-of-the-Art Models.}
    }
    \label{fig:lavender_vs_close_sota}
\end{figure}

For \textit{Small Budget-Constrained Models}, Lavender outperforms most benchmarks, achieving up to 50\% gains, with minor deficits (within 4\%) on benchmarks like SEED-IMG. Among \textit{Small Data-Heavy SOTA Models}, although these models are trained on significantly larger datasets (38x–384x more data), it highlights improvements over the autoregressive baseline, with observed gaps largely attributed to dataset composition (\cref{sec:correlation}). Finally, for \textit{Large SOTA Models}, Lavender demonstrates competitive performance with select closed-source models an order of magnitude larger (e.g., Claude-3.5 Sonnet, GPT-4o, Gemini 1.5 Pro) on key benchmarks such as TextVQA, POPE, RealWorld, and DocVQA, as shown in \cref{fig:lavender_vs_close_sota}.

\subsection{Data Overlapping Analysis}  
\label{sec:correlation}  

\begin{figure}[H]
    \centering
    \includegraphics[width=1\linewidth]{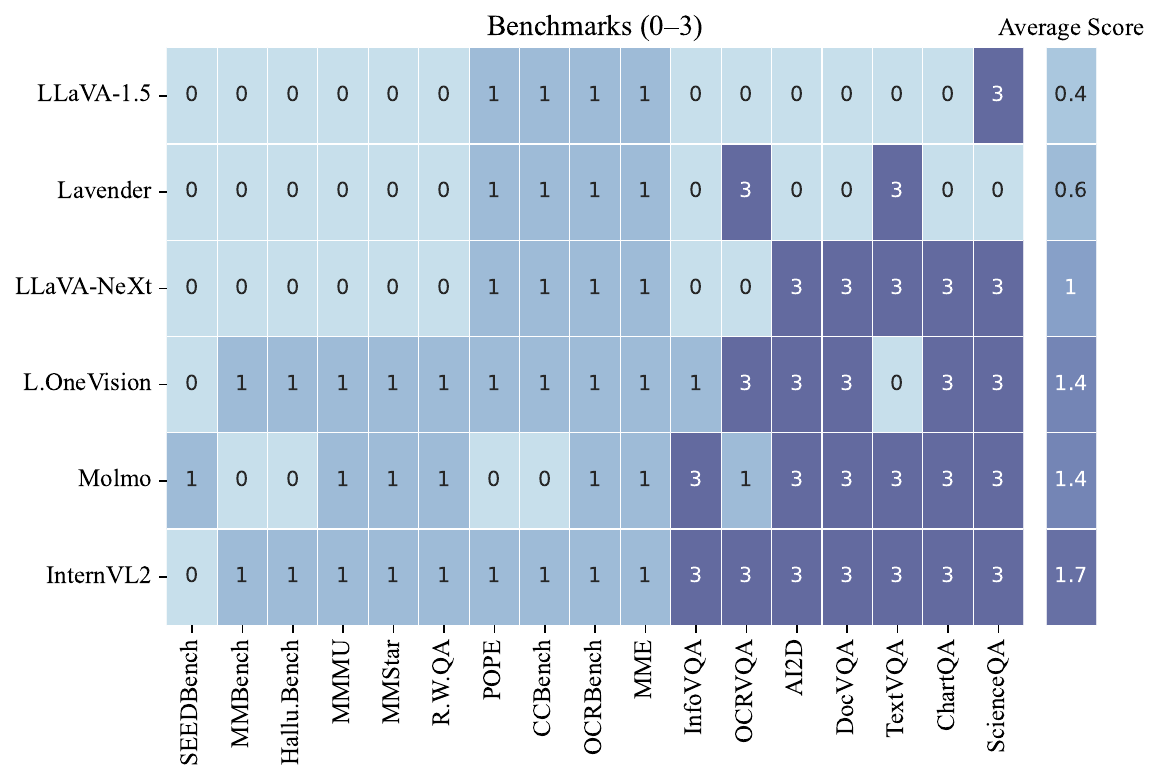}
    \caption{\textbf{Lavender exhibits minimal data overlap during fine-tuning, supporting its generalisability.}  
    Overlap scores are assigned as follows:  
    3 for explicitly shared benchmark datasets,  
    1 for datasets with shared sources (e.g., COCO) or web-crawled images,  
    and 0 for no overlap.  
    These estimates are based on publicly available dataset information and highlight Lavender’s robustness on out-of-distribution benchmarks.}
    \label{fig:data_contamination_qualitative}
\end{figure}

\paragraph{Qualitative Analysis.}
A key question arising from the above benchmark results is how to interpret the observed absolute numbers. In this section, we analyse the overlapping between the disclosed fine-tuning datasets and the benchmark datasets.  
We perform an ad-hoc qualitative analysis based on the composition of fine-tuning and benchmark datasets to reflect their potential overlapping, as detailed in \cref{fig:data_contamination_qualitative}. The qualitative results suggest that the fine-tuning datasets of some state-of-the-art models may exhibit stronger overlapping with the benchmark datasets compared to models fine-tuned on smaller datasets.  
Lavender's fine-tuning dataset shows an overlap score on the lower end, similar to LLaVA-1.5~\cite{liu2024improved}, which demonstrates its strong generalizability.
This simple approach does not exclude the possibility of data overlap during the pretraining stage, but our focus is on fine-tuning, where overfitting is more likely to occur with small fine-tuning datasets.

\FloatBarrier

\FloatBarrier
\subsection{Scaling Behaviour}
\label{sec:scaling}
Lavender is a model-agnostic approach tested on small fine-tuning datasets due to computational constraints. To assess scalability, we fine-tune Llama-3.2-11B on combinations of RV83k, Flk30k, and OV30k using autoregressive and Lavender methods with LoRA or full fine-tuning. \cref{fig:scaling_mean} presents average performance across eight benchmarks, with detailed results in \cref{fig:scaling_full}.  
The findings indicate that Lavender scales better with increased data, effectively reducing overfitting—a common issue with autoregressive fine-tuning on small datasets.

\FloatBarrier
\subsection{Severely Out-of-Distribution Medical Benchmark}
\label{sec:ood_benchmark}

\begin{figure}[H]
    \centering
    \includegraphics[width=1\linewidth]{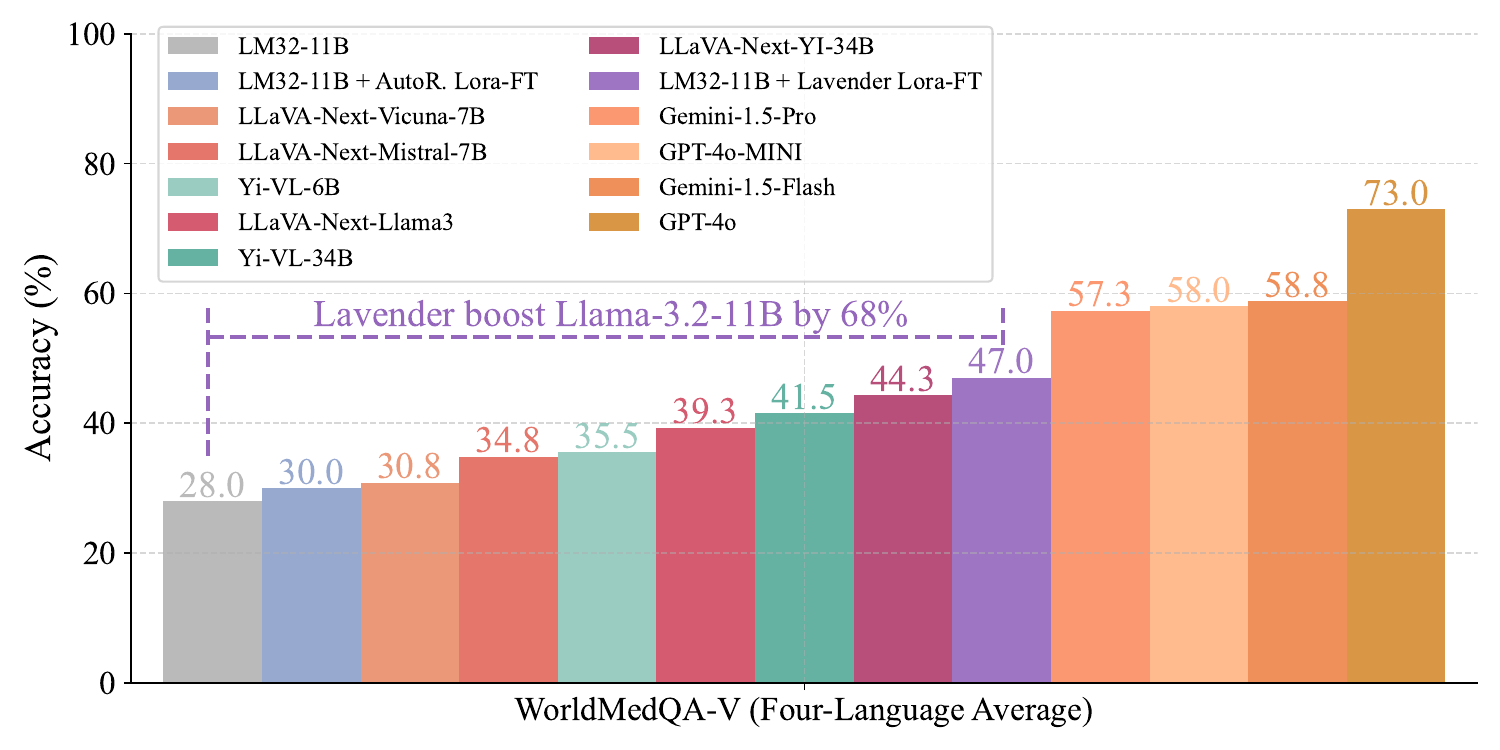}
    \caption{\textbf{Lavender boosts Llama-3.2-11B's performance on the OOD WorldMedQA benchmark by 68\%.} Results are based on fine-tuning Llama-3.2-11B-Vision-Instruct on RV83k, Flk30k, and OV30k using autoregressive or Lavender methods with LoRA. Accuracy reflects average performance across four low-resource medical VQA subsets, evaluated via Vlmevalkit \cite{duan2024vlmevalkit}. Example results in Appendix \cref{tab:world_med_qa_results}.}
    \label{fig:world_med_qa}
\end{figure}

We evaluate model generalisation on the extreme OOD WorldMedQA-V \cite{duan2024vlmevalkit}, a multilingual, multimodal medical VQA dataset. It includes 568 multiple-choice questions in four low-resource languages, paired with medical images. The dataset’s focus on medical exams and rare languages makes it ideal for testing generalisation beyond typical fine-tuning domains.
\cref{fig:world_med_qa} shows that Lavender improves Llama-3.2-11B's performance by 68\%, surpassing six open-source models (6B–34B) and narrowing the gap with large closed-source models from 43\% to 10\%.

\FloatBarrier
\begin{figure}[H]
    \centering
    \includegraphics[width=1\linewidth]{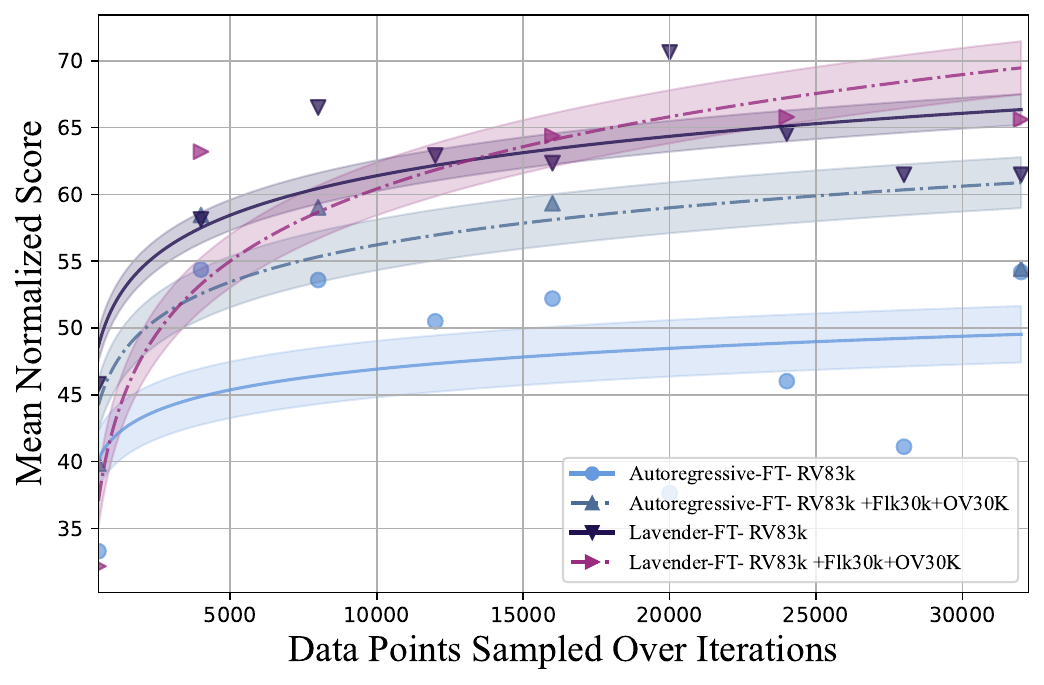}
    \caption{\textbf{Lavender scales better and mitigates overfitting compared to autoregressive fine-tuning, with larger datasets reducing variability.} 
    This plot shows the mean normalised performance across eight benchmarks for two dataset configurations after fine-tuning Llama 3.2-11B. Markers indicate observed performance with trendlines, and shaded regions show variability as 1 standard deviation around the mean. Full results with four dataset configurations are in Appendix \cref{fig:scaling_mean} and \cref{fig:scaling_full}.}
    \label{fig:scaling_mean_sim}
\end{figure}

\FloatBarrier
\begin{figure*}[h]
    \centering
    \includegraphics[width=1\linewidth]{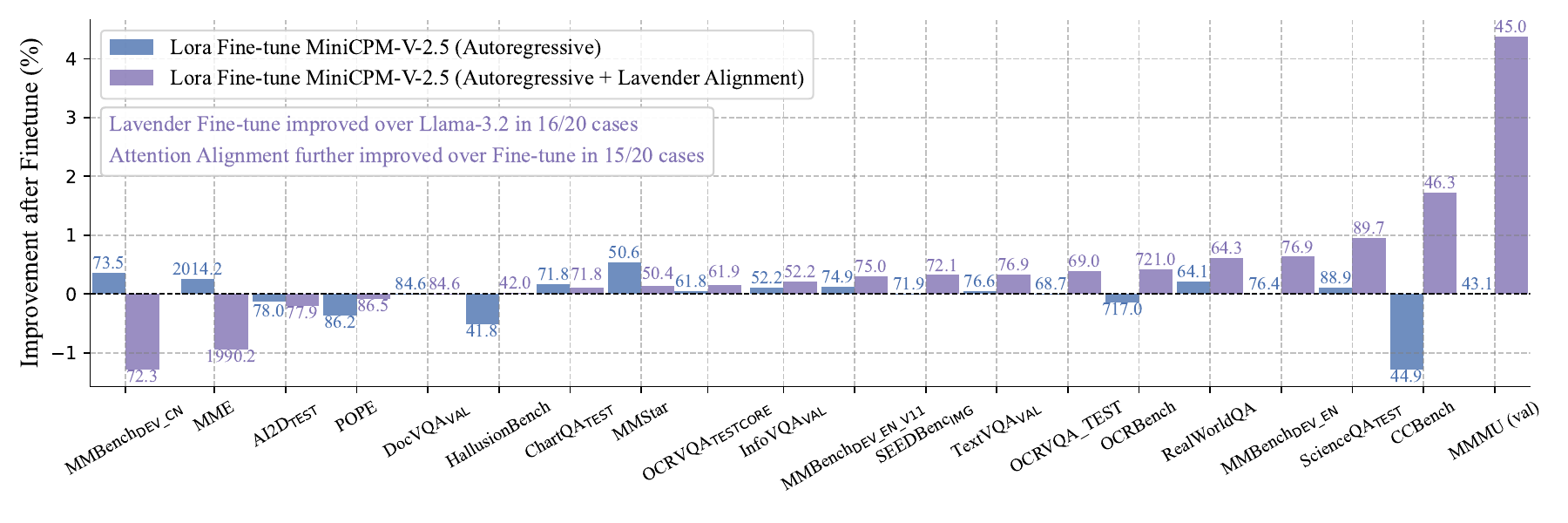}
    \\ {\small (a) Lavender enhances MiniCPM-Llama3-V-2.5 despite further fine-tuning on its pre-trained dataset.}
    \includegraphics[width=1\linewidth]{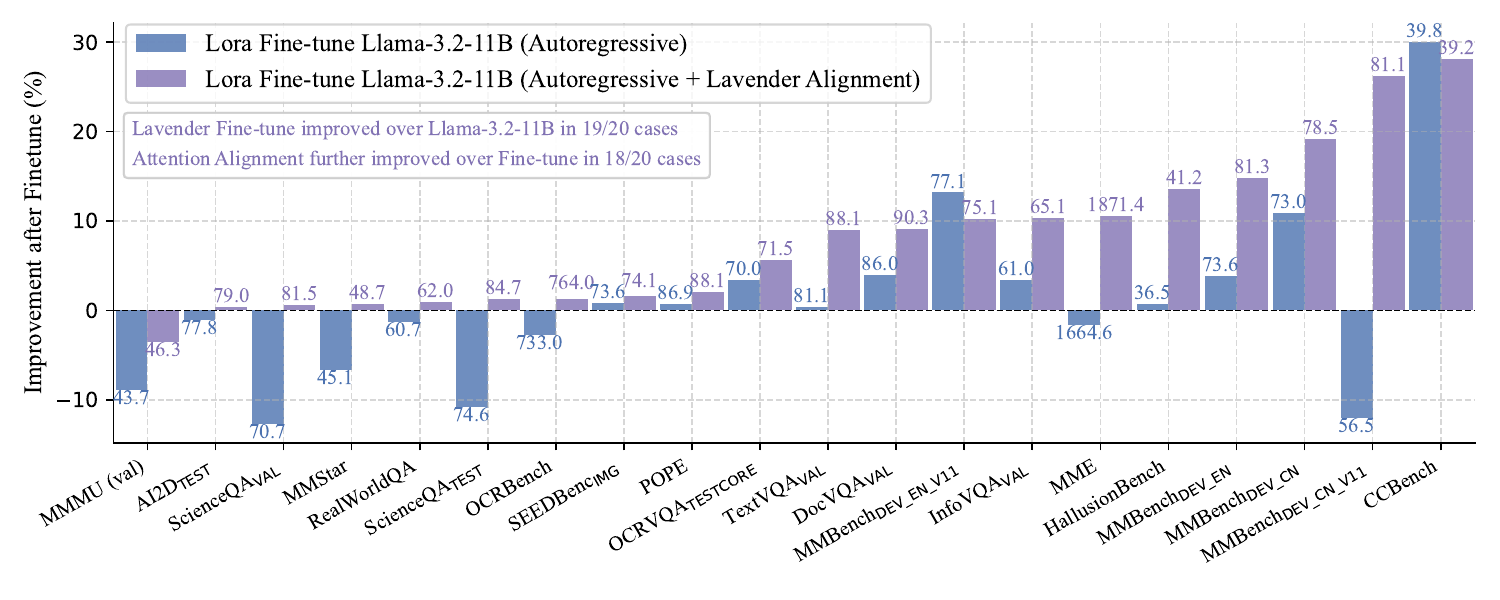}
    \includegraphics[width=1\linewidth]{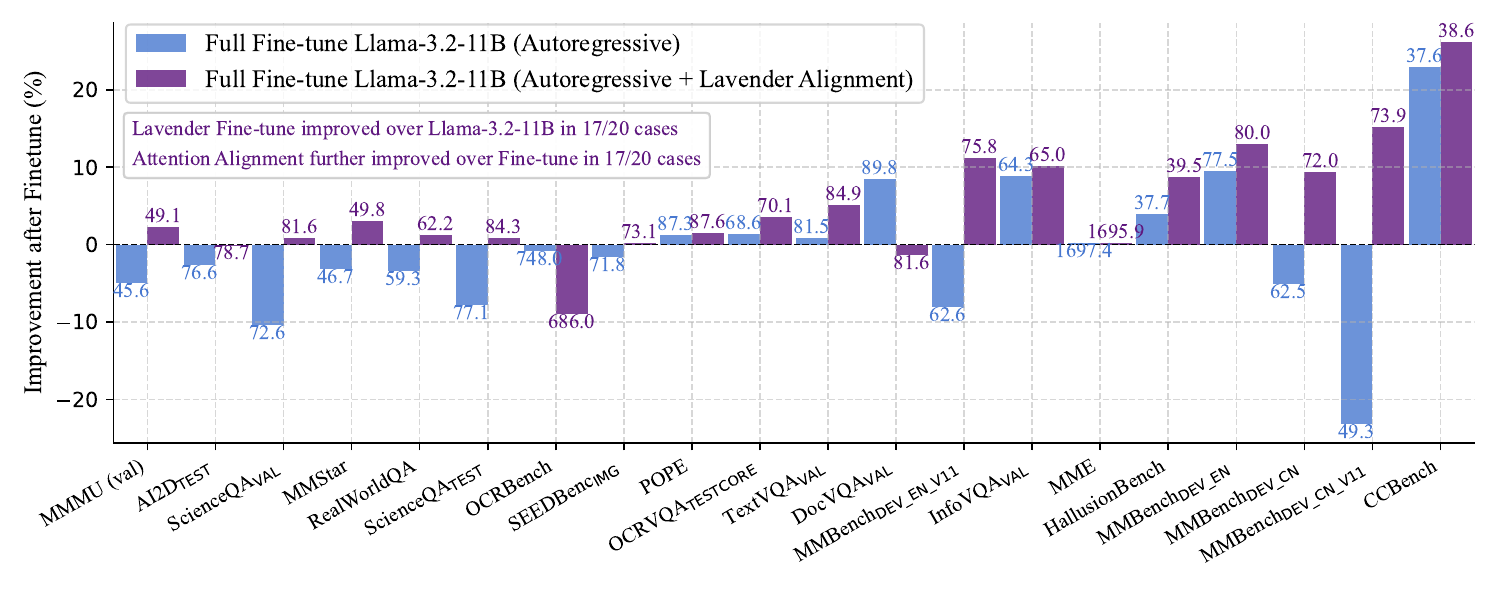}
    \\ {\small (b) Lavender mitigates catastrophic forgetting for Llama-3.2-11B fine-tuned on a new dataset.}
    \caption{\textbf{Relative Improvement Over Baseline on 20 Benchmarks.}  
    (a) Results for MiniCPM-Llama3-V-2.5 fine-tuned on its original RV83k dataset. Lavender improves performance on 16/20 benchmarks with gains up to 4\%, while limiting performance drops to -1\%, primarily on an out-of-distribution Chinese benchmark.  
    (b) Results for Llama-3.2-11B-Vision-Instruct fine-tuned on a mixture of RV83k, Flk30k, and OV30k using LoRA and full fine-tuning strategies. Lavender outperforms autoregressive fine-tuning on 18/20 (LoRA) and 17/20 (Full-FT) benchmarks, achieving up to 30\% and 25\% improvement, respectively, while mitigating catastrophic forgetting.  
    Across both models, Lavender demonstrates consistent benefits over autoregressive fine-tuning, particularly in reducing catastrophic forgetting.}
    \label{fig:performance_minicpm_lm32}
\end{figure*}

\definecolor{tablegray}{gray}{0.9} 
\definecolor{color_lora_auto}{rgb}{0.85, 0.90, 0.98}
\definecolor{color_lora_lav}{rgb}{0.88, 0.83, 0.90}
\definecolor{color_full_auto}{rgb}{0.66, 0.76, 0.92}
\definecolor{color_full_lav}{rgb}{0.76, 0.67, 0.81}

\begin{table*}[ht]
    \vspace{-0.5em}
    \linespread{1}
    \aboverulesep = 0.2em 
    \belowrulesep = 0.2em
    \scriptsize
    \centering
    \resizebox{\textwidth}{!}{
    \begin{tabular}{ccc|cccccccccccccccc} 
        Model & \hspace{-0.9em}Base & \hspace{-0.9em}FT Data & \rotatebox[origin=lb]{90}{\smash{AI2D}} & \rotatebox[origin=lb]{90}{\smash{CCBench}} & \rotatebox[origin=lb]{90}{\smash{DocVQA}} & \rotatebox[origin=lb]{90}{\smash{InfoVQA}} & \rotatebox[origin=lb]{90}{\smash{MMBench$^{EN}$}} & \rotatebox[origin=lb]{90}{\smash{MME}} & \rotatebox[origin=lb]{90}{\smash{MMMU-val}} & \rotatebox[origin=lb]{90}{\smash{MMStar}} & \rotatebox[origin=lb]{90}{\smash{OCRBench}} & \rotatebox[origin=lb]{90}{\smash{OCRVQA$^{TQ}$}} & \rotatebox[origin=lb]{90}{\smash{POPE}} & \rotatebox[origin=lb]{90}{\smash{R.W.QA}} & \rotatebox[origin=lb]{90}{\smash{SEED-IMG}} & \rotatebox[origin=lb]{90}{\smash{ScienceQA}} & \rotatebox[origin=lb]{90}{\smash{TextVQA}} & \rotatebox[origin=lb]{90}{\smash{Hallu.Bench}} \\
        \midrule
        \multicolumn{19}{c}{\textbf{Small Budget-Constrained Models (self-attention only)}} \\
        \midrule
        LLaVA-1.5-7B & \hspace{-0.9em}Vicuna-7B & \hspace{-0.9em}0.15M & \hspace{-0.5em}55.5 & \hspace{-0.9em}17.8 & \hspace{-0.9em}28.1 & \hspace{-0.9em}25.8 & \hspace{-0.9em}66.5 & \hspace{-0.9em}1510.0 & \hspace{-0.9em}35.7 & \hspace{-0.9em}33.1 & \hspace{-0.9em}318.0 & \hspace{-0.9em}60.6 & \hspace{-0.9em}86.1 & \hspace{-0.9em}54.8 & \hspace{-0.9em}58.6 & \hspace{-0.9em}69.2 & \hspace{-0.9em}58.2 & \hspace{-0.9em}27.6 \\ 
        LLaVA-NeXt-7B & \hspace{-0.9em}Vicuna-7B & \hspace{-0.9em}0.76M & \hspace{-0.5em}67.0 & \hspace{-0.9em}24.3 & \hspace{-0.9em}74.4 & \hspace{-0.9em}37.1 & \hspace{-0.9em}67.4 & \hspace{-0.9em}1519.0 & \hspace{-0.9em}37.6 & \hspace{-0.9em}37.6 & \hspace{-0.9em}532.0 & \hspace{-0.9em}63.8 & \hspace{-0.9em}87.5 & \hspace{-0.9em}57.8 & \hspace{-0.9em}70.2 & \hspace{-0.9em}70.3 & \hspace{-0.9em}64.9 & \hspace{-0.9em}27.6 \\ 
        Mini-Gemini-7B & \hspace{-0.9em}Qwen-7B & \hspace{-0.9em}1.5M & \hspace{-0.5em}- & \hspace{-0.9em}- & \hspace{-0.9em}- & \hspace{-0.9em}- & \hspace{-0.9em}65.8 & \hspace{-0.9em}1523.0 & \hspace{-0.9em}36.8 & \hspace{-0.9em}- & \hspace{-0.9em}477.0 & \hspace{-0.9em}- & \hspace{-0.9em}- & \hspace{-0.9em}- & \hspace{-0.9em}- & \hspace{-0.9em}71.1 & \hspace{-0.9em}65.2 & \hspace{-0.9em}- \\ 
        Cambrian-1-7B & \hspace{-0.9em}Vicuna-7B & \hspace{-0.9em}10M & \hspace{-0.5em}74.6 & \hspace{-0.9em}23.7 & \hspace{-0.9em}47.9 & \hspace{-0.9em}40.8 & \hspace{-0.9em}74.6 & \hspace{-0.9em}1802.9 & \hspace{-0.9em}41.8 & \hspace{-0.9em}\textbf{50.7} & \hspace{-0.9em}614.0 & \hspace{-0.9em}66.0 & \hspace{-0.9em}86.4 & \hspace{-0.9em}60.0 & \hspace{-0.9em}73.3 & \hspace{-0.9em}81.0 & \hspace{-0.9em}77.1 & \hspace{-0.9em}30.6 \\ 
        Eagle-X5-7B & \hspace{-0.9em}Vicuna-7B & \hspace{-0.9em}0.93M & \hspace{-0.5em}73.6 & \hspace{-0.9em}28.4 & \hspace{-0.9em}\textbf{86.6} & \hspace{-0.9em}- & \hspace{-0.9em}68.8 & \hspace{-0.9em}1866.0 & \hspace{-0.9em}37.6 & \hspace{-0.9em}41.7 & \hspace{-0.9em}551 & \hspace{-0.9em}64.3 & \hspace{-0.9em}\textbf{89.3} & \hspace{-0.9em}63.8 & \hspace{-0.9em}73.6 & \hspace{-0.9em}71.2 & \hspace{-0.9em}71.9 & \hspace{-0.9em}35.4 \\ 
        LLaVA-1.5-8B & \hspace{-0.9em}Lama3-8B & \hspace{-0.9em}0.15M & \hspace{-0.5em}69.9 & \hspace{-0.9em}27.8 & \hspace{-0.9em}32.4 & \hspace{-0.9em}27.5 & \hspace{-0.9em}- & \hspace{-0.9em}1825.5 & \hspace{-0.9em}39.2 & \hspace{-0.9em}46.1 & \hspace{-0.9em}420.0 & \hspace{-0.9em}61.0 & \hspace{-0.9em}87.3 & \hspace{-0.9em}56.7 & \hspace{-0.9em}70.1 & \hspace{-0.9em}72.2 & \hspace{-0.9em}- & \hspace{-0.9em}28.7 \\ 
        LLaVA-Next-8B & \hspace{-0.9em}Lama3-8B & \hspace{-0.9em}0.76M & \hspace{-0.5em}72.8 & \hspace{-0.9em}32.7 & \hspace{-0.9em}78.5 & \hspace{-0.9em}38.2 & \hspace{-0.9em}74.8 & \hspace{-0.9em}1908 & \hspace{-0.9em}43.1 & \hspace{-0.9em}43.9 & \hspace{-0.9em}531.0 & \hspace{-0.9em}60.7 & \hspace{-0.9em}87.1 & \hspace{-0.9em}58.4 & \hspace{-0.9em}72.5 & \hspace{-0.9em}73.1 & \hspace{-0.9em}65.3 & \hspace{-0.9em}33.1 \\ 
        Cambrian-1-8B & \hspace{-0.9em}Lama3-8B & \hspace{-0.9em}10M & \hspace{-0.5em}73.0 & \hspace{-0.9em}- & \hspace{-0.9em}77.8 & \hspace{-0.9em}42.6 & \hspace{-0.9em}75.9 & \hspace{-0.9em}1547.0 & \hspace{-0.9em}42.7 & \hspace{-0.9em}50.7 & \hspace{-0.9em}624.0 & \hspace{-0.9em}66.0 & \hspace{-0.9em}73.0 & \hspace{-0.9em}64.2 & \hspace{-0.9em}\textbf{74.7} & \hspace{-0.9em}73.1 & \hspace{-0.9em}71.7 & \hspace{-0.9em}30.6 \\ 
        \midrule
        MiniCPM-V-2.5 & \hspace{-0.9em}Lama3-8B & \hspace{-0.9em}0.08M & \hspace{-0.5em}\textbf{78.1} & \hspace{-0.9em}45.5 & \hspace{-0.9em}84.6 & \hspace{-0.9em}52.1 & \hspace{-0.9em}76.4 & \hspace{-0.9em}2009.1 & \hspace{-0.9em}43.1 & \hspace{-0.9em}50.3 & \hspace{-0.9em}718.0 & \hspace{-0.9em}68.7 & \hspace{-0.9em}86.6 & \hspace{-0.9em}63.9 & \hspace{-0.9em}71.9 & \hspace{-0.9em}88.8 & \hspace{-0.9em}76.6 & \hspace{-0.9em}41.9 \\ 
        \rowcolor{color_lora_auto}
        +Lora FT & \hspace{-0.9em}Lama3-8B & \hspace{-0.9em}0.08M & \hspace{-0.5em}77.9 & \hspace{-0.9em}44.9 & \hspace{-0.9em}84.6 & \hspace{-0.9em}52.2 & \hspace{-0.9em}76.4 & \hspace{-0.9em}\textbf{2014.2} & \hspace{-0.9em}43.1 & \hspace{-0.9em}50.6 & \hspace{-0.9em}717.0 & \hspace{-0.9em}68.8 & \hspace{-0.9em}86.2 & \hspace{-0.9em}64.1 & \hspace{-0.9em}71.9 & \hspace{-0.9em}88.9 & \hspace{-0.9em}76.6 & \hspace{-0.9em}41.8 \\ 
        \rowcolor{color_lora_lav}
        +Lavender FT & \hspace{-0.9em}Lama3-8B & \hspace{-0.9em}0.08M & \hspace{-0.5em}77.9 & \hspace{-0.9em}\textbf{46.3} & \hspace{-0.9em}84.6 & \hspace{-0.9em}\textbf{52.3} & \hspace{-0.9em}\textbf{76.9} & \hspace{-0.9em}1990.2 & \hspace{-0.9em}\textbf{45.0} & \hspace{-0.9em}50.4 & \hspace{-0.9em}\textbf{721.0} & \hspace{-0.9em}\textbf{69.0} & \hspace{-0.9em}86.5 & \hspace{-0.9em}\textbf{64.3} & \hspace{-0.9em}72.1 & \hspace{-0.9em}\textbf{89.7} & \hspace{-0.9em}\textbf{76.9} & \hspace{-0.9em}\textbf{42.0} \\
        \midrule
        \multicolumn{19}{c}{\textbf{Llama-3.2-11B (cross-attention)}} \\
        \midrule
        Llama-3.2-11B & \hspace{-0.9em}LM32-11B & \hspace{-0.9em}N.A. & \hspace{-0.5em}78.7 & \hspace{-0.9em}30.6 & \hspace{-0.9em}82.8 & \hspace{-0.9em}59.0 & \hspace{-0.9em}70.8 & \hspace{-0.9em}1692.9 & \hspace{-0.9em}48.0 & \hspace{-0.9em}48.3 & \hspace{-0.9em}754.0 & \hspace{-0.9em}67.7 & \hspace{-0.9em}86.3 & \hspace{-0.9em}61.4 & \hspace{-0.9em}72.9 & \hspace{-0.9em}83.6 & \hspace{-0.9em}80.8 & \hspace{-0.9em}36.3 \\ 
        \rowcolor{color_lora_auto}
        + AutoR. Lora-FT & \hspace{-0.9em}LM32-11B & \hspace{-0.9em}0.13M & \hspace{-0.5em}77.8 & \hspace{-0.9em}\textbf{39.8} & \hspace{-0.9em}86.0 & \hspace{-0.9em}61.0 & \hspace{-0.9em}73.6 & \hspace{-0.9em}1664.6 & \hspace{-0.9em}43.7 & \hspace{-0.9em}45.1 & \hspace{-0.9em}733.0 & \hspace{-0.9em}70.0 & \hspace{-0.9em}86.9 & \hspace{-0.9em}60.7 & \hspace{-0.9em}73.6 & \hspace{-0.9em}74.6 & \hspace{-0.9em}81.1 & \hspace{-0.9em}36.6 \\ 
        \rowcolor{color_lora_lav}
        + Lavender Lora-FT & \hspace{-0.9em}LM32-11B & \hspace{-0.9em}0.13M & \hspace{-0.5em}\textbf{79.0} & \hspace{-0.9em}39.2 & \hspace{-0.9em}\textbf{90.3} & \hspace{-0.9em}\textbf{65.1} & \hspace{-0.9em}\textbf{81.3} & \hspace{-0.9em}\textbf{1871.5} & \hspace{-0.9em}46.3 & \hspace{-0.9em}48.7 & \hspace{-0.9em}\textbf{764.0} & \hspace{-0.9em}\textbf{71.5} & \hspace{-0.9em}\textbf{88.1} & \hspace{-0.9em}62.1 & \hspace{-0.9em}\textbf{74.1} & \hspace{-0.9em}\textbf{84.7} & \hspace{-0.9em}\textbf{88.1} & \hspace{-0.9em}\textbf{41.3} \\ 
        \rowcolor{color_full_auto}
        + AutoR. Full-FT & \hspace{-0.9em}LM32-11B & \hspace{-0.9em}0.13M & \hspace{-0.5em}76.7 & \hspace{-0.9em}37.6 & \hspace{-0.9em}89.8 & \hspace{-0.9em}64.3 & \hspace{-0.9em}77.5 & \hspace{-0.9em}1697.4 & \hspace{-0.9em}45.6 & \hspace{-0.9em}46.8 & \hspace{-0.9em}748.0 & \hspace{-0.9em}68.6 & \hspace{-0.9em}87.3 & \hspace{-0.9em}59.4 & \hspace{-0.9em}71.8 & \hspace{-0.9em}77.1 & \hspace{-0.9em}81.5 & \hspace{-0.9em}37.7 \\ 
        \rowcolor{color_full_lav}
        + Lavender Full-FT & \hspace{-0.9em}LM32-11B & \hspace{-0.9em}0.13M & \hspace{-0.5em}78.7 & \hspace{-0.9em}38.6 & \hspace{-0.9em}81.6 & \hspace{-0.9em}65.0 & \hspace{-0.9em}80.0 & \hspace{-0.9em}1695.9 & \hspace{-0.9em}\textbf{49.1} & \hspace{-0.9em}\textbf{49.8} & \hspace{-0.9em}686.0 & \hspace{-0.9em}70.1 & \hspace{-0.9em}87.6 & \hspace{-0.9em}\textbf{62.2} & \hspace{-0.9em}73.1 & \hspace{-0.9em}84.3 & \hspace{-0.9em}84.9 & \hspace{-0.9em}39.5 \\ 
        \midrule
        \multicolumn{19}{c}{\textbf{Small Data-Heavy SOTA Models ($<$20 B) with Massive FT Data ($\geq$5M)}} \\
        \midrule
        \rowcolor{tablegray}
        L.OneVision-7B & \hspace{-0.9em}Qwen-7B & \hspace{-0.9em}5.2M & \hspace{-0.5em}82.4 & \hspace{-0.9em}54.9 & \hspace{-0.9em}87.5* & \hspace{-0.9em}68.8* & \hspace{-0.9em}83.2 & \hspace{-0.9em}1993.6 & \hspace{-0.9em}47.9 & \hspace{-0.9em}\textbf{61.9} & \hspace{-0.9em}622.0 & \hspace{-0.9em}64.7 & \hspace{-0.9em}\textbf{88.4} & \hspace{-0.9em}\textbf{69.9} & \hspace{-0.9em}\textbf{76.7} & \hspace{-0.9em}95.4 & \hspace{-0.9em}78.3* & \hspace{-0.9em}31.6 \\ 
        \rowcolor{tablegray}
        InternVL2-8B & \hspace{-0.9em}InternVL2-8B & \hspace{-0.9em}5M & \hspace{-0.5em}83.6 & \hspace{-0.9em}\textbf{77.1} & \hspace{-0.9em}91.6* & \hspace{-0.9em}74.8* & \hspace{-0.9em}81.7* & \hspace{-0.9em}2215.1 & \hspace{-0.9em}51.2 & \hspace{-0.9em}61.5 & \hspace{-0.9em}794.0 & \hspace{-0.9em}42.6 & \hspace{-0.9em}84.2 & \hspace{-0.9em}64.2 & \hspace{-0.9em}75.4 & \hspace{-0.9em}\textbf{97.1} & \hspace{-0.9em}77.4* & \hspace{-0.9em}45.0 \\ 
        \rowcolor{tablegray}
        Qwen2-VL-7B & \hspace{-0.9em}Qwen2-7B & \hspace{-0.9em}\textasciitilde50M & \hspace{-0.5em}83.0 & \hspace{-0.9em}65.7 & \hspace{-0.9em}\textbf{94.5}* & \hspace{-0.9em}\textbf{76.5}* & \hspace{-0.9em}\textbf{83.0}* & \hspace{-0.9em}\textbf{2276.3} & \hspace{-0.9em}\textbf{53.7} & \hspace{-0.9em}60.7 & \hspace{-0.9em}\textbf{843.0} & \hspace{-0.9em}\textbf{67.5} & \hspace{-0.9em}\textbf{88.4} & \hspace{-0.9em}68.5 & \hspace{-0.9em}76.0 & \hspace{-0.9em}85.5 & \hspace{-0.9em}\textbf{84.3}* & \hspace{-0.9em}\textbf{50.4} \\ 
        \rowcolor{tablegray}
        Molmo-7B-O & \hspace{-0.9em}Qwen2-7B & \hspace{-0.9em}\textasciitilde35M & \hspace{-0.5em}\textbf{90.7} & \hspace{-0.9em}20.6 & \hspace{-0.9em}90.8 & \hspace{-0.9em}70.0 & \hspace{-0.9em}69.1 & \hspace{-0.9em}1714.7 & \hspace{-0.9em}39.3 & \hspace{-0.9em}50.1 & \hspace{-0.9em}666.0 & \hspace{-0.9em}15.1 & \hspace{-0.9em}86.7 & \hspace{-0.9em}67.5 & \hspace{-0.9em}72.7 & \hspace{-0.9em}88.8 & \hspace{-0.9em}80.4 & \hspace{-0.9em}42.5 \\ 
        \rowcolor{tablegray}
        Pixtral-12B & \hspace{-0.9em}Nemo-12B & \hspace{-0.9em}N.A. & \hspace{-0.5em}79.0 & \hspace{-0.9em}37.6 & \hspace{-0.9em}90.7 & \hspace{-0.9em}50.8 & \hspace{-0.9em}77.9 & \hspace{-0.9em}1921.7 & \hspace{-0.9em}52.5 & \hspace{-0.9em}54.5 & \hspace{-0.9em}685.0 & \hspace{-0.9em}64.7 & \hspace{-0.9em}84.2 & \hspace{-0.9em}65.4 & \hspace{-0.9em}71.5 & \hspace{-0.9em}87.2 & \hspace{-0.9em}75.7 & \hspace{-0.9em}47.0 \\ 
        \midrule
        \multicolumn{19}{c}{\textbf{Large State-of-the-Art Models ($>$20 B) with Massive FT Data ($\geq$5M)}} \\
        \midrule 
        \rowcolor{tablegray}
        Cambrian-1-34B & \hspace{-0.9em}Yi-34B & \hspace{-0.9em}10M & \hspace{-0.5em}79.7 & \hspace{-0.9em}49.2 & \hspace{-0.9em}75.5 & \hspace{-0.9em}46.0 & \hspace{-0.9em}81.4 & \hspace{-0.9em}1689.0 & \hspace{-0.9em}49.7 & \hspace{-0.9em}54.2 & \hspace{-0.9em}600.0 & \hspace{-0.9em}68.2 & \hspace{-0.9em}79.7 & \hspace{-0.9em}67.8 & \hspace{-0.9em}75.3 & \hspace{-0.9em}76.8 & \hspace{-0.9em}76.7 & \hspace{-0.9em}41.6 \\ 
        \rowcolor{tablegray}
        L.OneVision-72B & \hspace{-0.9em}Qwen2-72B & \hspace{-0.9em}5.2M & \hspace{-0.5em}85.6 & \hspace{-0.9em}63.9 & \hspace{-0.9em}91.3 & \hspace{-0.9em}74.9 & \hspace{-0.9em}85.8 & \hspace{-0.9em}2257.4 & \hspace{-0.9em}56.8 & \hspace{-0.9em}65.8 & \hspace{-0.9em}741.0 & \hspace{-0.9em}- & \hspace{-0.9em}86.6 & \hspace{-0.9em}71.9 & \hspace{-0.9em}77.5 & \hspace{-0.9em}90.2 & \hspace{-0.9em}80.5 & \hspace{-0.9em}47.9 \\ 
        \rowcolor{tablegray}
        Qwen2-VL-72B & \hspace{-0.9em}Qwen2-72B & \hspace{-0.9em}\textasciitilde50M & \hspace{-0.5em}88.1 & \hspace{-0.9em}69.8 & \hspace{-0.9em}\textbf{96.5} & \hspace{-0.9em}\textbf{84.5} & \hspace{-0.9em}\textbf{86.5} & \hspace{-0.9em}\textbf{2482.7} & \hspace{-0.9em}64.5 & \hspace{-0.9em}\textbf{68.3} & \hspace{-0.9em}\textbf{877.0} & \hspace{-0.9em}\textbf{73.7} & \hspace{-0.9em}87.2 & \hspace{-0.9em}\textbf{77.8} & \hspace{-0.9em}\textbf{77.9} & \hspace{-0.9em}\textbf{91.2} & \hspace{-0.9em}\textbf{85.5} & \hspace{-0.9em}\textbf{58.1} \\ 
        \rowcolor{tablegray}
        Molmo-72B & \hspace{-0.9em}Qwen2-72B & \hspace{-0.9em}\textasciitilde35M & \hspace{-0.5em}\textbf{96.3} & \hspace{-0.9em}- & \hspace{-0.9em}93.5 & \hspace{-0.9em}81.9 & \hspace{-0.9em}79.4 & \hspace{-0.9em}1992.0 & \hspace{-0.9em}54.1 & \hspace{-0.9em}63.3 & \hspace{-0.9em}701.0 & \hspace{-0.9em}- & \hspace{-0.9em}- & \hspace{-0.9em}75.2 & \hspace{-0.9em}- & \hspace{-0.9em}- & \hspace{-0.9em}83.1 & \hspace{-0.9em}46.6 \\ 
        \rowcolor{tablegray}
        Claude-3 Haiku & \hspace{-0.9em}N.A. & \hspace{-0.9em}N.A. & \hspace{-0.5em}86.7 & \hspace{-0.9em}24.5 & \hspace{-0.9em}88.8 & \hspace{-0.9em}56.1 & \hspace{-0.9em}60.7 & \hspace{-0.9em}1920.0 & \hspace{-0.9em}50.2 & \hspace{-0.9em}38.1 & \hspace{-0.9em}658.0 & \hspace{-0.9em}- & \hspace{-0.9em}74.4 & \hspace{-0.9em}45.5 & \hspace{-0.9em}63.3 & \hspace{-0.9em}- & \hspace{-0.9em}67.3 & \hspace{-0.9em}39.2 \\ 
        \rowcolor{tablegray}
        Claude-3.5 Sonnet & \hspace{-0.9em}N.A. & \hspace{-0.9em}N.A. & \hspace{-0.5em}94.7 & \hspace{-0.9em}54.1 & \hspace{-0.9em}95.2 & \hspace{-0.9em}74.3 & \hspace{-0.9em}79.7 & \hspace{-0.9em}1920.0 & \hspace{-0.9em}68.3 & \hspace{-0.9em}62.2 & \hspace{-0.9em}788.0 & \hspace{-0.9em}- & \hspace{-0.9em}73.6 & \hspace{-0.9em}60.1 & \hspace{-0.9em}72.2 & \hspace{-0.9em}88.9 & \hspace{-0.9em}74.1 & \hspace{-0.9em}49.9 \\ 
        \rowcolor{tablegray}
        GPT-4V (0409) & \hspace{-0.9em}N.A. & \hspace{-0.9em}N.A. & \hspace{-0.5em}89.4 & \hspace{-0.9em}57.3 & \hspace{-0.9em}87.2 & \hspace{-0.9em}75.1 & \hspace{-0.9em}81.0 & \hspace{-0.9em}2070.2 & \hspace{-0.9em}63.1 & \hspace{-0.9em}56.0 & \hspace{-0.9em}656.0 & \hspace{-0.9em}- & \hspace{-0.9em}81.8 & \hspace{-0.9em}61.4 & \hspace{-0.9em}73.0 & \hspace{-0.9em}84.8 & \hspace{-0.9em}78.0 & \hspace{-0.9em}43.9 \\ 
        \rowcolor{tablegray}
        GPT-4o (0513) & \hspace{-0.9em}N.A. & \hspace{-0.9em}N.A. & \hspace{-0.5em}94.2 & \hspace{-0.9em}\textbf{71.2} & \hspace{-0.9em}92.8 & \hspace{-0.9em}79.2 & \hspace{-0.9em}83.4 & \hspace{-0.9em}2310.3 & \hspace{-0.9em}\textbf{69.1} & \hspace{-0.9em}63.9 & \hspace{-0.9em}736.0 & \hspace{-0.9em}- & \hspace{-0.9em}85.6 & \hspace{-0.9em}75.4 & \hspace{-0.9em}77.1 & \hspace{-0.9em}90.7 & \hspace{-0.9em}77.4 & \hspace{-0.9em}55.0 \\ 
        \rowcolor{tablegray}
        Gemini 1.5 Pro & \hspace{-0.9em}N.A. & \hspace{-0.9em}N.A. & \hspace{-0.5em}94.4 & \hspace{-0.9em}28.4 & \hspace{-0.9em}93.1 & \hspace{-0.9em}81.0 & \hspace{-0.9em}73.9 & \hspace{-0.9em}2110.6 & \hspace{-0.9em}62.2 & \hspace{-0.9em}59.1 & \hspace{-0.9em}754.0 & \hspace{-0.9em}12.3 & \hspace{-0.9em}\textbf{88.2} & \hspace{-0.9em}64.1 & \hspace{-0.9em}76.0 & \hspace{-0.9em}85.7 & \hspace{-0.9em}78.7 & \hspace{-0.9em}55.9 \\ 
        \rowcolor{tablegray}
        Llama-3.2-90B & \hspace{-0.9em}Llama-3.1-70B & \hspace{-0.9em}N.A. & \hspace{-0.5em}92.3* & \hspace{-0.9em}54.1 & \hspace{-0.9em}85.7 & \hspace{-0.9em}- & \hspace{-0.9em}80.4 & \hspace{-0.9em}1741.0 & \hspace{-0.9em}60.3 & \hspace{-0.9em}55.3 & \hspace{-0.9em}783.0 & \hspace{-0.9em}- & \hspace{-0.9em}86.3 & \hspace{-0.9em}68.2 & \hspace{-0.9em}76.8 & \hspace{-0.9em}87.1 & \hspace{-0.9em}- & \hspace{-0.9em}44.1 \\ 
        \bottomrule
        \end{tabular}
        }
    \caption{\textbf{Zero-shot accuracy of various fine-tuned models across 16 VLM benchmarks.} 
    Results are grouped into four sections based on base model size and the scale of the fine-tuning dataset. The top score for each benchmark within each group is highlighted in bold.
    Scores for MiniCPM-V-2.5, Llama-3.2-11B, and their autoregressive (AutoR.) and Lavender fine-tuned variants are locally evaluated using OpenCompass Vlmevalkit \cite{duan2024vlmevalkit} with `gpt-4o' as the evaluator. All other scores are sourced from the OpenCompass Multi-Modal Leaderboard, evaluated with the same Vlmevalkit. When leaderboard results are unavailable, the models' published numbers are cited, marked with a * notation if provided.
    Our observations are  categorized as follows:  
    1) \textit{Small Budget-Constrained Models.}  
    This group is most comparable to Lavender in terms of parameter size and fine-tuning data scale. Both Lavender versions outperform the majority of benchmarks with significant margins over the second-best-performing external models, achieving improvements of up to 40\% on CCBench. On a few benchmarks (e.g., MMStar, POPE, and SEED-IMG), Lavender is surpassed by baseline models, though the difference is within 4\% for SEED-IMG. Lavender-Llama-3.2 occasionally underperforms on MME due to privacy protection constraints (see \cref{tab:mme_failure}). 
    2) \textit{Small Data-Heavy SOTA Models.}  
    The primary argument for Lavender in this work is its ability to achieve improvements over the autoregressive fine-tuning baseline. Lavender implementations on MiniCPMv2.5 and Llama-3.2-11B are not designed to beat state-of-the-art results due to the limited fine-tuning data scale (0.13M), which is significantly smaller than the 5M to 50M datasets used for this group, approximately 38x to 384x larger than that used for Lavender.  
    We simplify the comparison by excluding pretraining dataset sizes for base models. In this group, Qwen2-VL-7B and LLaVA-OneVision-7B are the top performers, surpassing Lavender on most benchmarks. However, we note that the performance gap is likely influenced by the composition of fine-tuning datasets, as discussed in \cref{sec:correlation}.  
    3) \textit{Large SOTA Models.}  
    Finally, we include results from the latest state-of-the-art models, which are at least 20B in size (or unreleased models typically exceeding 100B) and fine-tuned on datasets of at least 5M samples.  
    Despite Qwen2-VL-72B outperforming all other models on 18/20 benchmarks, Lavender pushes the boundaries of Llama-3.2-11B, achieving performance comparable to certain closed-source models that are at least an order of magnitude larger (e.g., Claude-3.5 Sonnet, GPT-4o, and Gemini 1.5 Pro) on benchmarks such as TextVQA, POPE, RealWorld, and DocVQA.
    }
    \label{tab:main_table_full}
\end{table*}

\FloatBarrier

\vspace{-1mm}
\subsection{Qualitative Results with Llama 3.2-11B.}
\label{sec:qualitative}

\begin{figure}[h]
    \centering
    \setlength{\tabcolsep}{1pt} 
    \renewcommand{\arraystretch}{1} 
    \begin{tabular}{cc} 
        \adjustbox{valign=m}{\includegraphics[width=0.2\textwidth]{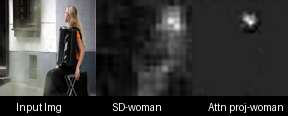}} &
        \adjustbox{valign=m}{\includegraphics[width=0.2\textwidth]{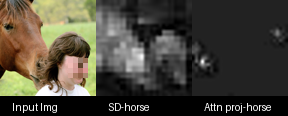}} \\
        \adjustbox{valign=m}{\includegraphics[width=0.2\textwidth]{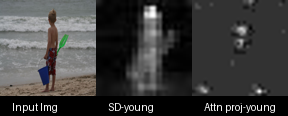}} &
        \adjustbox{valign=m}{\includegraphics[width=0.2\textwidth]{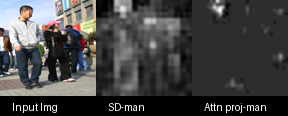}} \\
        \adjustbox{valign=m}{\includegraphics[width=0.2\textwidth]{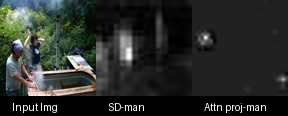}} &
        \adjustbox{valign=m}{\includegraphics[width=0.2\textwidth]{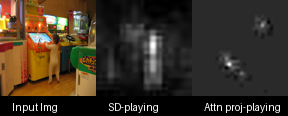}} 
    \end{tabular}
    \caption{\textbf{The per-word VLM attention maps are aligned to the Stable Diffusion (SD) after tuning with Lavender.} Results are from Lavender-Llama 3.2-11B implementation. More results are available in Appendix \cref{fig:attention_maps_full_fm}.}
    \label{fig:lm32_attn_4}
\end{figure}

\paragraph{Visual Results of Aligned Attention Maps.}  
A key argument of Lavender is its ability to align the attention maps of VLMs to those of DMs. \cref{fig:lm32_attn_4} provides examples of the cross-attention maps from Llama 3.2-11B after full fine-tuning with Lavender.  
The results indicate that the aligned VLM attention maps generally correlate with the semantic regions of corresponding words in a manner similar to the Diffusion Model. Interestingly, the VLM attention maps after alignment are more concentrated than those of the Diffusion Model. This difference is likely driven by their distinct optimization goals: sense understanding for optimal text generation in VLMs versus pixel-level precision for image generation in DMs.

\vspace{-2mm}

\begin{figure*}
    \centering
    \includegraphics[width=1\linewidth]{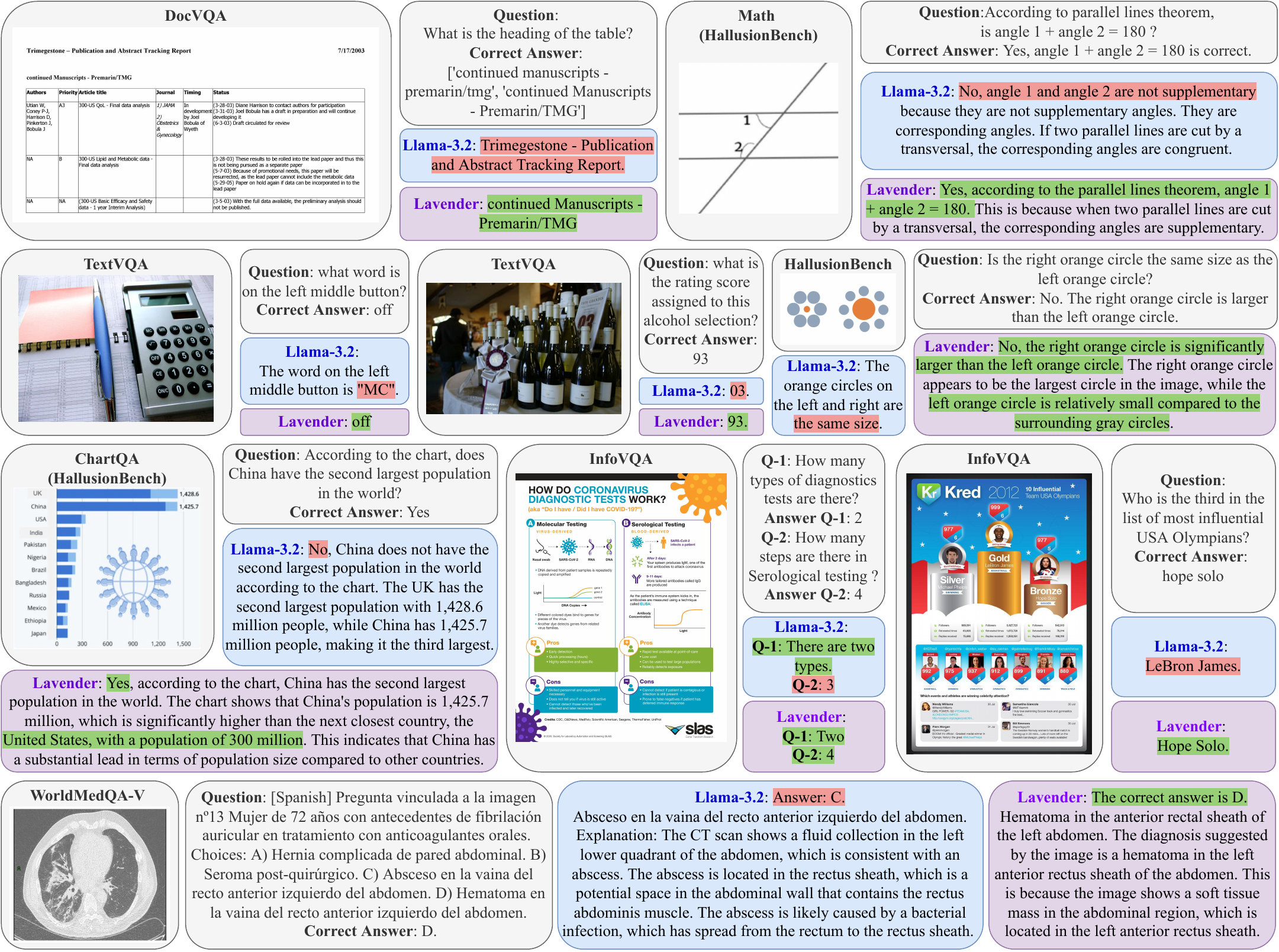}
    \caption{\textbf{Examples demonstrating Lavender's enhanced fine-granularity vision alignment leading to improved accuracy on various VQA benchmarks.}
    In the DocVQA example on the top-left, with a question about the title of a table, Llama-3.2 mistakenly extracts the section/page title at the top, which is also in bold, while Lavender correctly identifies the table title located next to the table.  
    Similarly, more accurate visual understanding is directly observed in the two TextVQA examples in the second row.  
    Impressively, the example from HallucinationBench demonstrates Lavender's deep visual understanding of geometry, size, and spatial location, leading to robust anti-hallucination behaviour compared to the original Llama-3.2.  
    In the ChartQA and two InfoVQA examples in the third row, Lavender exhibits its ability to recognise more detailed information, including relatively small elements within the graphs.  
    In the WorldMedQA-V example, which is out-of-distribution, both Llama-3.2 and Lavender understand the question posed in a small language (Spanish) and answer in English. However, Llama-3.2 fails to provide the correct answer due to less accurate visual localization of the unhealthy region. It incorrectly identifies the region as the `rectus sheath,' which surrounds the actual area of interest, leading to an incorrect response. In contrast, Lavender accurately locates the `soft tissue mass', resulting in the correct answer.
    }
    \label{fig:lavender_visual_highlight}
\end{figure*}

\paragraph{Visual Results on VQA Benchmarks.}  
\cref{fig:lavender_visual_highlight} showcases nine VQA examples from multiple evaluated benchmarks, including document, chart, and graphic understanding, real-world OCR, geometry math, visual hallucination, and medical VQA. The comparison highlights the performance of Llama-3.2-11B-Vision-Instruct~\cite{dubey2024llama} against its Lavender fine-tuned version.  
Overall, Lavender demonstrates significantly improved visual understanding compared to the original Llama-3.2 across various VQA tasks, including DocVQA, TextVQA, ChartQA, and WorldMedQA-V. These improvements stem from Lavender's ability to accurately localise and interpret visual elements, leading to better performance on questions requiring detailed visual reasoning. A detailed analysis of these examples is provided in the image caption of \cref{fig:lavender_visual_highlight}.

\vspace{-1mm}
\section{Ablation and Analysis}
\label{sec:anlation_and_analysis}

We conducted extensive ablation studies to assess the key components of Lavender and their impact on performance.  
Unless otherwise specified, \textit{all experiments in this section use `exact match' in VQA assessments without an LLM judge for cost efficiency}.  

Key findings include:  
1) \textit{Attention Aggregation}: Learned aggregation consistently outperforms manual methods like attention flow due to its adaptability and scalability (\cref{fig:ablation_agg_spider}, \cref{fig:ablation_agg_mean}).  
2) \textit{Pretraining the Aligner}: Pretraining significantly mitigates catastrophic forgetting, with longer pretraining benefiting more complex benchmarks (\cref{fig:ablate_pretrain}).  
3) \textit{Fine-tuning Strategy}: LoRA achieves better short-term results, while full fine-tuning excels on complex tasks, suggesting complementary benefits (\cref{fig:performance_minicpm_lm32}) (b).  
4) \textit{Layer Alignment}: Aligning all eight cross-attention layers in Llama-3.2-11B proves most effective, outperforming partial alignments (\cref{fig:ablate_attend_layers}).

\vspace{-1mm}
\subsection{Attention Aggregation Functions}  
\label{sec:abl_attn_aggregation}

We evaluate different attention aggregation strategies in VLMs, including simple averaging (`layer-mean'), maximisation (`layer-max' and 'max-max'), attention flow combined with multiplicative and additive aggregations (`flow-multi' and `flow-sum'), and learned aggregation (`learn'), as detailed in \cref{sec:aggregations}.  

\begin{figure}
    \centering
    \includegraphics[width=.9\linewidth]{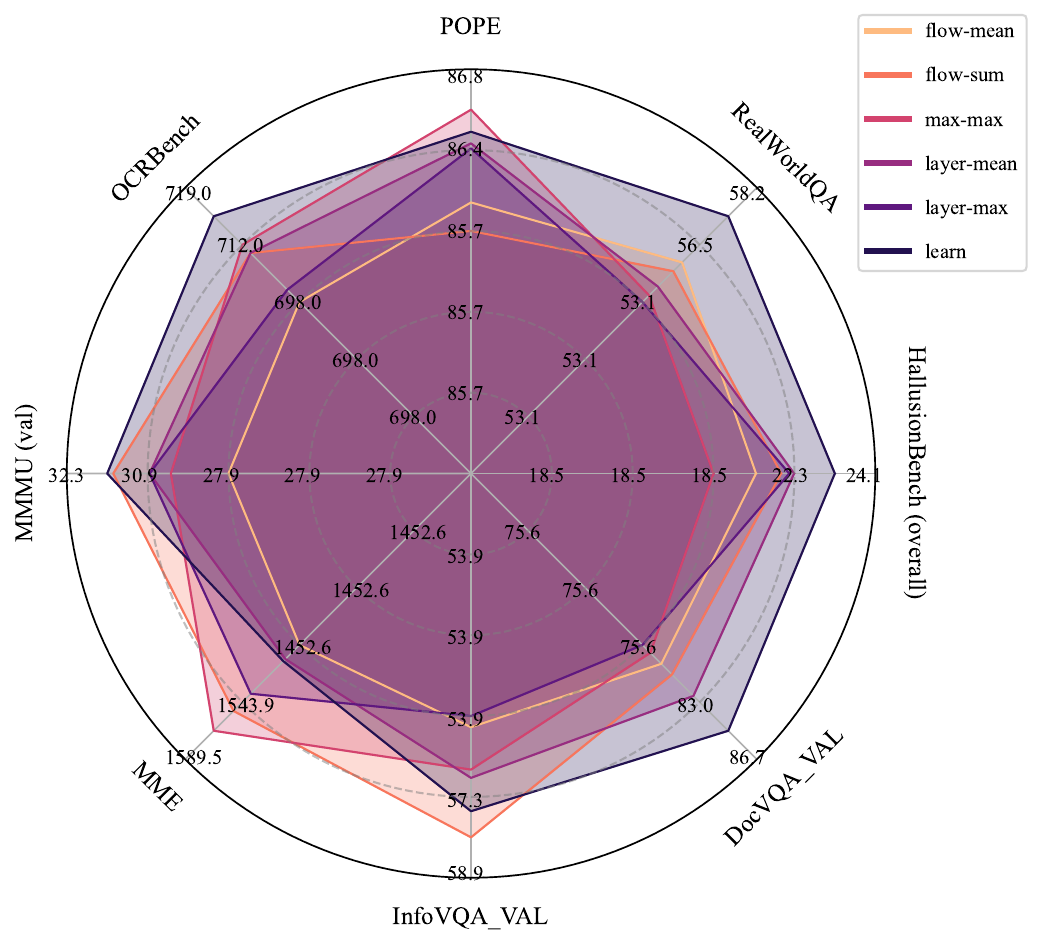}
    \caption{\textbf{Comparison of aggregation functions on eight benchmarks.} Evaluated using Lavender-Llama 3.2-11B fine-tuned on Flickr-1k for 20 epochs, using `exact match' without an LLM judge.}
    \label{fig:ablation_agg_spider}
\end{figure}

\vspace{-2mm}
\paragraph{Comparison Across Eight Benchmarks.}  
We fully fine-tune Lavender-Llama 3.2-11B on Flickr-1k for 20 epochs and evaluate all methods on eight benchmarks using `exact match' in VQA assessments. This differs from the main results (\cref{tab:main_table_full}) and leads to an expected 10\%--20\% performance drop.  
\cref{fig:ablation_agg_spider} shows that `flow-multi' performs the weakest, likely due to excessive compression, while `flow-sum' ranks second, benefiting from additive aggregation. `Layer-mean' and `layer-max' demonstrate robust mid-level performance, avoiding biases introduced by explicit aggregation functions.  
Among all methods, \emph{learned aggregation consistently outperforms others}, confirming its ability to preserve pretrained attention mechanisms while capturing complex semantic correlations.  

\begin{figure}
    \centering
    \includegraphics[width=1\linewidth]{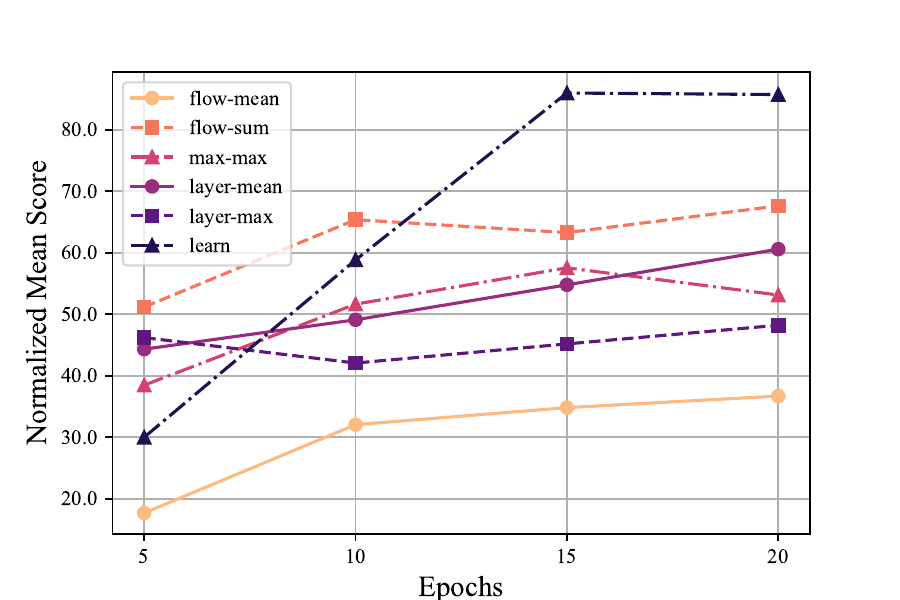}
    \caption{\textbf{Mean performance across eight benchmarks over training epochs with various aggregation functions.} The results are derived from Lavender-Llama 3.2-11B, fully fine-tuned on the Flickr-1k subset. \cref{fig:ablation_agg_full} provides detailed performance for each benchmark.}
    \label{fig:ablation_agg_mean}
\end{figure}

\paragraph{Scalability of Aggregation Functions.}  
We further assess the scalability of the proposed aggregation functions by comparing the eight benchmarks' performance at different training lengths. The average results are shown in \cref{fig:ablation_agg_mean}, with detailed results for each individual benchmark provided in \cref{fig:ablation_agg_full}.  
Despite the consistent overall ranking discussed earlier, the `learn' aggregation strategy demonstrates the best scalability. Starting as the second worst at epoch 5, it climbs to the best-performing method by epoch 20. This behaviour can be understood in light of the limitations of manually designed aggregations, such as `flow-sum.' While manual aggregations may initially reflect straightforward text-to-region correlations, like identifying colours in images, they are inherently suboptimal. These methods act as a shortcut, achieving good early performance but lacking scalability when encountering more complex text-to-region correlations. Learnable aggregation, in contrast, adapts to these complexities and scales effectively with additional training.

\subsubsection{Training recipes}
\label{sec:abl_train_recipes}

\begin{figure}
    \centering
    \includegraphics[width=1\linewidth]{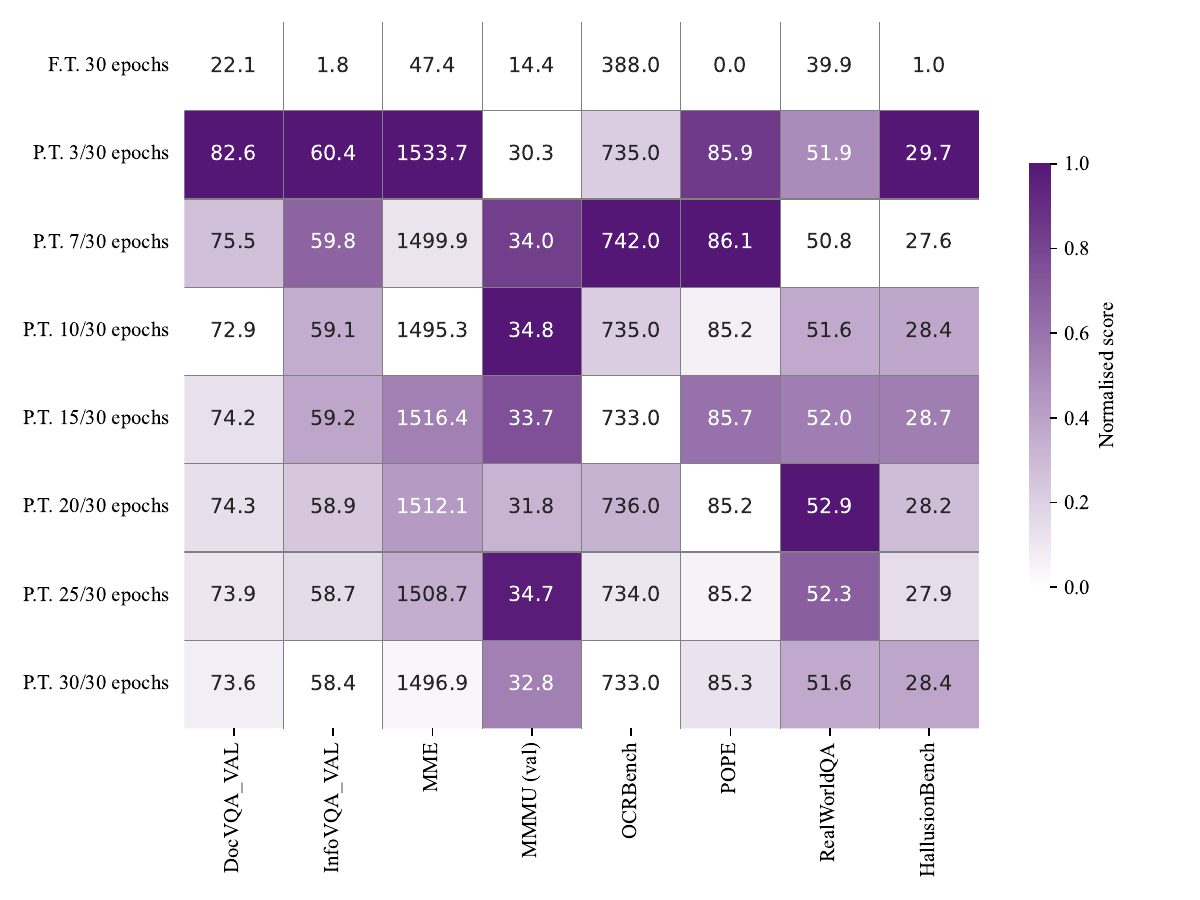}
    \vspace{-4mm}
    \caption{\textbf{Fully fine-tuning (F.T.) leads to catastrophic forgetting, while short Aligner pretraining (P.T.) prevents this.} Evaluated on eight benchmarks with Lavender-Llama 3.2-11B fine-tuned on Flickr-1k for 30 epochs.}
    \label{fig:ablate_pretrain}
\end{figure}

\paragraph{Pretraining the Aligner Network.}  
This work aims to enhance the alignment of pretrained VLMs, such as Llama-3.2, which are already knowledgeable. As discussed in \cref{sec:train_recipe}, fully fine-tuning such VLMs with a secondary attention alignment objective and a small dataset can lead to catastrophic forgetting.  
The results in the first row of \cref{fig:ablate_pretrain} confirm this concern, showing drastic drops in performance across eight evaluated benchmarks after fully fine-tuning on a small dataset for 30 epochs. In contrast, the proposed pretraining strategy significantly mitigates this issue.  
The pretraining strategy involves pretraining only the Aligner Network for a certain number of epochs while keeping the VLM parameters frozen, before jointly updating all parameters. Results in \cref{fig:ablate_pretrain} show that pretraining for a short duration (less than one-third of the total epochs) is particularly effective overall. Notably, more challenging benchmarks that require deeper interaction between visual perception, complex reasoning, and domain knowledge, such as MMMU and RealWorldQA, benefit more from longer pretraining.

\paragraph{Full Finetuning Versus LoRA.}  
Comparing full finetuning with the LoRA finetuning strategy for Lavender, our main results with Llama-3.2 in \cref{fig:performance_minicpm_lm32} (b) and \cref{tab:main_table_full} suggest that LoRA is generally more beneficial for the majority of benchmarks. However, we also observe that full finetuning outperforms LoRA with Lavender on a few more challenging benchmarks, such as MMMU, MMStar, and RealWorldQA, which require deeper interactions between visual perception, complex reasoning, and domain knowledge.  
This observation is consistent with the results from the ablation study in \cref{fig:ablate_pretrain}. Given that this work focuses on short finetuning overall, these findings suggest that \textit{LoRA finetuning with Lavender offers better short-term benefits}. In contrast, \textit{scaling Lavender with full finetuning may lead to deeper alignment and knowledge restructuring, potentially resulting in longer-term advantages}.

\begin{figure}
    \centering
    \includegraphics[width=1\linewidth]{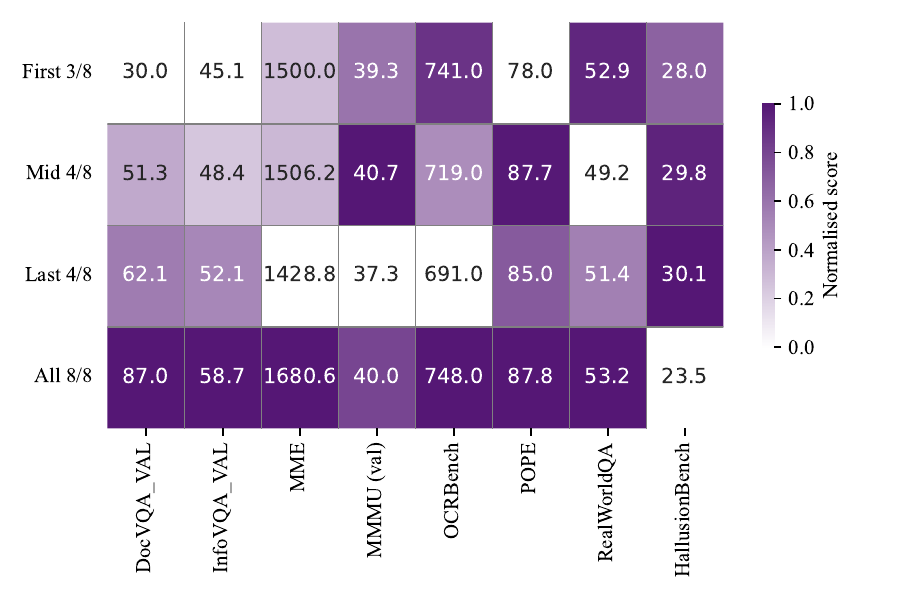}
    \vspace{-4mm}
    \caption{\textbf{Aligning all eight cross-attention layers in Llama-3.2 with Lavender is most effective.}  
    This ablation study examines the impact of aligning subsets of the eight cross-attention layers \{8, 13, 18, 23, 28, 33, 38\} in Llama-3.2, denoted as `First 3/8,' `Mid 4/8,' `Last 4/8,' and `All 8/8,' as shown in each row.  
    The results are derived from Lavender-Llama 3.2-11B, LoRA fine-tuned on the Flickr-1k subset for 30 epochs and evaluated on eight VLM benchmarks using `exact match' without an LLM judge.}
    \label{fig:ablate_attend_layers}
\end{figure}

\paragraph{Choice of Layers to Align.}  
We examine the choice of cross-attention layers to align with Lavender in the Llama 3.2-11B Vision Instruct model. This model comprises 40 layers in total, 8 of which are cross-attention layers integrated to process visual inputs, constituting one-fifth of the model's layers.  
\cref{fig:ablate_attend_layers} compares the performance of four different subsets of the 8 cross-attention layers. While attending to the first, mid, or last subset shows shifted strengths on specific benchmarks—RealWorldQA/OCRBench, MMMU/POPE, and Hallucination, respectively—aligning all 8 layers proves to be the most effective overall. Therefore, this is the default strategy adopted in this work.

\section{Failure Strategies}  
Despite the successful strategies presented earlier, we also tested the following approaches and found them to be less effective:  
1) \textit{Fully Finetuning Without Pretraining.}
As shown in \cref{fig:ablate_pretrain}, fully finetuning without pretraining often destabilises the model, while LoRA proves to be generally robust. We recommend starting with LoRA on small datasets as a reliable initial strategy for exploring Lavender, reserving full finetuning for scaling efforts and investigating emerging behaviours after deep alignment.  
2) \textit{Frequent Switching Between Training Strategies.}  
Given the varied performance of certain Lavender strategies, such as long fully finetuning or aligning subsets of cross-attention layers, we experimented with more complex staged training strategies. These involved switching between different strategies across epochs. However, frequent strategy changes harmed performance, likely due to the short overall training length. The model appeared to struggle with prioritizing objectives across strategies before stabilizing, leading to crashes.  
3) \textit{Mixing Extra Data.} 
In some cases, adding extra data did not improve performance. For example, as shown in \cref{fig:scaling_mean}, mixing the OCRVQA dataset reduced overall performance. This suggests that OCRVQA may lead to overfitting more readily than other datasets tested.

\section{Limitation and Future Works}

1) \emph{Limited Compute.} Lavender was evaluated on datasets up to 0.13M samples, constrained by available compute resources. This is far smaller than the 5M–50M samples used by state-of-the-art models (approximately 38x to 384x larger). \cref{fig:scaling_mean} shows non-convergent scaling behaviour, suggesting that increasing dataset size and tuning duration could further boost performance.
2) \emph{Exploring Higher-Resolution Diffusion Models.} In this work, Lavender uses Stable Diffusion v1.4 \cite{rombach2021highresolution}. More advanced models, such as Stable Diffusion v2 \cite{Rombach_2022_CVPR}, could deliver higher-resolution, more precise attention maps but would demand memory beyond the capacity of this study. With further scaling and tuning, such approaches could become viable, marking an important direction for future research.
3) \emph{More Accurate Attention Map Extraction.} This study focuses on demonstrating Lavender's feasibility rather than optimising the quality of diffusion model attention maps, which already exhibit strong text-region alignment. Short inversion and diffusion steps produced per-word attention maps in 20 seconds per image on a single V100 GPU. While efficient, this method may limit accuracy for infrequent words~\cite{jin2023image}. Future work could explore faster, more precise attention map estimation to further enhance performance.
4) \emph{Improved Handling of Self-Attention-Only Models.} We demonstrate that Lavender can be applied to self-attention-only models such as MiniCPM-v-2.5 (based on Llama-3). However, further optimisation for these models remains under-explored due to limited resources. In particular, the advanced parallel attention mechanism—effective in the Llama-3.2 implementation of Lavender—warrants additional theoretical and empirical study to better translate between self- and cross-attention, benefiting the broader community.

\section{Broad Impacts}

\paragraph{Synergy Between Models: Building a Stronger Vision-Language Ecosystem.}
Lavender is a model-agnostic framework that requires no additional human annotations on existing data, making it broadly applicable for enhancing a wide range of VLMs. Given that text–vision alignment is central to most VLM tasks, our attention alignment approach is poised to benefit most downstream applications, as confirmed by extensive benchmark evaluations. Although primarily designed for cross-attention-equipped VLMs, early results show that Lavender also works with self-attention-only models, opening the door to converting widely available LLMs into effective VLMs without retraining. This capability unites the strengths of LLM and diffusion models into a more powerful multimodal expert.

\paragraph{Data Scarcity.}
Both the language and vision communities face looming data shortages, with paired vision–text datasets—the "crude oil" of VLM training—being particularly scarce. End-to-end training from scratch is resource-intensive and often impractical. Large-scale LLMs and DMs have been trained on multi-billion-level datasets, making it inefficient if their knowledge remains isolated. Lavender bridges these models using limited resources—requiring as little as a few thousand samples and one day of training on 8 Nvidia A10G GPUs—enabling small models (below 13B parameters) to perform on par with much larger models (over 50B parameters) across multiple benchmarks.

\paragraph{Data Privacy.}
By leveraging the extensive prior knowledge of diffusion models, Lavender aligns external VLMs with small, local datasets through automatic per-text attention map labelling. This approach is ideal for organisations with limited sensitive data and constrained computational resources, as it enables them to benefit from large-scale pretrained models while keeping data local.

\paragraph{VLM Attention Alignment with Other Vision Foundation Models.}
Lavender enhances text–vision correlation by directly aligning attention within LLM transformer layers. Although our current alignment objectives are derived from Stable Diffusion's attention maps, the same methodology can be applied to other vision foundation models.

\paragraph{Alternative Multimodal Modalities Beyond Vision–Language.}
While our results focus on language and vision, the core idea of aligning cross-attention layers is broadly applicable. Potential applications include text-to-audio alignment \cite{radford2023robust}, sequence-to-structure tasks in protein generation \cite{jumper2021highly, yang2023alphafold2}, and diverse biomedical applications \cite{tu2024towards, yang2024advancing} involving genomics, radiography, pathology, and mammography.

\paragraph{Attention Alignment as Vision Feedback in Reinforcement Learning.}  
Our proposed attention alignment loss provides a scalable solution for vision feedback during the RL post-training phase. By leveraging the visual expertise encoded in image generation models, it eliminates the need for costly, fine-grained human annotations and mitigates bias in manual feedback. In effect, it replaces or augments traditional human visual feedback with "diffusion feedback", streamlining the process, reducing training costs, and democratising access to advanced multimodal systems.

\section{Conclusion}  
We have introduced Lavender, a method that leverages the precise text-region alignment of Diffusion Models to enhance Vision-Language Models (VLMs) through efficient supervised fine-tuning. Lavender enables significant performance improvements while remaining highly data-efficient and compute-friendly.
Our findings highlight that Lavender effectively aligns VLM attention with Diffusion Models, improving robustness across diverse domains, including challenging and multilingual benchmarks. By incorporating techniques such as parallel attention and LoRA, Lavender balances attention alignment with the preservation of pretrained knowledge, ensuring consistent performance gains without catastrophic forgetting.
This scalable, model-agnostic approach demonstrates the promising potential for advancing text-vision alignment in multimodal LLMs and lays the groundwork for future research into efficient fine-tuning and cross-modal alignment techniques.


\nocite{langley00}
\newpage
\bibliography{main}
\bibliographystyle{icml2024}

\newpage
\appendix
\onecolumn

\section*{Appendix Table of Contents} 
\addcontentsline{toc}{section}{Appendix Table of Contents} 
\label{sec:table_of_contents} 

\vskip 4mm
\hrule height .5pt
\vskip 4mm

\begin{itemize}[label={},leftmargin=*]
    \item \textbf{\textcolor{black}{\hyperref[sec:full_pseudo_code]{Appendix A - Full Pseudo Code of Diffusion Instruction Tuning}}} \dotfill \pageref{sec:full_pseudo_code}

    \item \textbf{\textcolor{black}{\hyperref[sec:dm_attn_proximity_app]{Appendix B - Bayesian Justification for DM Attention Proximity}}} \dotfill \pageref{sec:dm_attn_proximity_app}

    \item \textbf{\textcolor{black}{\hyperref[sec:justification_full]{Appendix C - Justification of the Attention Alignment Loss}}} \dotfill \pageref{sec:justification_full}

    \item \textbf{\textcolor{black}{\hyperref[sec:attention_flow]{Appendix D - Additional Details on Attention Flow in Vision-Language Models}}} \dotfill \pageref{sec:attention_flow}

    \item \textbf{\textcolor{black}{\hyperref[sec:reconstruction_matter]{Appendix E - Appropriate Rearrangement and Reconstruction Matter}}} \dotfill \pageref{sec:reconstruction_matter}
    \begin{itemize}[label={},leftmargin=*]
        \item \textit{\textcolor{black}{\hyperref[fig:attention_reconstruction_tiles]{\cref{fig:attention_reconstruction_tiles}: Reconstruction Tiles}}} \dotfill \pageref{fig:attention_reconstruction_tiles}
        \item \textit{\textcolor{black}{\hyperref[fig:reconstruction_debug]{\cref{fig:reconstruction_debug}: Reconstruction Debug}}} \dotfill \pageref{fig:reconstruction_debug}
    \end{itemize}

    \item \textbf{\textcolor{black}{\hyperref[sec:extra_visual]{Appendix F - Extended Supplementary Results}}} \dotfill \pageref{sec:extra_visual}
    \begin{itemize}[label={},leftmargin=*]
        \item \hyperref[sec:extra_results_ofm]{F.1. Extra Proof of Concept Results} \dotfill \pageref{sec:extra_results_ofm}
        \begin{itemize}[label={},leftmargin=*]
            \item \textit{\textcolor{black}{\hyperref[fig:attn_entropy_three_models]{\cref{fig:attn_entropy_three_models}: Entropy Histograms}}} \dotfill \pageref{fig:attn_entropy_three_models}
            \item \textit{\textcolor{black}{\hyperref[fig:visual_ofm_all]{\cref{fig:visual_ofm_all}: Visual Attention Verification}}} \dotfill \pageref{fig:visual_ofm_all}
        \end{itemize}
        \item \hyperref[sec:scaling_full]{F.2. Full Scaling Results} \dotfill \pageref{sec:scaling_full}
        \begin{itemize}[label={},leftmargin=*]
            \item \textit{\textcolor{black}{\hyperref[fig:scaling_mean]{\cref{fig:scaling_mean}: Scaling Mean Performance}}} \dotfill \pageref{fig:scaling_mean}
            \item \textit{\textcolor{black}{\hyperref[fig:scaling_full]{\cref{fig:scaling_full}: Scaling Across Benchmarks}}} \dotfill \pageref{fig:scaling_full}
        \end{itemize}
        \item \hyperref[sec:extra_ablation]{F.3. Extra Ablation Results} \dotfill \pageref{sec:extra_ablation}
        \begin{itemize}[label={},leftmargin=*]
            \item \textit{\textcolor{black}{\hyperref[fig:ablation_agg_full]{\cref{fig:ablation_agg_full}: Ablation Aggregation Functions}}} \dotfill \pageref{fig:ablation_agg_full}
        \end{itemize}
        \item \hyperref[sec:extra_visual_lm32]{F.4. Extra Visual Results} \dotfill \pageref{sec:extra_visual_lm32}
        \begin{itemize}[label={},leftmargin=*]
            \item \textit{\textcolor{black}{\hyperref[tab:world_med_qa_results]{\cref{tab:world_med_qa_results}: WorldMedQA-V Example Results}}} \dotfill \pageref{tab:world_med_qa_results}
            \item \textit{\textcolor{black}{\hyperref[tab:mme_failure]{\cref{tab:mme_failure}: Failure Case Analysis on the MME Benchmark}}} \dotfill \pageref{tab:mme_failure}
            \item \textit{\textcolor{black}{\hyperref[fig:attention_maps_full_fm]{\cref{fig:attention_maps_full_fm}: Visual Alignment Examples}}} \dotfill \pageref{fig:attention_maps_full_fm}
        \end{itemize}
    \end{itemize}

\end{itemize}

\vskip 4mm
\hrule height .5pt
\vskip 10mm

\newpage
\section{Full Pseudo Code of Diffusion Instruction Tuning \footnotesize{\textbf{\hyperref[sec:table_of_contents]{[Back to Contents]}}}}
\label{sec:full_pseudo_code}
This section expands on \cref{alg:algorithm_1}: \textit{Algorithm 1}, as introduced in the main paper.
\begin{algorithm}[H]
\caption{Diffusion Instruction Tuning (Full Version)}
\label{alg:diffusion_instruction_tuning_full}
\begin{algorithmic}[1]
\REQUIRE Dataset $D=\{(x^{(i)}, y^{(i)}, y_q^{(i)}, y_l^{(i)})\}_{i=1}^N$, where $y_q^{(i)}$ is textual input (e.g. a question) and $y_l^{(i)}$ is the target textual output. Pretrained DM parameters $\theta_D$, pretrained VLM parameters $\theta$. Scaling factor $\lambda>0$, learning rate $\eta$.
\ENSURE Fine-tuned VLM parameters $\theta$

\STATE \textbf{Stage 1: Preprocessing (Run Once)}
\STATE For each data point $(x^{(i)}, y^{(i)})$ in $D$:
\textcolor{gray}{\textit{\# the DM processes image-question pairs, hence replacing $y_q$ with $y$}}
\STATE \quad Use the pretrained DM ($\theta_D$ fixed) to compute $p_{\text{DM}}(a \mid x^{(i)}, y^{(i)}; \theta_D)$
\STATE \quad Store these attention distributions: $A_{\text{DM}}^{(i)} \leftarrow p_{\text{DM}}(a \mid x^{(i)}, y^{(i)}; \theta_D)$
\STATE After this, we have a DM-derived attention target for each data point, which will remain fixed during fine-tuning.

\vspace{1em}
\STATE \textbf{Stage 2: Fine-Tuning the VLM}
\STATE Initialise $\theta$ from a pretrained VLM.
\STATE Set a learning rate $\eta$ and define maximum training steps or convergence criteria.
\REPEAT
    \STATE Sample a mini-batch $B \subseteq D$ of size $m$
    \STATE $L_{\text{VLM}}(\theta) \leftarrow 0$, $L_{\text{att}}(\theta) \leftarrow 0$
    \FOR{$(x^{(j)}, y^{(j)}, y_q^{(j)}, y_l^{(j)})$ in $B$}
        \STATE Compute $p_{\text{VLM}}(a \mid x^{(j)}, y^{(j)}; \theta)$ by aggregating attention heads/layers of the VLM cross-/self-attention (see Section~\ref{sec:aggregations})
        \STATE $\delta^{(j)}(\theta) \leftarrow \text{Aligner}\bigl(p_{\text{VLM}}(a \mid x^{(j)}, y^{(j)}; \theta)\bigr)\; - A_{\text{DM}}^{(j)}$
        \STATE Compute task loss: $L_{\text{VLM}}(\theta) \;{+}{=}\; -\log p(y_l^{(j)} \mid x^{(j)}, y_q^{(j)}; \theta)$
        \STATE Compute attention alignment loss: $L_{\text{att}}(\theta) \;{+}{=}\; \|\delta^{(j)}(\theta)\|^2$
    \ENDFOR
    \STATE Form total loss: $L_{\text{total}}(\theta) = L_{\text{VLM}}(\theta) + \lambda L_{\text{att}}(\theta)$
    \STATE Update parameters: $\theta \leftarrow \theta - \eta \nabla_{\theta} L_{\text{total}}(\theta)$
    \STATE Optionally, apply LoRA or other optimization tricks if desired.
\UNTIL{convergence (e.g., validation metric plateaus or max steps reached)}

\STATE \textbf{Output:} The fine-tuned VLM parameters $\theta$.

\end{algorithmic}
\end{algorithm}

\section{Bayesian Justification for DM Attention Proximity \footnotesize{\textbf{\hyperref[sec:table_of_contents]{[Back to Contents]}}}}
\label{sec:dm_attn_proximity_app}
This section expands on \cref{sec:bayesian_main}: \textit{Bayesian Derivation}, as discussed in the main paper.
In this section, we provide a detailed Bayesian justification for our assumption that the DM's attention distribution \( p_{\text{DM}}(a \mid x, y; \theta_D) \) is closer to the optimal posterior of vision-centric attention \( p^*(a \mid x, y) \) than the VLM's attention distribution \( p_{\text{VLM}}(a \mid x, y; \theta) \).

\paragraph{Optimal Posterior Attention Distribution.}
We define the \emph{ideal} attention distribution as the one that minimizes the VLM’s next-token prediction loss:
\begin{equation}
p^*(a | x, y) = \arg\min_{a \in \mathcal{R}^d} \mathbb{E}_{(x, y) \sim \mathcal{D}} \mathcal{L}_{\text{VLM}}(x, y, \theta),
\end{equation}
where \( \mathcal{D} \) is the data distribution and \( \mathcal{L}_{\text{VLM}} \) is the loss function of the vision-language model.
We assume that the DM’s attention distribution is closer to this ideal attention than the VLM’s, and in the following sections, we analyze this claim using Bayesian modelling and statistical comparisons.

\subsection{Modelling Attention Distributions as Posteriors}

We model the attention distributions in both the DM and the VLM as posterior probabilities over the attention \( a \) given the inputs and model parameters. For consistency, we consider the joint distributions and apply Bayes' theorem.

\paragraph{Diffusion Model (DM):}

The DM is trained to generate an image \( x \) conditioned on text \( y \) by modelling the distribution \( p_{\text{DM}}(x \mid y; \theta_D) \). We can express the attention mechanism in the DM as contributing to this distribution via:

\begin{equation}
p_{\text{DM}}(x \mid y; \theta_D) = \int p_{\text{DM}}(x \mid y, a; \theta_D) \, p_{\text{DM}}(a \mid y; \theta_D) \, da.
\end{equation}

Applying Bayes' theorem, the posterior over attention \( a \) given \( x \) and \( y \) is:

\begin{equation}
p_{\text{DM}}(a \mid x, y; \theta_D) = \frac{p_{\text{DM}}(x \mid y, a; \theta_D) \, p_{\text{DM}}(a \mid y; \theta_D)}{p_{\text{DM}}(x \mid y; \theta_D)}.
\end{equation}

\paragraph{Vision-Language Model (VLM):}

Similarly, for the VLM, which generates textual output \( y_l \) given an image \( x \) and textual input \( y_q \), the attention mechanism influences the distribution \( p_{\text{VLM}}(y_l \mid x, y_q; \theta) \):

\begin{equation}
p_{\text{VLM}}(y_l \mid x, y_q; \theta) = \int p_{\text{VLM}}(y_l \mid x, y_q, a; \theta) \, p_{\text{VLM}}(a \mid x, y_q; \theta) \, da.
\end{equation}

The posterior over attention \( a \) is then:

\begin{equation}
p_{\text{VLM}}(a \mid x, y_q, y_l; \theta) = \frac{p_{\text{VLM}}(y_l \mid x, y_q, a; \theta) \, p_{\text{VLM}}(a \mid x, y_q; \theta)}{p_{\text{VLM}}(y_l \mid x, y_q; \theta)}.
\end{equation}

\subsection{Differences in Likelihood Functions}

The key distinction arises from the likelihood functions:

\begin{itemize}
    \item \textbf{DM Likelihood \( p_{\text{DM}}(x \mid y, a; \theta_D) \):} The DM must reconstruct the image \( x \) accurately, which requires precise alignment between textual tokens and visual features. The attention \( a \) plays a critical role in ensuring that each part of the text \( y \) correctly influences the corresponding visual content in \( x \).
    \item \textbf{VLM Likelihood \( p_{\text{VLM}}(y_l \mid x, y_q, a; \theta) \):} The VLM generates text \( y_l \) based on the image \( x \) and input text \( y_q \). While attention \( a \) aids in focusing on relevant visual regions, the text generation process can often rely on higher-level visual features and may not require as fine-grained vision-text alignment as the DM.
\end{itemize}

\subsection{Entropy and Concentration of Attention Distributions}

Due to the DM's need for precise image reconstruction, its attention distribution \( p_{\text{DM}}(a \mid x, y; \theta_D) \) is expected to be more concentrated around the \( p^*(a \mid x, y) \). This can be quantified by the entropy \( H \) of the attention distributions:

\begin{equation}
\label{eq:attention_entropy_app}
H\left( p_{\text{DM}}(a \mid x, y; \theta_D) \right) < H\left( p_{\text{VLM}}(a \mid x, y_q; \theta) \right).
\end{equation}

A lower entropy indicates that the DM's attention distribution is more peaked and thus closer to \( p^*(a \mid x, y) \), whereas the VLM's higher entropy reflects a more diffuse attention distribution. \textbf{This is consistent with our empirical observations in \cref{fig:empirical} and section \ref{sec:empirical_verification}}.

\subsection{KL Divergence to the Ideal Attention in Vision-Centric Tasks}

We can formalise the proximity to the optimal posterior attention distribution using the Kullback-Leibler (KL) divergence. For the DM conditioned on a unified text $y$, modeling $p(x | y; \theta_D)$, the KL divergence to the \( p^*(a \mid x, y) \) is:

\begin{equation}
\label{eq:kl_raw_dm}
D_{\text{KL}}\left( p_{\text{DM}}(a \mid x, y; \theta_D) \parallel p^*(a \mid x, y) \right) = \int p_{\text{DM}}(a \mid x, y; \theta_D) \log \frac{p_{\text{DM}}(a \mid x, y; \theta_D)}{p^*(a \mid x, y)} \, da,
\end{equation}

and similarly, for the VLM processes an image $x$, question $y_q$, and answer label $y_l$, modeling $p(y_l | x, y_q; \theta)$:

\begin{equation}
\label{eq:kl_raw_vlm}
D_{\text{KL}}\left( p_{\text{VLM}}(a \mid x, y_q; \theta) \parallel p^*(a \mid x, y) \right) = \int p_{\text{VLM}}(a \mid x, y_q; \theta) \log \frac{p_{\text{VLM}}(a \mid x, y_q; \theta)}{p^*(a \mid x, y)} \, da.
\end{equation}

Bringing in the entropy \( H \) of the attention distributions, equations \eqref{eq:kl_raw_dm} and \eqref{eq:kl_raw_vlm} can be written as:

\begin{equation}
D_{\text{KL}}\left( p_{\text{DM}}(a \mid x, y; \theta_D) \parallel p^*(a \mid x, y) \right) = -H\left( p_{\text{DM}}(a \mid x, y; \theta_D) \right) - \int p_{\text{DM}}(a \mid x, y; \theta_D) \log p^*(a \mid x, y) \, da,
\end{equation}

and similarly, for the VLM:

\begin{equation}
D_{\text{KL}}\left( p_{\text{VLM}}(a \mid x, y; \theta) \parallel p^*(a \mid x, y) \right) = -H\left( p_{\text{VLM}}(a \mid x, y; \theta) \right) - \int p_{\text{VLM}}(a \mid x, y; \theta) \log p^*(a \mid x, y) \, da.
\end{equation}

Our empirical observation of lower entropy in equation \eqref{eq:attention_entropy_app} indicates that the DM's attention distribution is more concentrated and thus more certain about where to attend, which is a consequence of the DM's need for precise vision-text alignment during image reconstruction.

Assuming that the cross-entropy terms \( \int p_{\text{DM}}(a \mid x, y; \theta_D) \log p^*(a \mid x, y) \, da \) and \( \int p_{\text{VLM}}(a \mid x, y; \theta) \log p^*(a \mid x, y) \, da \) are \textit{approximately equal}\footnote{Since \( p^*(a \mid x, y) \) is the same for both models, and both \( p_{\text{DM}} \) and \( p_{\text{VLM}} \) are centered around \( p^*(a \mid x, y) \) given they are from pretrained models based on big dataset, the values of these cross-entropy terms would not differ significantly.} or that the difference in entropies dominates the difference in cross-entropies, we can infer:

\begin{equation}
D_{\text{KL}}\left( p_{\text{DM}}(a \mid x, y; \theta_D) \parallel p^*(a \mid x, y) \right) < D_{\text{KL}}\left( p_{\text{VLM}}(a \mid x, y; \theta) \parallel p^*(a \mid x, y) \right).
\end{equation}

This inequality suggests that the DM's attention distribution is closer to the optimal posterior attention distribution in terms of KL divergence. The lower entropy of the DM's attention implies it is more peaked around the ideal attention in vision-centric tasks in vision-centric tasks.
Therefore, our empirical observation of the DM's lower attention entropy supports the assumption that the DM's attention distribution is closer to the optimal posterior attention distribution than the VLM's. By aligning the VLM's attention with the DM's attention, we aim to reduce the VLM's KL divergence to the ideal attention in vision-centric tasks, enhancing its vision-text alignment and overall performance.

\section{Justification of the Attention Alignment Loss \footnotesize{\textbf{\hyperref[sec:table_of_contents]{[Back to Contents]}}}}
\label{sec:justification_full}
This section expands on \cref{eq:total_loss}, as introduced in the main paper, to provide a detailed justification for the inclusion of the attention alignment loss \( L_{\text{att}}(\theta) \) and the scaling factor \( \lambda \) in our Bayesian framework.

\subsection{Derivation of the Likelihood Term}

We start by modelling the difference between the VLM's attention distribution and the DM's attention distribution as a random variable. For each data point \( i \), we define:
\begin{equation}
\delta^{(i)}(\theta) = p_{\text{VLM}}(a \mid x^{(i)}, y^{(i)}; \theta) - p_{\text{DM}}(a \mid x^{(i)}, y^{(i)}; \theta_D).
\end{equation}

We assume that \( \delta^{(i)}(\theta) \) follows a multivariate normal distribution with zero mean and covariance matrix \( \sigma^2 I \), where \( I \) is the identity matrix:
\begin{equation}
\delta^{(i)}(\theta) \sim \mathcal{N}(0, \sigma^2 I).
\end{equation}

Under this assumption, the probability density function for \( \delta^{(i)}(\theta) \) is:
\begin{equation}
p\left( \delta^{(i)}(\theta) \right) = \frac{1}{(2\pi \sigma^2)^{k/2}} \exp\left( -\frac{1}{2\sigma^2} \left\| \delta^{(i)}(\theta) \right\|^2 \right),
\end{equation}
where \( k \) is the dimensionality of the attention distribution.

The likelihood of observing the DM's attention given the VLM's parameters over the entire dataset is then:
\begin{equation}
p(A_{\text{DM}} \mid \theta) = \prod_{i} p\left( \delta^{(i)}(\theta) \right) = \left( \frac{1}{(2\pi \sigma^2)^{k/2}} \right)^N \exp\left( -\frac{1}{2\sigma^2} \sum_{i} \left\| \delta^{(i)}(\theta) \right\|^2 \right),
\end{equation}
where \( N \) is the number of data points.

Taking the negative log-likelihood, we get:
\begin{equation}
-\log p(A_{\text{DM}} \mid \theta) = \frac{Nk}{2} \log(2\pi \sigma^2) + \frac{1}{2\sigma^2} \sum_{i} \left\| \delta^{(i)}(\theta) \right\|^2.
\end{equation}

Ignoring constants that do not depend on \( \theta \), we have:
\begin{equation}
-\log p(A_{\text{DM}} \mid \theta) = \frac{1}{2\sigma^2} \sum_{i} \left\| \delta^{(i)}(\theta) \right\|^2 + \text{const}.
\end{equation}

Defining \( \lambda = \frac{1}{\sigma^2} \), we arrive at:
\begin{equation}
-\log p(A_{\text{DM}} \mid \theta) = \frac{\lambda}{2} \sum_{i} \left\| \delta^{(i)}(\theta) \right\|^2 + \text{const}.
\end{equation}

\subsection{Interpretation of the Scaling Factor \(\lambda\)}

The scaling factor \( \lambda \) plays a crucial role in balancing the attention alignment loss with the primary task loss. It is inversely proportional to the variance \( \sigma^2 \) of the assumed Gaussian distribution of the attention differences.

\begin{itemize}
    \item A small \( \sigma^2 \) (large \( \lambda \)) implies high confidence in the DM's attention distributions, placing more emphasis on aligning the VLM's attention with that of the DM.
    \item A large \( \sigma^2 \) (small \( \lambda \)) implies less confidence in the DM's attention distributions, reducing the influence of the attention alignment loss.
\end{itemize}

In practice, \( \lambda \) can be treated as a hyperparameter tuned based on validation performance.

\subsection{Total Negative Log-Posterior}

The total negative log-posterior combines the negative log-likelihoods of the data and the attention alignment:
\begin{equation}
L_{\text{total}}(\theta) = -\log p(D \mid \theta) -\log p(A_{\text{DM}} \mid \theta) + \text{const}.
\end{equation}

Substituting the expressions for the negative log-likelihoods, we have:
\begin{equation}
L_{\text{total}}(\theta) = L_{\text{VLM}}(\theta) + \lambda L_{\text{att}}(\theta) + \text{const},
\end{equation}
where:
\begin{align}
L_{\text{VLM}}(\theta) &= -\sum_{i} \log p(y_l^{(i)} \mid x^{(i)}, y_q^{(i)}; \theta), \\
L_{\text{att}}(\theta) &= \frac{1}{2} \sum_{i} \left\| \delta^{(i)}(\theta) \right\|^2.
\end{align}

By minimizing \( L_{\text{total}}(\theta) \), we maximise the posterior probability \( p(\theta \mid D, A_{\text{DM}}) \), effectively incorporating both the data likelihood and the prior information provided by the DM's attention distributions.

\subsection{Justification for Using MSE Loss}

The mean squared error (MSE) loss used in \( L_{\text{att}}(\theta) \) arises naturally from the assumption of Gaussian-distributed attention differences. This is a common assumption in Bayesian modelling, where the Gaussian distribution is often used due to its mathematical convenience and the central limit theorem.
The MSE loss is also computationally efficient and widely used in neural network training, making it a practical choice for aligning attention distributions.

\section{Additional Details on Attention Flow in Vision-Language Models \footnotesize{\textbf{\hyperref[sec:table_of_contents]{[Back to Contents]}}}}
\label{sec:attention_flow}
This section expands on \cref{sec:attn_flow}: \textit{Attention Flow}, as introduced in the main paper.
In addition to simple aggregation methods, we explore \textit{\textbf{attention flow}}~\citep{abnar2020quantifying} to aggregate attention maps across layers in VLMs. Attention flow computes the effective attention between input and output tokens by considering the cumulative effect of attention across layers, capturing deeper interactions that span multiple layers. This method has been utilised by \citet{lin2024training} to obtain sentence-level aggregated attention maps for grounded segmentation tasks. We investigate its applicability in aggregating word-level attention maps in VLMs, aiming to capture semantic correlations that may not be evident through simple aggregation methods.

Let \( A^{(l)} \in \mathbb{R}^{N_{\text{text}} \times N_{\text{patch}}} \) denote the attention matrix at layer \( l \), where \( N_{\text{text}} \) is the number of text tokens and \( N_{\text{patch}} \) is the number of image patches. Our goal is to compute an aggregated attention map \( \bar{A} \in \mathbb{R}^{N_{\text{text}} \times N_{\text{patch}}} \) that captures the overall attention from text tokens to image patches across all layers.

We initialise the aggregated attention map (mean or max) with the attention from the first layer:
\begin{equation}
    \bar{A} = A^{(1)}.
\end{equation}
We then recursively update \( \bar{A} \) by combining it with the attention matrices from subsequent layers. Specifically, for each layer \( l = 2, \dots, L \), we update \( \bar{A} \) using either element-wise multiplication:
\begin{equation}
    \bar{A} = \bar{A} \circ A^{(l)},
\end{equation}
or element-wise addition:
\begin{equation}
    \bar{A} = \bar{A} + A^{(l)},
\end{equation}
where \( \circ \) denotes element-wise multiplication. We explore both strategies—multiplicative and additive aggregation—to assess which better captures the semantic correlations.

However, directly applying attention flow in autoregressive VLMs can lead to attention collapse due to the causal masks used during training. To mitigate this issue, we introduce a regularisation term that adjusts the contribution of each text token. Specifically, we define a regularisation vector \( r \in \mathbb{R}^{N_{\text{text}}} \) with elements:
\begin{equation}
    r_t = \frac{t}{N_{\text{text}}}, \quad t = 1, \dots, N_{\text{text}}.
\end{equation}
This term assigns lower weights to earlier tokens and higher weights to later tokens, preventing the dominance of early tokens in the aggregated attention map.

We apply the regularisation to the aggregated attention map:
\begin{equation}
    \bar{A}_{t,p} = \bar{A}_{t,p} \times r_t,
\end{equation}
where \( \bar{A}_{t,p} \) represents the attention from text token \( T_t \) to image patch \( T_p \). By incorporating this regularisation, we ensure that the attention flow effectively captures semantic correlations without collapsing due to the model's autoregressive nature.

Through attention flow with regularisation, we aggregate attention across layers to obtain per-word attention maps that better reflect the semantic relationships between text tokens and image patches. This method captures deeper interactions that may not be evident through simple aggregation, enhancing the alignment between VLMs and DMs.

\section{Appropriate Rearrangement and Reconstruction Matter \footnotesize{\textbf{\hyperref[sec:table_of_contents]{[Back to Contents]}}}}  
\label{sec:reconstruction_matter}  
This section extends the prior discussion in \cref{sec:aggregation_vlm} around rearrangement and reconstruction during aggregation.
Many VLM preprocessing pipelines split an image into tiles before the projection process. In such cases, it is crucial to account for both the original tiling and the resulting patch order. Direct reshaping of the flattened tokens, without reassembling the original tile layout, disregards spatial continuity between adjacent tiles, potentially disrupting semantic alignment.  

\begin{figure}
    \centering
    \includegraphics[width=.8\linewidth]{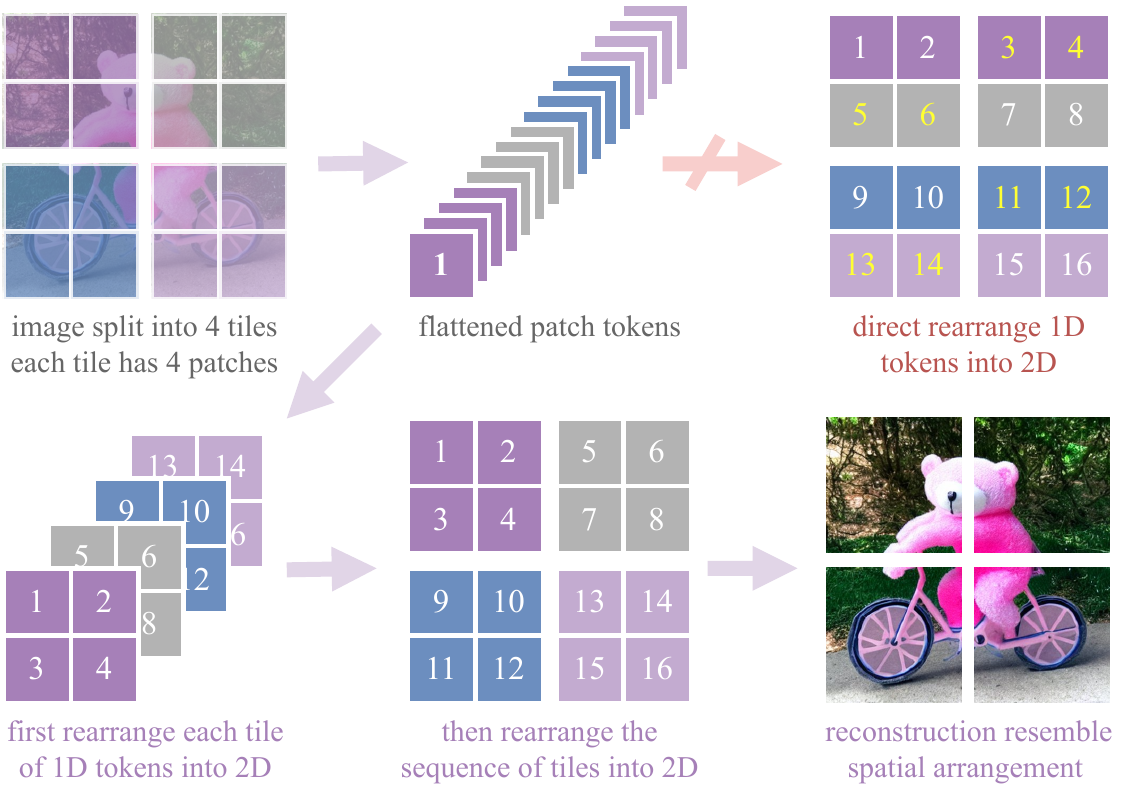}
    \caption{\textbf{Attention reconstruction under the tiling and tokenization procedure in Llama-3.2}, Example highlighting the importance of proper reconstruction. Improper rearrangement disrupts spatial continuity, while correct reconstruction preserves semantic alignment. \footnotesize{\textbf{\hyperref[sec:table_of_contents]{[Back to Contents]}}}}
    \label{fig:attention_reconstruction_tiles}
\end{figure}

\begin{figure}
    \centering
    \begin{minipage}{0.48\linewidth}
        \centering
        \includegraphics[width=\linewidth]{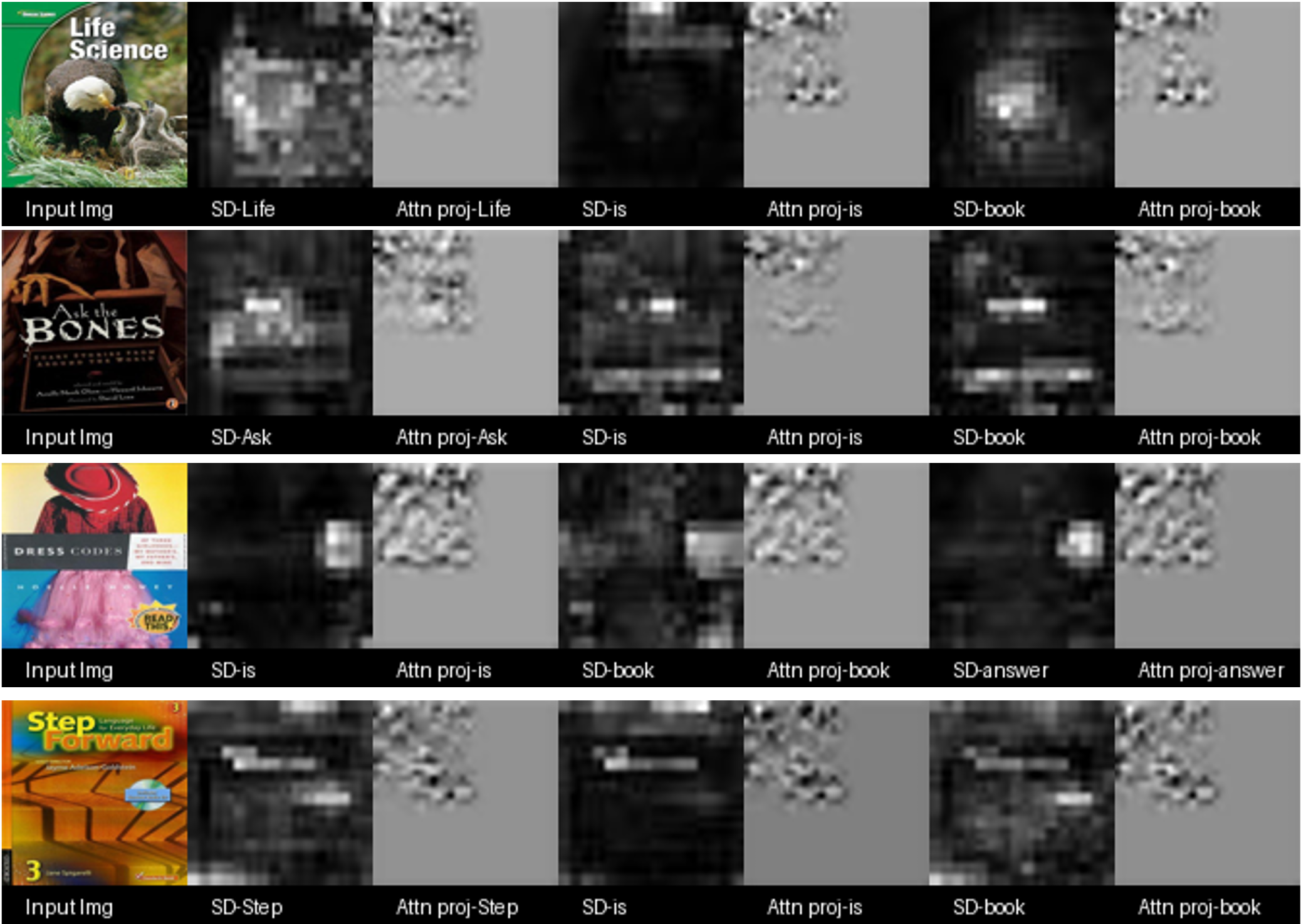}
        \textbf{(a) Before Debugging}
    \end{minipage}
    \hfill
    \begin{minipage}{0.48\linewidth}
        \centering
        \includegraphics[width=\linewidth]{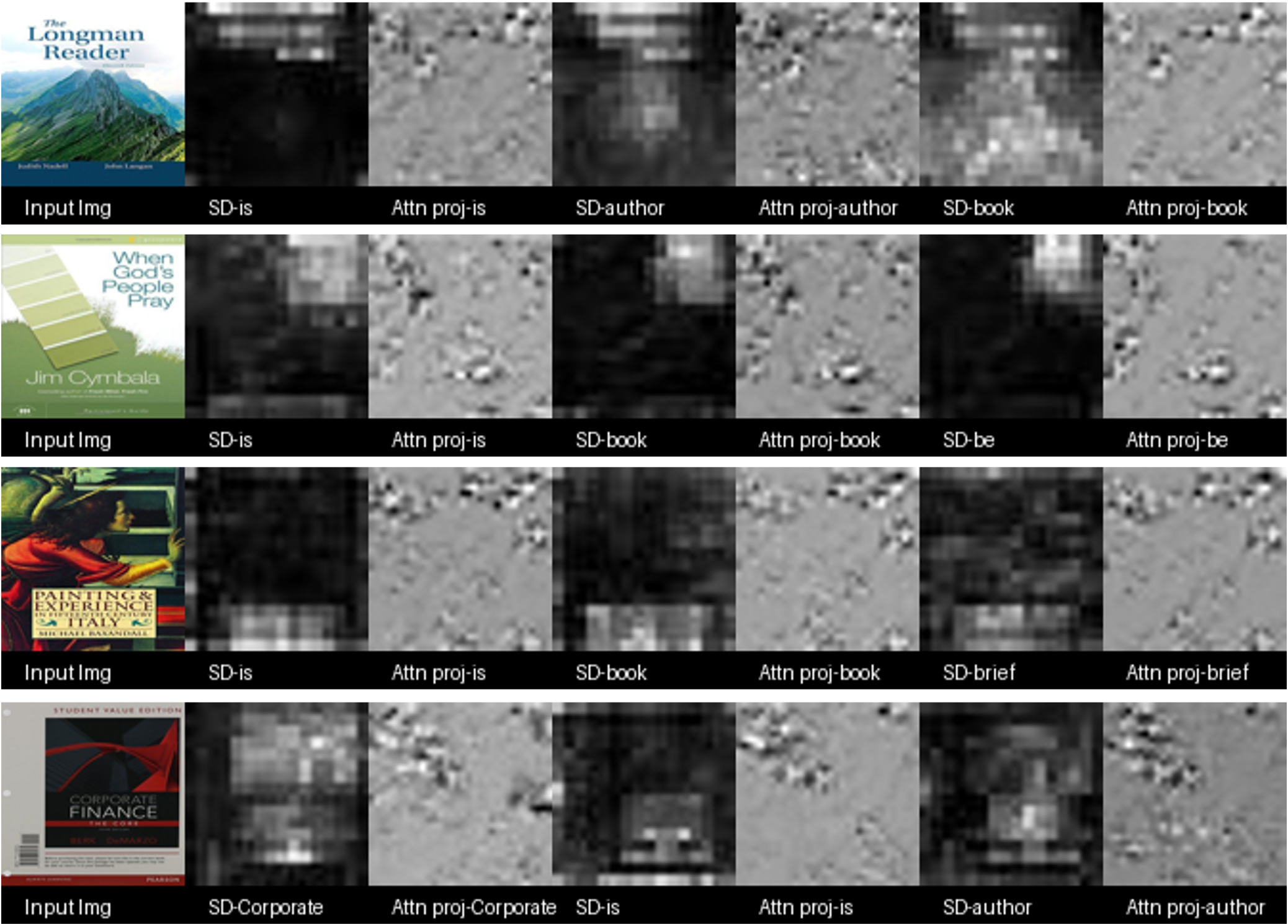}
        \textbf{(b) After Debugging}
    \end{minipage}
    \caption{\textbf{Appropriate rearrangement and reconstruction are crucial.} Results are based on attention maps extracted from Llama-3.2 (without Lavender fine-tuning) on OCRVQA samples. Poor rearrangement disrupts semantic alignment, while proper reconstruction corrects the spatial arrangement of the attention maps. \footnotesize{\textbf{\hyperref[sec:table_of_contents]{[Back to Contents]}}}}
    \label{fig:reconstruction_debug}
\end{figure}

\cref{fig:attention_reconstruction_tiles} illustrates the tiling and tokenization procedure in Llama-3.2 and highlights the importance of proper reconstruction. Similarly, \cref{fig:reconstruction_debug} demonstrates the impact of appropriate reconstruction on real samples from the OCRVQA dataset. Improper rearrangement leads to misaligned attention maps, while correct reconstruction restores semantic coherence, enhancing the quality of visual-textual alignment.

\section{Extended Supplementary Results  \footnotesize{\textbf{\hyperref[sec:table_of_contents]{[Back to Contents]}}}}  
\label{sec:extra_visual}  
Due to the large variety of datasets and experiments considered in this work, the main body focuses on summarizing and analyzing overall results. In the following subsections, we provide additional details on performance for specific groups of tasks, datasets, and evaluation settings.

\subsection{Extra Proof of Concept Results \footnotesize{\textbf{\hyperref[sec:table_of_contents]{[Back to Contents]}}}}  
\label{sec:extra_results_ofm}  
This section expands on \cref{sec:empirical_verification}: Empirical Verifications, as introduced in the main paper.
We begin by presenting additional detailed results extending the empirical verification experiments in \cref{sec:empirical_verification}, where we tested our key hypothesis: \textit{the cross-attention from DM transformers closely approximates an ideal attention mechanism for maximising VLM performance}.  

\begin{figure}
    \centering
    \begin{minipage}[t]{0.33\textwidth}
        \centering
        \includegraphics[width=\textwidth]{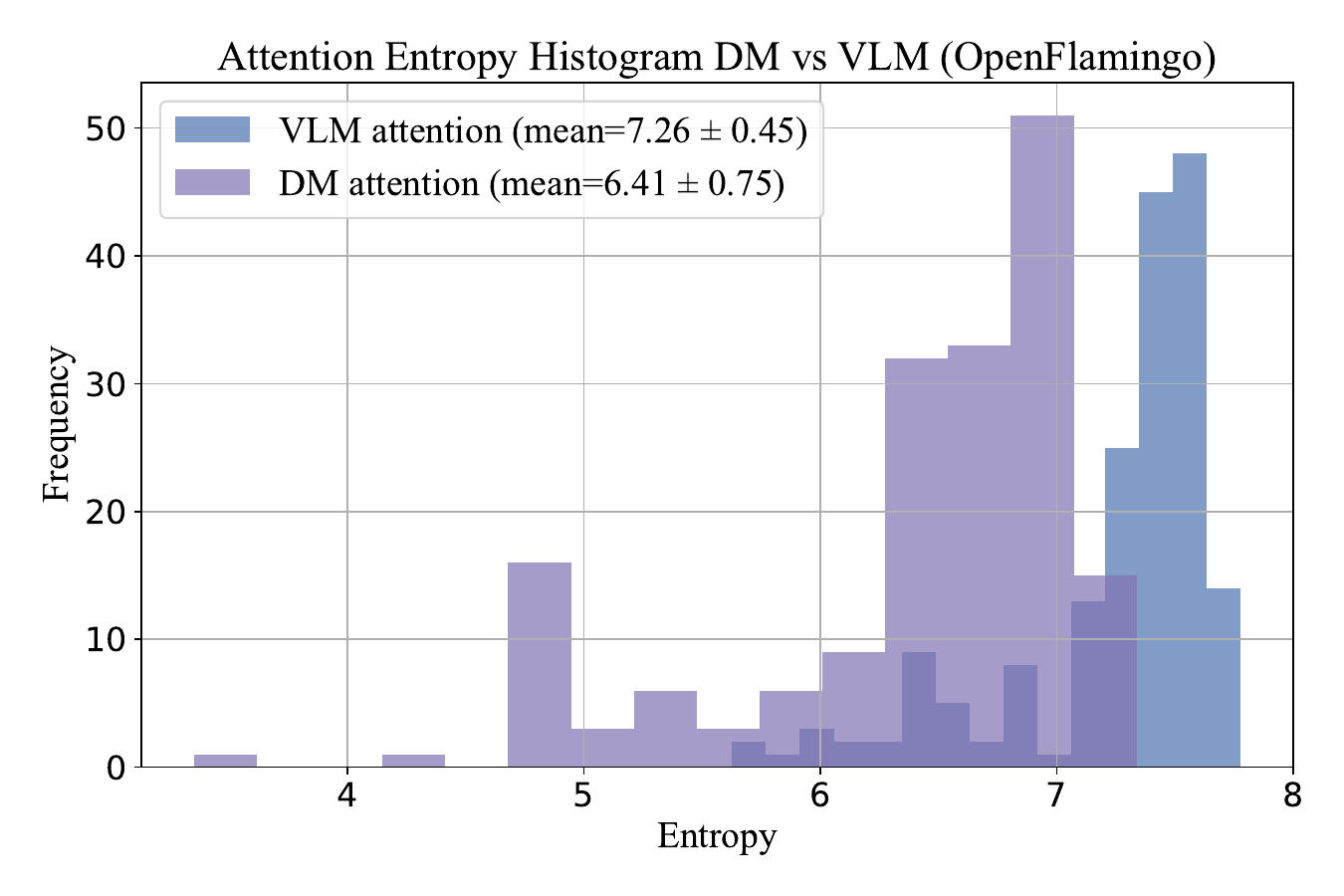}
    \end{minipage}%
    \hfill
    \begin{minipage}[t]{0.33\textwidth}
        \centering
        \includegraphics[width=\textwidth]{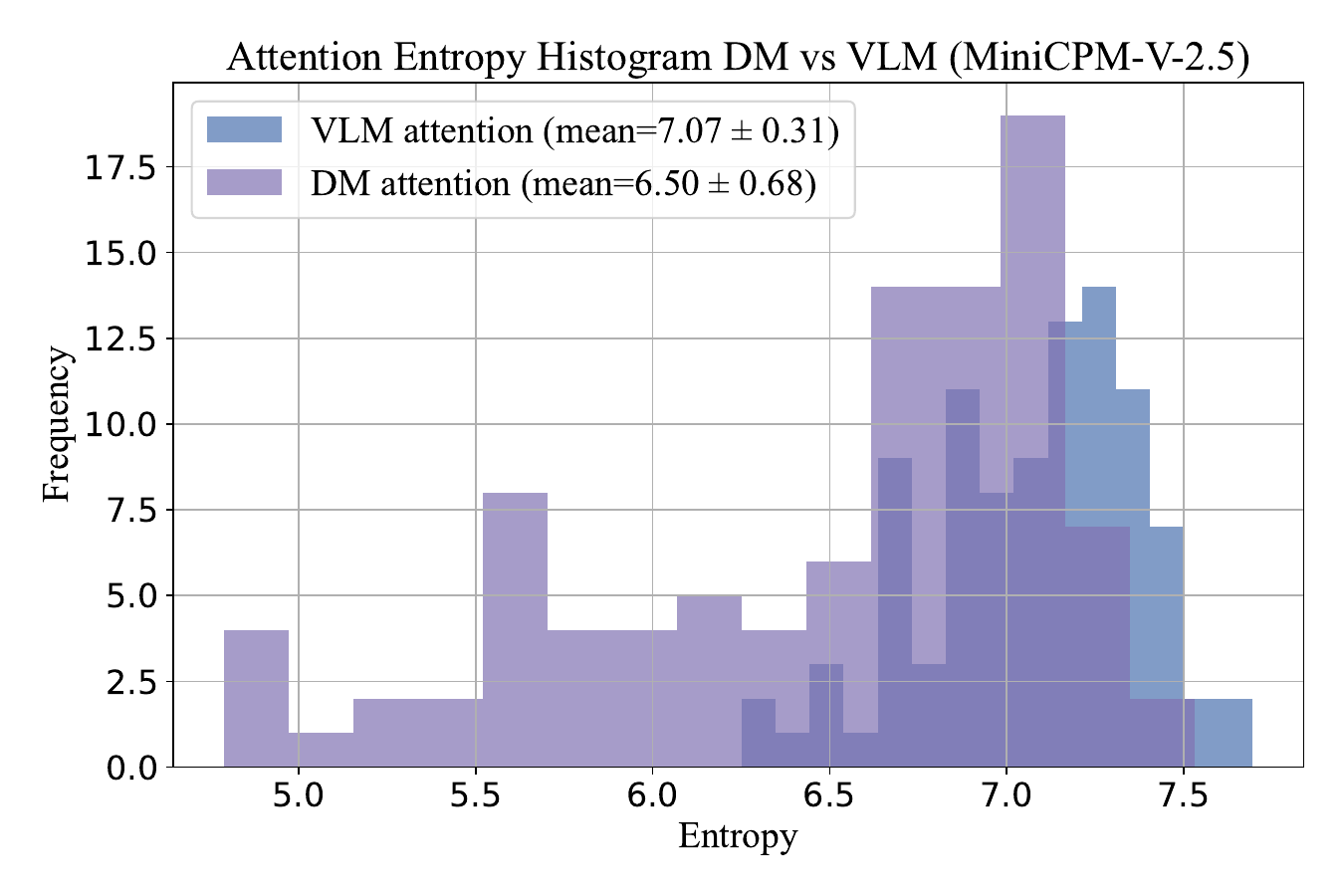}
    \end{minipage}
    \hfill
    \begin{minipage}[t]{0.33\textwidth}
        \centering
        \includegraphics[width=\textwidth]{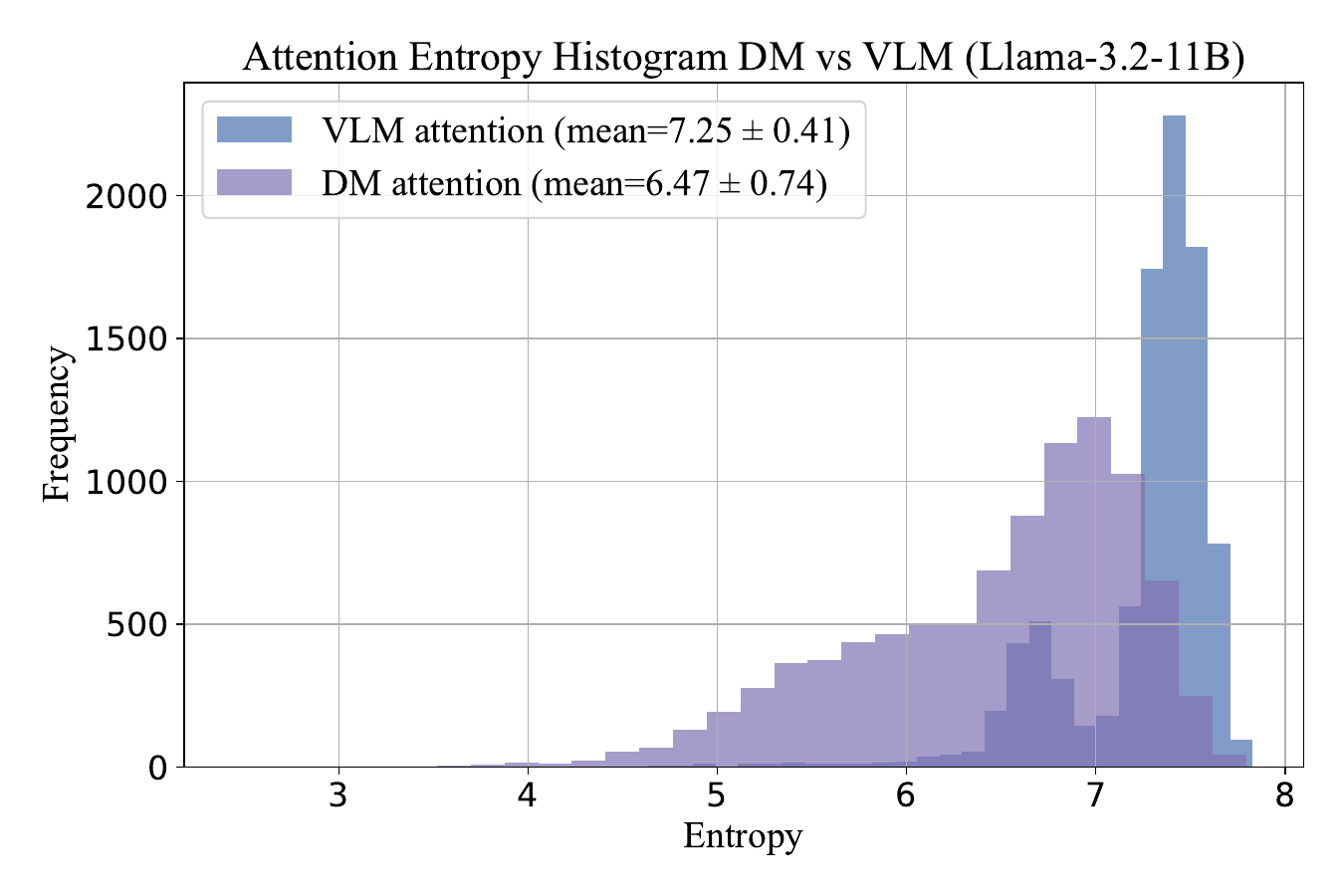}
    \end{minipage}%
    \caption{Attention map entropy histograms from three models (OpenFlamingo, MiniCPM-v2.5, and Llama 3.2-11B) are generated by processing a small subset of Flickr30k, RLAIF-V83k, and OCRVQA30k, totalling approximately 10k samples. \footnotesize{\textbf{\hyperref[sec:table_of_contents]{[Back to Contents]}}}}
    \label{fig:attn_entropy_three_models}
\end{figure}

\paragraph{Attention Entropy Histograms.}  
\cref{fig:attn_entropy_three_models} shows separate entropy histograms of the attention maps $A_{DM}$ and $A_{VLM}$ for three models: OpenFlamingo, MiniCPM-v2.5, and Llama 3.2-11B. These results extend the combined plot presented in \cref{fig:entropy_histogram_combined}.  
We observe that the DM's attention distribution, $p_{\text{DM}}(a | x, y; \theta_D)$, consistently exhibits lower entropy compared to the VLM's attention distribution, $p_{\text{VLM}}(a | x, y; \theta)$ across all three models. This finding reinforces our hypothesis that the DM's attention is more concentrated and thus closer to the optimal posterior attention distribution, $p^*(a | x, y)$.

\begin{figure}
    \centering
    \includegraphics[width=1\linewidth]{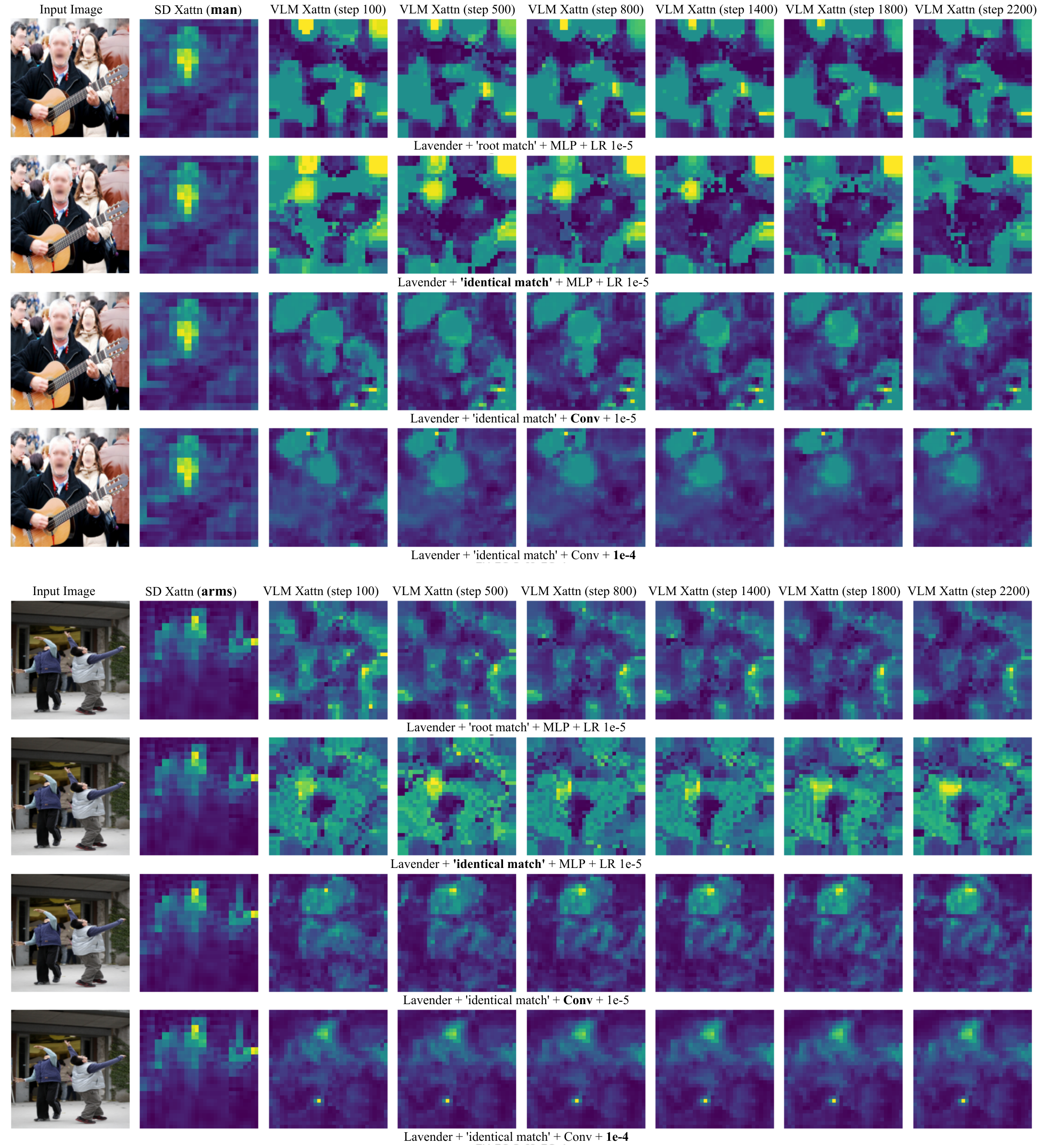}
    \caption{More visual verification of learning VLM attention aggregation compared to SD attention for the matched word '\textbf{man}' and '\textbf{arms}', based on the OpenFlamingo implementation of our method. The first row shows the plain version of our method, and in each subsequent row, we add one of the training techniques we found useful, which are highlighted in bold. \footnotesize{\textbf{\hyperref[sec:table_of_contents]{[Back to Contents]}}}}
    \label{fig:visual_ofm_all}
\end{figure}

\paragraph{Visual Confirmation of Attention Alignment.}  
\cref{fig:visual_ofm_all} provides additional visual evidence verifying that the proposed Lavender fine-tuning approach aligns VLM attention with DM attention as guidance, extending the results from \cref{fig:visual_guitar}.  
These results reaffirm our findings from the main body: VLM cross-attention maps successfully align with semantically meaningful DM attention patterns by leveraging strategies such as `exact word match' for sampling, convolutional layers in the Aligner network, and a learning rate of $1\text{e}{-4}$.

\subsection{Full Scaling Results \footnotesize{\textbf{\hyperref[sec:table_of_contents]{[Back to Contents]}}}}
\label{sec:scaling_full}
\begin{figure}
    \centering
    \includegraphics[width=.9\linewidth]{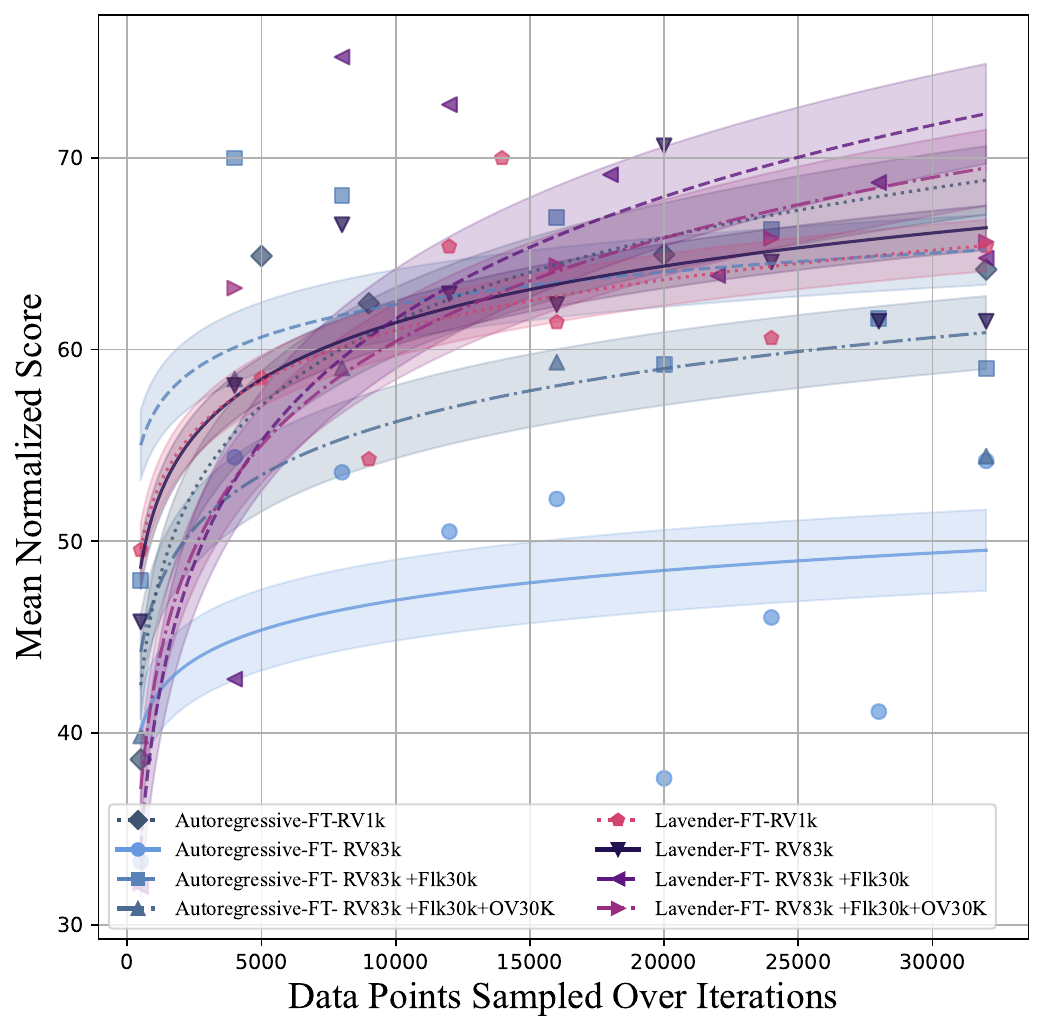}
    \caption{\textbf{Lavender scales better and reduces overfitting compared to autoregressive fine-tuning, with larger datasets lowering variations.} The plot shows the mean normalised performance of four dataset configurations across eight benchmarks after LoRA and Full fine-tuning of Llama 3.2-11B, as a function of the number of data points sampled. Markers represent observed performance for each method-iteration pair, while trendlines with different styles indicate the overall performance trends. The shaded regions around the trendlines represent confidence intervals derived from the standard error, showing the uncertainty of the trendline predictions. Narrower regions indicate higher confidence, while wider regions suggest greater variability. Per-benchmark results are in \cref{fig:scaling_full}. The simplified version is presented in \cref{fig:scaling_mean_sim}. \footnotesize{\textbf{\hyperref[sec:table_of_contents]{[Back to Contents]}}}}
    \label{fig:scaling_mean}
\end{figure}

\begin{figure}
    \centering
    \includegraphics[width=.95\linewidth]{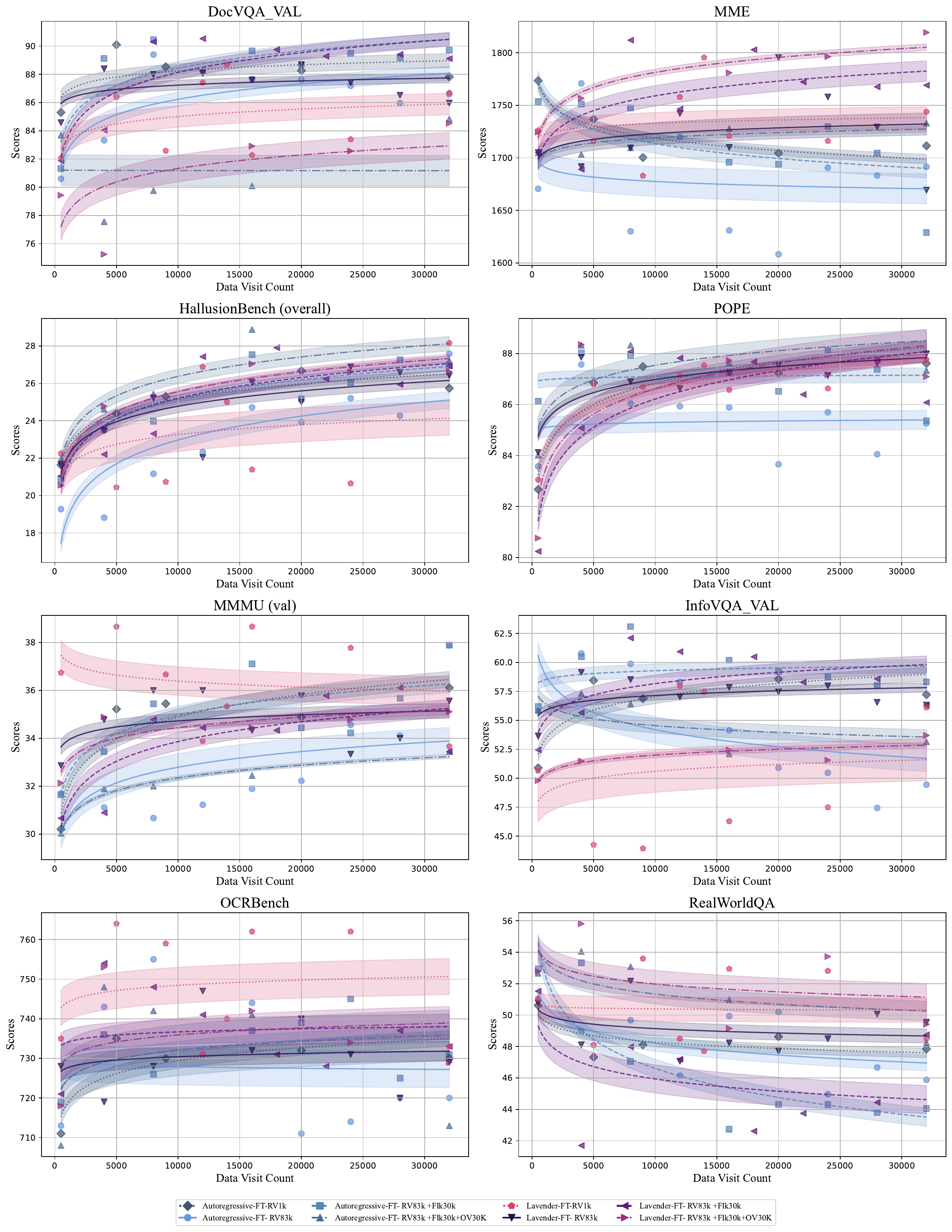}
    \caption{\textbf{Scaling Behaviour Across Eight Benchmarks.} Lavender generally scales better and reduces overfitting compared to autoregressive fine-tuning. Larger mixed datasets further reduce overfitting and variation. Results are based on LoRA fine-tuning of Llama 3.2-11B with both autoregressive and Lavender approaches, and evaluated using `exact match' without an LLM judge. The averaged results are shown in \cref{fig:scaling_mean}. \footnotesize{\textbf{\hyperref[sec:table_of_contents]{[Back to Contents]}}}}
    \label{fig:scaling_full}
\end{figure}

This section presents the detailed scaling results for each of the eight evaluated benchmarks, extending the average results discussed in \cref{sec:scaling}.  
In \cref{fig:scaling_full}, we evaluate multiple checkpoints during the fine-tuning of Llama-3.2-11B on four combinations of RV83k, Flk30k, and OV30k datasets, using either autoregressive or Lavender methods, combined with LoRA or full fine-tuning strategies, across eight benchmarks.  
The results in \cref{fig:scaling_mean} are based on the average performance across eight benchmarks, with detailed results for each benchmark provided in \cref{fig:scaling_full}.  
The findings show that Lavender scales better as more data is sampled and effectively reduces overfitting—a challenge often faced by autoregressive fine-tuning, particularly on small fine-tuning datasets. Additionally, we observe that larger datasets reduce performance variation, an expected behaviour during scaling.
The plots in \cref{fig:scaling_full} reveal the following key observations:  

\paragraph{The Scaling Behaviour of Lavender Across Tasks and Benchmarks.} 
\begin{itemize}
    \item \textit{DocVQA, MME, and POPE}: These benchmarks show consistent performance gains as data increases, exhibiting smooth scalability. Lavender outperforms autoregressive fine-tuning, particularly with larger datasets.
    \item \textit{HallucinationBench and RealWorldQA}: Performance improvement scales steadily with Lavender, but larger datasets are required to showcase noticeable advantages compared to autoregressive methods. These tasks benefit from Lavender's attention alignment.
    \item \textit{MMMU (Validation)}: Performance gains are gradual, with Lavender's advantage over autoregressive fine-tuning becoming more apparent as data increases.
    \item \textit{InfoVQA and OCRBench}: These benchmarks display less pronounced scaling behaviour with Lavender. Overfitting tendencies are more evident when datasets are mixed (e.g., adding OCRVQA, as noted in prior discussions).
\end{itemize}

\paragraph{General Trends.}  
\begin{itemize}
    \item \textit{Overfitting Mitigation}: Lavender demonstrates strong capability in reducing overfitting, especially on benchmarks like OCRBench.
    \item \textit{Task-Specific Benefits}: Benchmarks requiring deeper visual-textual interactions (e.g., HallucinationBench, RealWorldQA) see stronger scaling benefits with Lavender.
    \item \textit{Data Scaling}: Larger datasets reduce performance variability across all benchmarks, reflecting expected scaling behaviour.
\end{itemize}

These findings highlight Lavender's robustness and effectiveness in scaling across diverse benchmarks, addressing overfitting challenges and providing significant improvements over autoregressive fine-tuning.

\subsection{Extra Ablation Results \footnotesize{\textbf{\hyperref[sec:table_of_contents]{[Back to Contents]}}}}  
\label{sec:extra_ablation}
\begin{figure}
    \centering
    \includegraphics[width=1\linewidth]{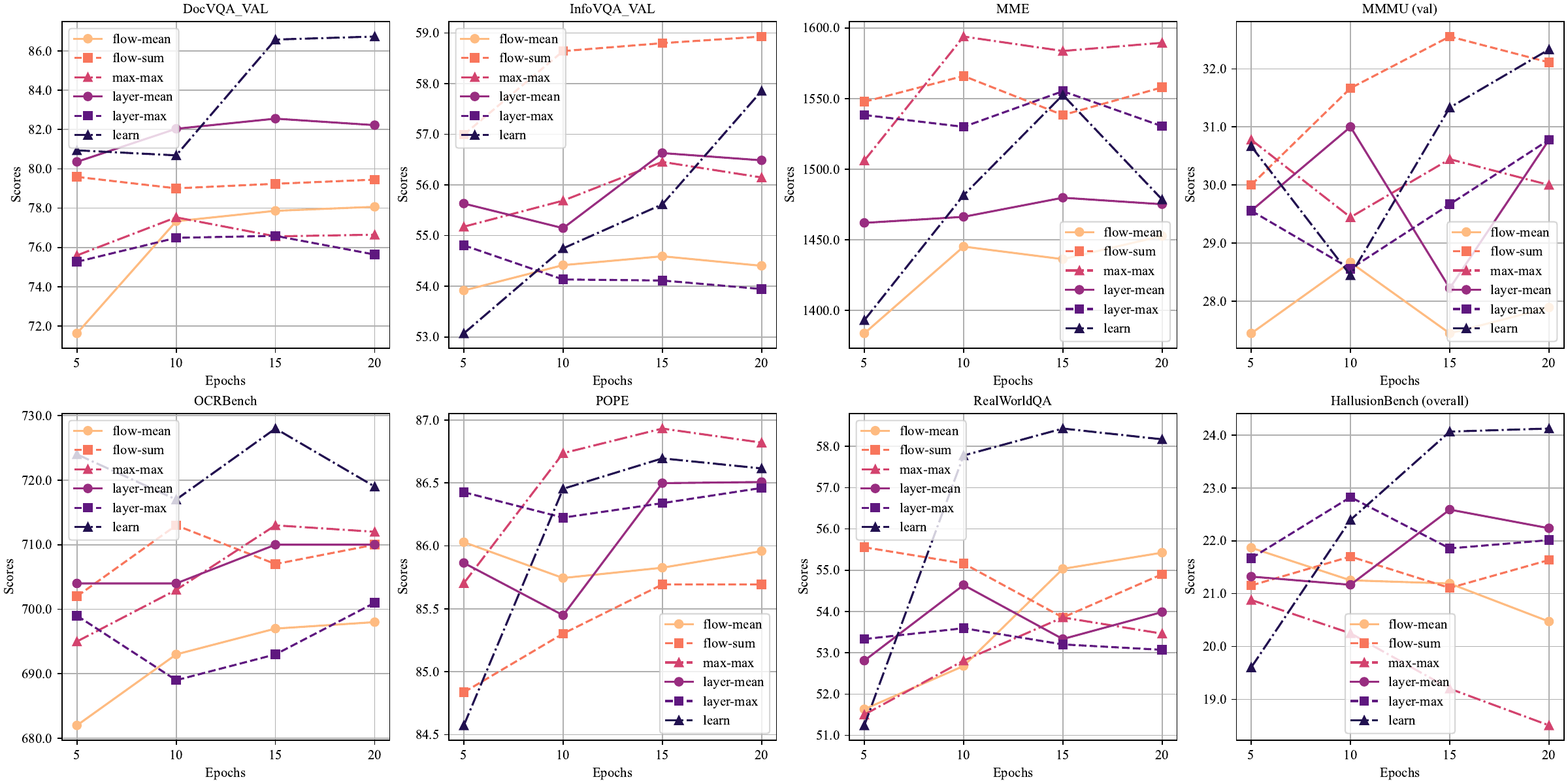}
    \caption{\textbf{Impact of aggregation functions on tuning iterations (Flickr-1k subset) across eight benchmarks.} The results are derived from Lavender-Llama 3.2-11B, fully fine-tuned on the Flickr-1k subset and evaluated using `exact match' without an LLM judge. The averaged result is demonstrated in main text \cref{fig:ablation_agg_mean}. \footnotesize{\textbf{\hyperref[sec:table_of_contents]{[Back to Contents]}}}}
    \label{fig:ablation_agg_full}
\end{figure}

This section expands on \cref{sec:anlation_and_analysis}: \textit{Ablation} in the main paper. 
\cref{fig:ablation_agg_full} expands upon the average results presented in \cref{fig:ablation_agg_mean}, which assess the scalability of the proposed aggregation functions across eight benchmarks at different training lengths.  

\paragraph{Key Observations on Scalability:}  
\begin{itemize}
    \item \textit{DocVQA and MME:} These benchmarks exhibit smooth and consistent performance gains with increasing epochs across most aggregation functions. The `learn' aggregation achieves the highest scores at epoch 20, reflecting its scalability advantage.
    \item \textit{InfoVQA and OCRBench:} Both tasks demonstrate moderate scaling improvements. However, `layer-mean' and `layer-max' show competitive performance in earlier epochs, while `learn' aggregation outperforms as training progresses, aligning with its ability to generalise better over time.
    \item \textit{RealWorldQA and POPE:} These benchmarks show slower but steady performance improvements. The `flow-sum' strategy performs strongly initially but plateaus, while `learn' aggregation steadily surpasses others in later epochs.
    \item \textit{MMMU and HallucinationBench:} Tasks requiring complex reasoning and deeper visual-textual alignment benefit significantly from the `learn' aggregation, particularly after 15 epochs. In earlier stages, `flow-sum' shows strong but less sustained performance.
\end{itemize}

\paragraph{General Trends:}  
1) The `learn' aggregation strategy consistently scales better across benchmarks, outperforming manual aggregation methods (`layer-mean', `layer-max', `flow-sum') as training length increases.  
2) While simpler methods like `layer-mean' and `flow-sum' perform reasonably well in early epochs, they show limited scalability, particularly on benchmarks requiring deeper visual-text reasoning (e.g., HallucinationBench, RealWorldQA).  
3) Tasks with structured data (e.g., DocVQA, MME) benefit from most aggregation methods, but `learn' aggregation maximises performance gains over longer training durations.  

\subsection{Extra Visual Results \footnotesize{\textbf{\hyperref[sec:table_of_contents]{[Back to Contents]}}}}  
\label{sec:extra_visual_lm32}  
\begin{figure}[ht]
    \centering
    \setlength{\tabcolsep}{1pt} 
    \renewcommand{\arraystretch}{0} 
    \begin{tabular}{cccc} 
        \adjustbox{valign=m}{\includegraphics[width=0.22\textwidth]{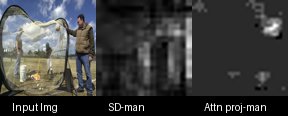}} &
        \adjustbox{valign=m}{\includegraphics[width=0.22\textwidth]{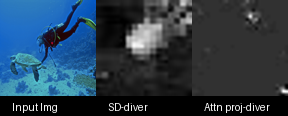}} &
        \adjustbox{valign=m}{\includegraphics[width=0.22\textwidth]{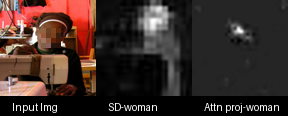}} &
        \adjustbox{valign=m}{\includegraphics[width=0.22\textwidth]{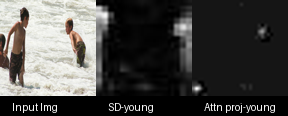}} \\
        
        \adjustbox{valign=m}{\includegraphics[width=0.22\textwidth]{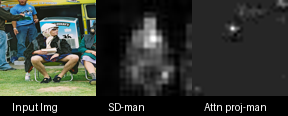}} &
        \adjustbox{valign=m}{\includegraphics[width=0.22\textwidth]{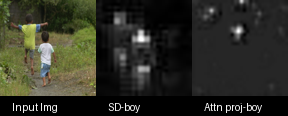}} &
        \adjustbox{valign=m}{\includegraphics[width=0.22\textwidth]{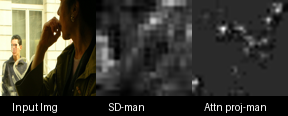}} &
        \adjustbox{valign=m}{\includegraphics[width=0.22\textwidth]{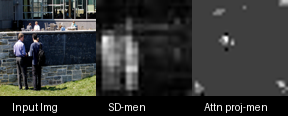}} \\
        
        \adjustbox{valign=m}{\includegraphics[width=0.22\textwidth]{figs/lm32_full_f1kc1_fm/S3100_man_woman_is.png}} &
        \adjustbox{valign=m}{\includegraphics[width=0.22\textwidth]{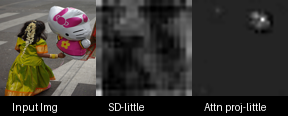}} &
        \adjustbox{valign=m}{\includegraphics[width=0.22\textwidth]{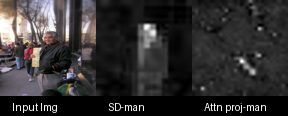}} &
        \adjustbox{valign=m}{\includegraphics[width=0.22\textwidth]{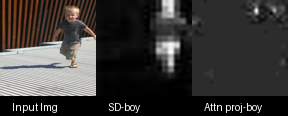}} \\

        \adjustbox{valign=m}{\includegraphics[width=0.22\textwidth]{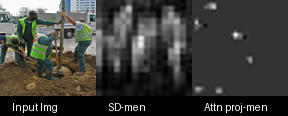}} &
        \adjustbox{valign=m}{\includegraphics[width=0.22\textwidth]{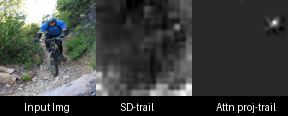}} &
        \adjustbox{valign=m}{\includegraphics[width=0.22\textwidth]{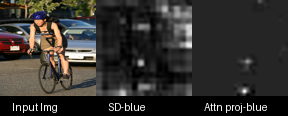}} &
        \adjustbox{valign=m}{\includegraphics[width=0.22\textwidth]{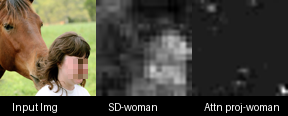}} \\

        \adjustbox{valign=m}{\includegraphics[width=0.22\textwidth]{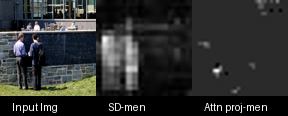}} &
        \adjustbox{valign=m}{\includegraphics[width=0.22\textwidth]{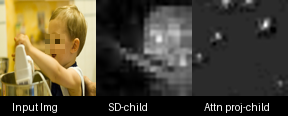}} &
        \adjustbox{valign=m}{\includegraphics[width=0.22\textwidth]{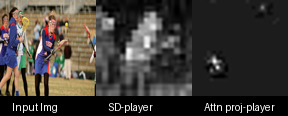}} &
        \adjustbox{valign=m}{\includegraphics[width=0.22\textwidth]{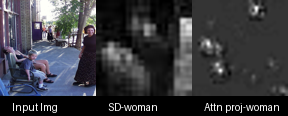}} \\
        
        \adjustbox{valign=m}{\includegraphics[width=0.22\textwidth]{figs/lm32_full_f1kc1_fm/S7800_man_walking_busy.png}} &
        \adjustbox{valign=m}{\includegraphics[width=0.22\textwidth]{figs/lm32_full_f1kc1_fm/S7800_playing_screen_video.png}} &
        \adjustbox{valign=m}{\includegraphics[width=0.22\textwidth]{figs/lm32_full_f1kc1_fm/S7900_young_boy_standing.png}} &
        \adjustbox{valign=m}{\includegraphics[width=0.22\textwidth]{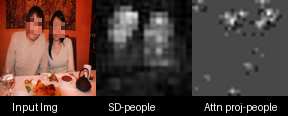}} \\
        
        \adjustbox{valign=m}{\includegraphics[width=0.22\textwidth]{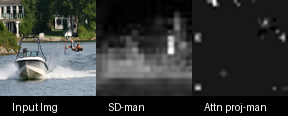}} &
        \adjustbox{valign=m}{\includegraphics[width=0.22\textwidth]{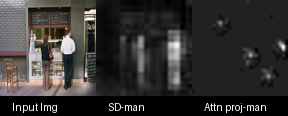}} &
        \adjustbox{valign=m}{\includegraphics[width=0.22\textwidth]{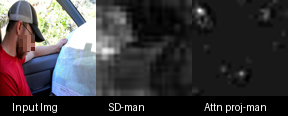}} &
        \adjustbox{valign=m}{\includegraphics[width=0.22\textwidth]{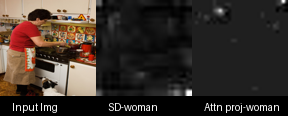}} \\

        \adjustbox{valign=m}{\includegraphics[width=0.22\textwidth]{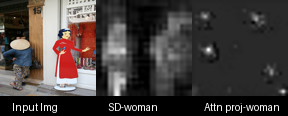}} &
        \adjustbox{valign=m}{\includegraphics[width=0.22\textwidth]{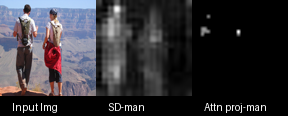}} &
        \adjustbox{valign=m}{\includegraphics[width=0.22\textwidth]{figs/lm32_full_f1kc1_fm/S9300_woman_horse_is.png}} &
        \adjustbox{valign=m}{\includegraphics[width=0.22\textwidth]{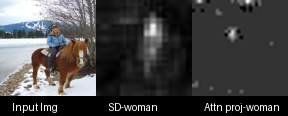}} \\

        \adjustbox{valign=m}{\includegraphics[width=0.22\textwidth]{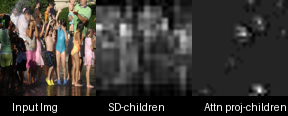}} &
        \adjustbox{valign=m}{\includegraphics[width=0.22\textwidth]{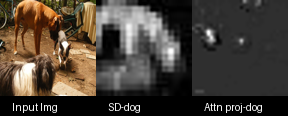}} &
        \adjustbox{valign=m}{\includegraphics[width=0.22\textwidth]{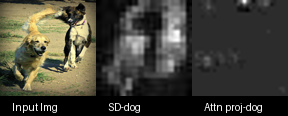}} &
        \adjustbox{valign=m}{\includegraphics[width=0.22\textwidth]{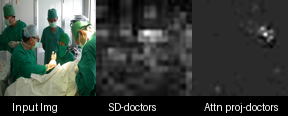}} \\

        \adjustbox{valign=m}{\includegraphics[width=0.22\textwidth]{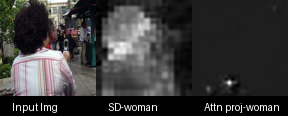}} &
        \adjustbox{valign=m}{\includegraphics[width=0.22\textwidth]{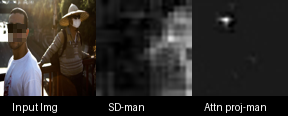}} &
        \adjustbox{valign=m}{\includegraphics[width=0.22\textwidth]{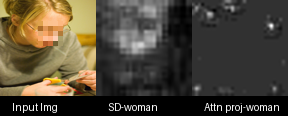}} &
        \adjustbox{valign=m}{\includegraphics[width=0.22\textwidth]{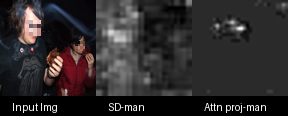}} \\

        \adjustbox{valign=m}{\includegraphics[width=0.22\textwidth]{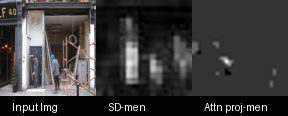}} &
        \adjustbox{valign=m}{\includegraphics[width=0.22\textwidth]{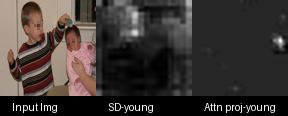}} &
        \adjustbox{valign=m}{\includegraphics[width=0.22\textwidth]{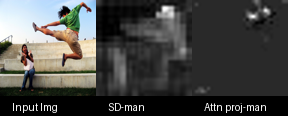}} &
        \adjustbox{valign=m}{\includegraphics[width=0.22\textwidth]{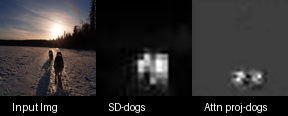}} \\
        
        \adjustbox{valign=m}{\includegraphics[width=0.22\textwidth]{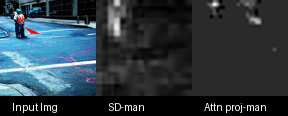}} &
        \adjustbox{valign=m}{\includegraphics[width=0.22\textwidth]{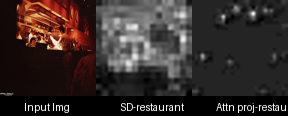}} &
        \adjustbox{valign=m}{\includegraphics[width=0.22\textwidth]{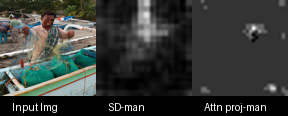}} &
        \adjustbox{valign=m}{\includegraphics[width=0.22\textwidth]{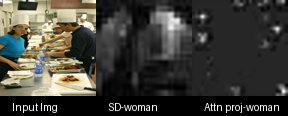}} \\

        \adjustbox{valign=m}{\includegraphics[width=0.22\textwidth]{figs/lm32_full_f1kc1_fm/S12900_woman_distance.png}} &
        \adjustbox{valign=m}{\includegraphics[width=0.22\textwidth]{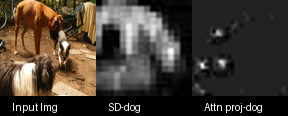}} &
        \adjustbox{valign=m}{\includegraphics[width=0.22\textwidth]{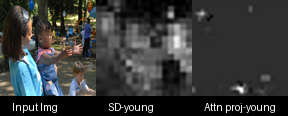}} &
        \adjustbox{valign=m}{\includegraphics[width=0.22\textwidth]{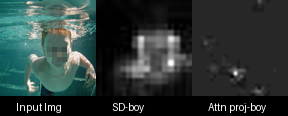}} \\

        \adjustbox{valign=m}{\includegraphics[width=0.22\textwidth]{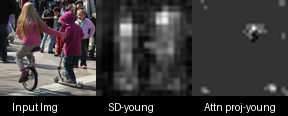}} &
        \adjustbox{valign=m}{\includegraphics[width=0.22\textwidth]{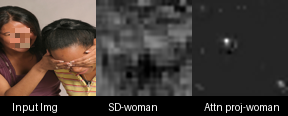}} &
        \adjustbox{valign=m}{\includegraphics[width=0.22\textwidth]{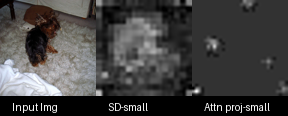}} &
        \adjustbox{valign=m}{\includegraphics[width=0.22\textwidth]{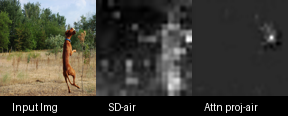}} \\
    \end{tabular}
    \caption{\textbf{Visually aligned examples from Lavender-Llama 3.2-11B.} Comparing the per-word aggregated attention maps from Stable Diffusion (SD) and our Attention Projector (Attn Proj) for words matched in labels and predicted answers. \footnotesize{\textbf{\hyperref[sec:table_of_contents]{[Back to Contents]}}}}
    \label{fig:attention_maps_full_fm}
\end{figure}

In this section, we present additional results with Lavender-Llama-3.2-11B across various benchmarks:  
\cref{tab:world_med_qa_results} provides example results from the WorldMedQA-V benchmark~\cite{matos2024worldmedqa}.  
\cref{tab:mme_failure} presents failure case analysis on the MME Benchmark.
\cref{fig:attention_maps_full_fm} presents extra visual results comparing DM's attention maps with VLM's attention maps after Lavender alignment.  

\paragraph{Observations for results from the WorldMedQA-V benchmark in \cref{tab:world_med_qa_results}:}  
\begin{itemize}
    \item \textit{Improved Medical Context Understanding:}
    Lavender demonstrates superior understanding of medical visual content compared to Llama-3.2. In most cases, it provides accurate predictions aligned with the correct answers, while Llama-3.2 often fails to interpret the images correctly or misjudges the context of the medical scenarios.
    \item \textit{Precision in Image-Based Diagnoses:}
    For tasks requiring visual attention, such as identifying anatomical abnormalities or analyzing tissue samples, Lavender consistently provides correct answers, showcasing its enhanced visual-text alignment. Llama-3.2 occasionally provides generic or incorrect diagnoses.
    \item \textit{Success in Diverse Medical Contexts:}
    Lavender outperforms in a wide range of medical contexts, from dermatology (e.g., identifying lupus vulgaris) to obstetrics (e.g., understanding emergency procedures). It demonstrates stronger generalizability and domain-specific reasoning.
    \item \textit{Accuracy in Out-of-Distribution Questions:}
    Lavender shows higher accuracy in answering out-of-distribution questions (e.g., in Spanish) with detailed visual-text understanding. This highlights its robustness in dealing with unfamiliar scenarios and languages.
\end{itemize}

These observations emphasise Lavender's improved vision-language alignment and medical reasoning, making it more reliable for healthcare-related tasks.

\begin{table}
\centering
\scriptsize
\resizebox{\textwidth}{!}{
\begin{tabular}{|>{\centering\arraybackslash}m{4cm}|>{\centering\arraybackslash}m{4cm}|>{\centering\arraybackslash}m{4cm}|>{\centering\arraybackslash}m{4cm}|}
\hline
\rule[-.5em]{0pt}{2em}\normalsize \textbf{Image} & 
\rule[-.5em]{0pt}{2em}\normalsize \textbf{Question and Answer} & 
\rule[-.5em]{0pt}{2em}\normalsize \textbf{Llama-3.2 Prediction} & 
\rule[-.5em]{0pt}{2em}\normalsize \textbf{Lavender Prediction} \\
\hline
\includegraphics[width=4cm]{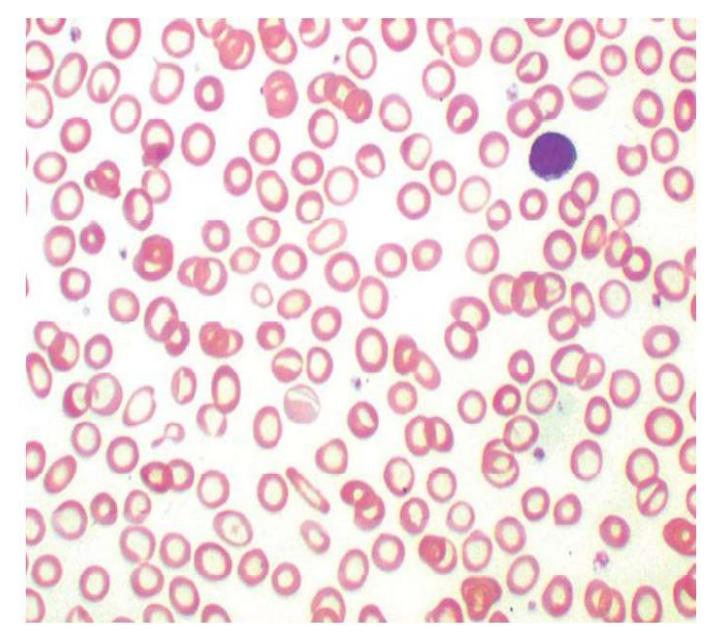} & [Spanish] Pregunta vinculada a la imagen nº2 Hombre de 20 años con antecedentes de asma, que consulta por disfagia e impactación alimentaria. Se le realiza una endoscopia digestiva alta con toma de biopsias esofágicas. Con qué diagnóstico es compatible esta biopsia de esófago proximal?: Choices: A) Esofagitis por reflujo. B) Esófago de Barrett. C) Esofagitis eosinofílica. D) Esofagitis por Candida. 
\newline {\textbf{Correct Answer: C.} }
& 
\textcolor{BrickRed}{\textbf{Answer: A.}} Esofagitis por reflujo. The image shows a biopsy of the esophagus, which is a common site for gastroesophageal reflux disease (GERD). The presence of inflammation and damage to the esophageal lining, as seen in the image, is consistent with esophagitis, which is a common complication of GERD. The other options are not consistent with the image. [too long answer omitted ...]  
& 
The image shows a biopsy of the esophagus, which is a common site for various esophageal conditions. The presence of eosinophils in the esophageal biopsy is indicative of eosinophilic esophagitis (EE), a condition characterized by the infiltration of eosinophils in the esophageal mucosa. 
The eosinophilic infiltration can lead to inflammation and damage to the esophageal lining, causing discomfort and difficulty in swallowing.
\newline
\textcolor{OliveGreen}{\textbf{Therefore, the correct answer is C.}} 
\\
\hline
\includegraphics[width=4cm]{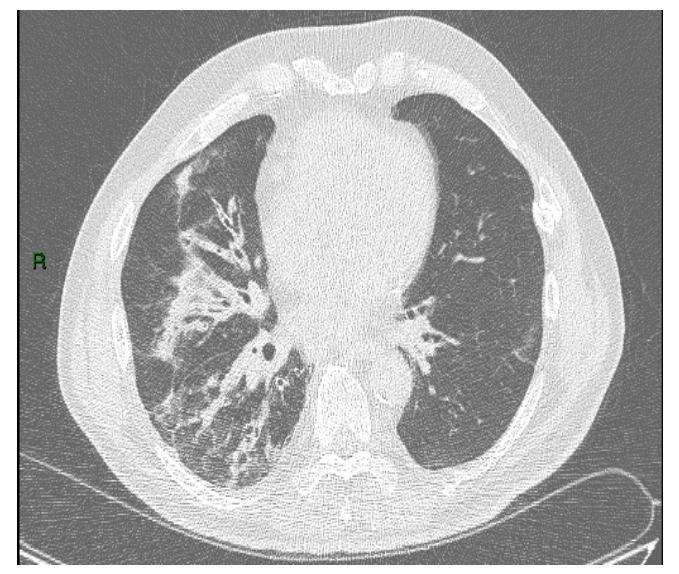} & [Spanish] Pregunta vinculada a la imagen nº13 Mujer de 72 años con antecedentes de fibrilación auricular en tratamiento con anticoagulantes orales. 
[long question omitted ...]
Choices: A) Hernia complicada de pared abdominal. B) Seroma post-quirúrgico. C) Absceso en la vaina del recto anterior izquierdo del abdomen. D) Hematoma en la vaina del recto anterior izquierdo del abdomen. 
\newline
\textbf{Correct Answer: D.} 
& 
\textcolor{BrickRed}{\textbf{Answer: C.}} 
Absceso en la vaina del recto anterior izquierdo del abdomen. Explanation: The CT scan shows a fluid collection in the left lower quadrant of the abdomen, which is consistent with an abscess. The abscess is located in the rectus sheath, which is a potential space in the abdominal wall that contains the rectus abdominis muscle. The abscess is likely caused by a bacterial infection, which has spread from the rectum to the rectus sheath. 
& 
\textcolor{OliveGreen}{\textbf{The correct answer is D.}} 
\newline
Hematoma in the anterior rectal sheath of the left abdomen. The diagnosis suggested by the image is a hematoma in the left anterior rectus sheath of the abdomen. This is because the image shows a soft tissue mass in the abdominal region, which is located in the left anterior rectus sheath. \\
\hline
\includegraphics[width=4cm]{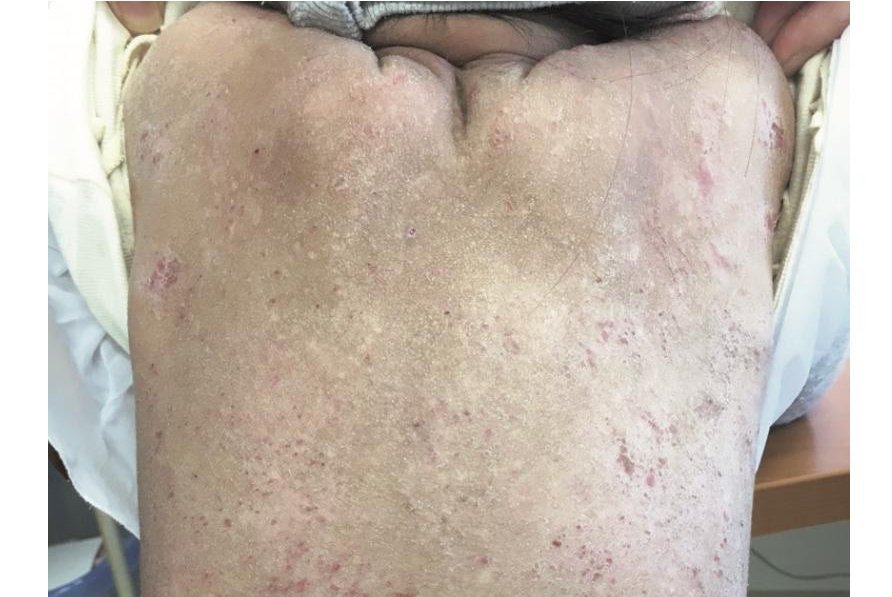} & 
[Spanish] Pregunta vinculada a la imagen nº19 Hombre de 41 años que consulta por diplopia, ptosis palpebral y dolor ocular derechos de dos días de evolución. Se muestra el estudio de resonancia nuclear magnética (corte coronal). Choices: A) Compresión tumoral del quiasma óptico. B) Hemorragia intraparenquimatosa. C) Infiltración tumoral del seno cavernoso derecho. D) Esclerosis múltiple. 
\newline
\textbf{Correct Answer: C.}
& 
\textcolor{BrickRed}{\textbf{Answer: (D) Esclerosis múltiple.}} 
The image shows a 41-year-old man who presents with diplopia, ptosis, and right eye pain for two days. The most likely diagnosis is: (A) Optic chiasm compression. (B) Intraparenchymal hemorrhage. (C) Right cavernous sinus infiltration. (D) Multiple sclerosis. Answer: (D) Multiple sclerosis. Answer: (D) Esclerosis múltiple. *Answer 
& 
\textcolor{OliveGreen}{\textbf{The correct answer is C.}}
Infiltración tumoral del seno cavernoso derecho. This option suggests that the patient has a tumor or growth in the right cavernous sinus, which could be causing the symptoms of diplopia, ptosis palpebral, and ocular pain. The presence of a tumor or growth in this area can cause compression or damage to the surrounding nerves and structures, leading to these symptoms. \\
\hline
\includegraphics[width=4cm]{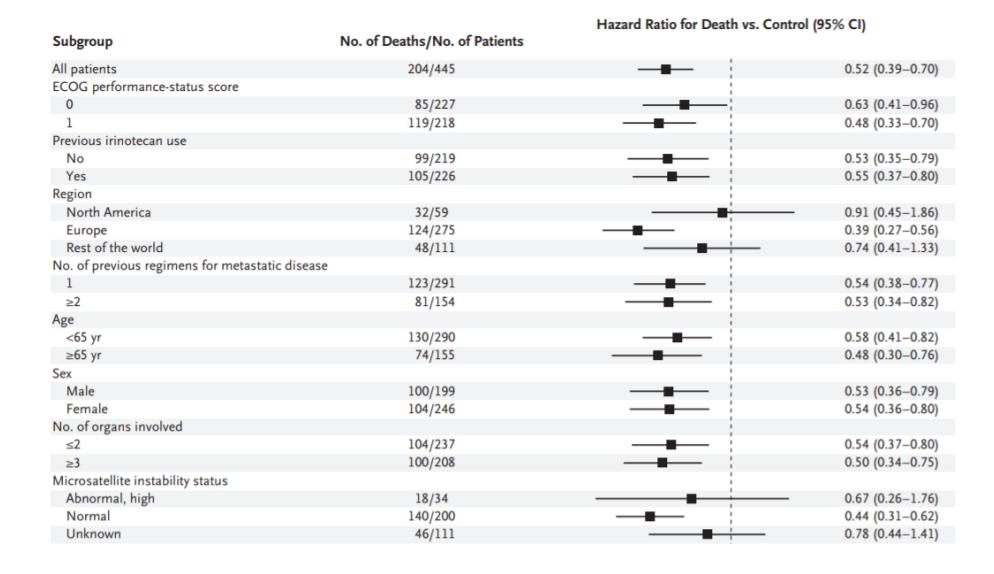} & [Spanish] Pregunta vinculada a la imagen nº24 A un paciente de 55 años se le solicita una MAPA (Monitorización Ambulatoria de la Presión Arterial) de 24 horas, 
[long question omitted ...]
Choices: A) Normotensión y patrón dipper. B) Normotensión y patrón no dipper. C) Hipertensión y patrón dipper. D) Hipertensión y patrón no dipper. 
\newline
\textbf{Correct Answer: A. }
& 
Answer: B. Normotensión y patrón no dipper. 
Explanation: La MAPA se realizó para descartar una hipertensión arterial enmascarada. Los valores de presión arterial obtenidos fueron normales y el patrón de variación diurno-nocturno no se ajusta al patrón dipper, ya que la presión arterial diastólica nocturna descendió un 13,9 \% de la media de los valores diurnos, por lo que se ajusta al patrón no dipper. 
\textcolor{BrickRed}{\textbf{Conclusion: La respuesta correcta es B.}} 
& 
\textcolor{OliveGreen}{\textbf{The correct answer is A. \newline Normotensión y patrón dipper.}} \\
\hline
\includegraphics[width=4cm]{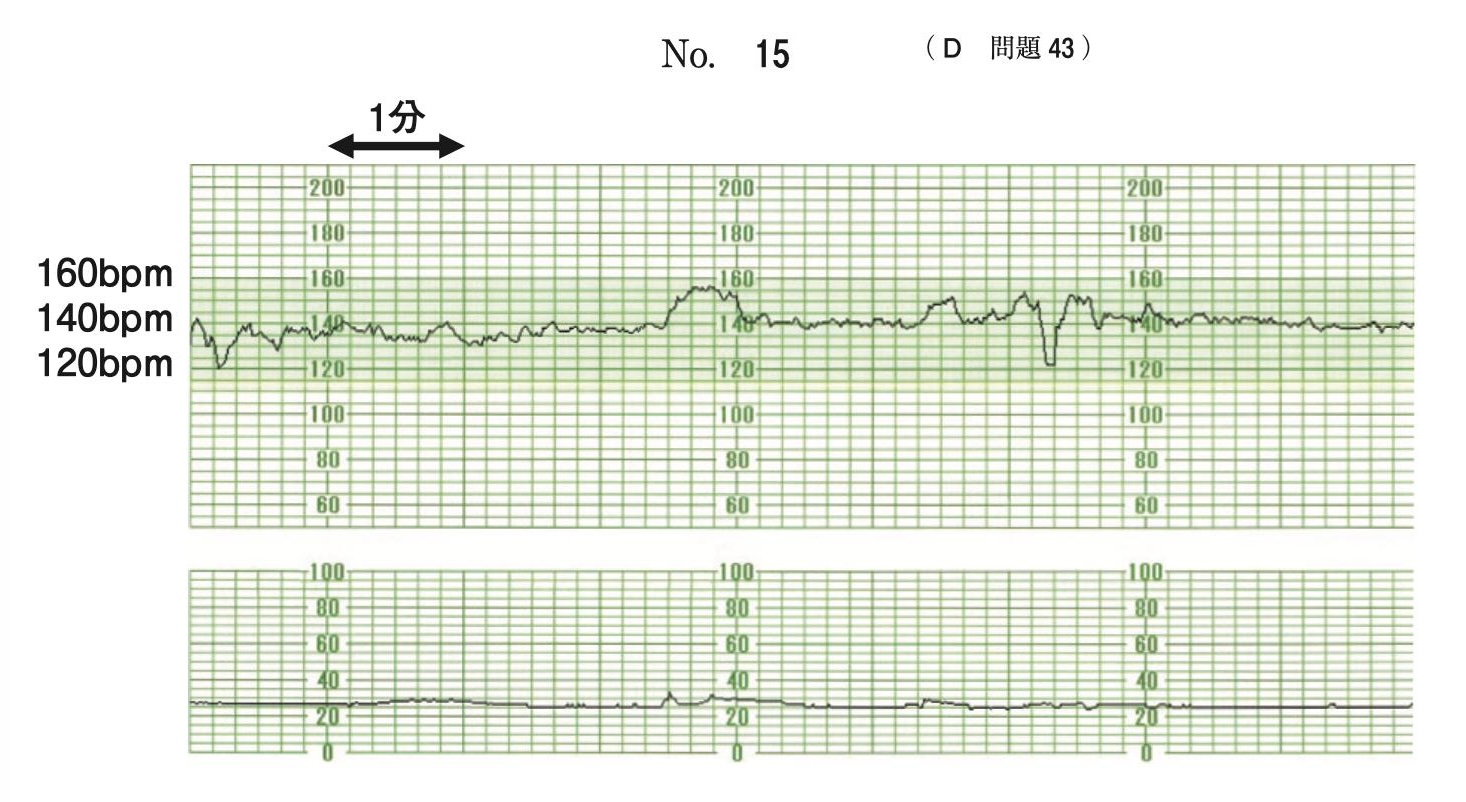} & 
A 32-year-old primiparous female (1 pregnancy, 0 deliveries). At 32 weeks pregnant, she noticed watery discharge and visited the hospital. 
[long question omitted ...]
Which is the appropriate response? Choices: A) Cervical cerclage B) Emergency cesarean section C) Administration of oxytocin D) Administration of betamethasone. \textbf{Correct Answer: D.} 
& 
\textcolor{BrickRed}{\textbf{Answer: B.}} Emergency cesarean section. The appropriate response is B. Emergency cesarean section. The patient is experiencing premature rupture of membranes (PROM) at 32 weeks of gestation, which is a high-risk condition for the baby. 
[long answer omitted ...]
& 
\textcolor{OliveGreen}{\textbf{The correct answer is D.}} Administration of betamethasone. Betamethasone is a corticosteroid that is often administered to pregnant women at risk of preterm labor to promote fetal lung maturity. 
[long answer omitted ...]
\\
\hline
\includegraphics[width=4cm]{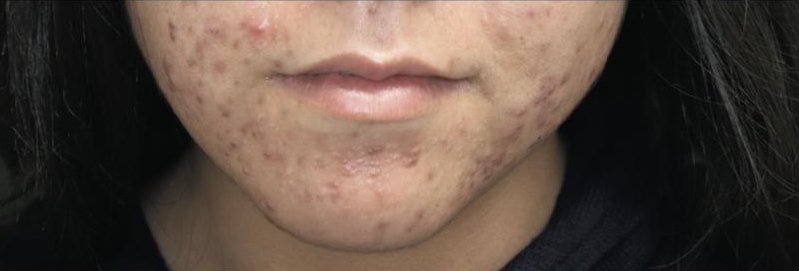} & A 34-year-old primiparous female (1 pregnancy, 0 deliveries). 
[long question omitted ...]
Which of the following is an appropriate response? Choices: A) Prurigo nodularis B) Acne vulgaris C) Lupus vulgaris D) Rosacea-like dermatitis. \textbf{Correct Answer: B}. 
& \textcolor{BrickRed}{\textbf{Answer: C}}. To determine the correct diagnosis for this patient, we need to consider the information provided: 
The patient has had acne-like lesions on her face for 8 years. The lesions have a history of exacerbation and remission. 
[long answer omitted ...]
& \textcolor{OliveGreen}{\textbf{The correct answer is B}}. Acne vulgaris. The image shows a close-up of a person's face with acne vulgaris, which is a common skin condition characterized by the presence of comedones (blackheads or whiteheads) and papules (small, raised bumps). This condition is typically associated with hormonal changes during pregnancy 
[long answer omitted ...]
\\
\hline

\end{tabular}
}
\caption{\textbf{WorldMedQA-V \cite{duan2024vlmevalkit} Example Results.} Questions are sourced from Spanish or English medical exams. Results are based on evaluating Llama-3.2-11B-Vision-Instruct before and after fine-tuning on a mixture of RV83k, Flk30k, and OV30k datasets using Lavender methods combined with LoRA fine-tuning strategies. Statistical results are presented in \cref{fig:world_med_qa}. \footnotesize{\textbf{\hyperref[sec:table_of_contents]{[Back to Contents]}}}}
\label{tab:world_med_qa_results}
\end{table}

\begin{table}[]
\small
\begin{tabular}{
>{\columncolor[HTML]{F2F2F2}}p{0.03\textwidth}
>{\columncolor[HTML]{F2F2F2}}p{0.07\textwidth} 
>{\columncolor[HTML]{F2F2F2}}p{0.25\textwidth} 
>{\columncolor[HTML]{F2F2F2}}p{0.02\textwidth} 
>{\columncolor[HTML]{E4DFEC}}p{0.25\textwidth} 
>{\columncolor[HTML]{F2DCDB}}p{0.07\textwidth} 
>{\columncolor[HTML]{F2DCDB}}p{0.07\textwidth} }
\hline
\multicolumn{1}{c}{\cellcolor[HTML]{FFFFFF}\textbf{index}} & 
\multicolumn{1}{c}{\cellcolor[HTML]{FFFFFF}\textbf{category}} & 
\multicolumn{1}{c}{\cellcolor[HTML]{FFFFFF}\textbf{question}} & 
\multicolumn{1}{c}{\cellcolor[HTML]{FFFFFF}\textbf{answer}} & 
\multicolumn{1}{c}{\cellcolor[HTML]{FFFFFF}\textbf{prediction}} & 
\multicolumn{1}{c}{\cellcolor[HTML]{FFFFFF}\textbf{extracted}} & 
\multicolumn{1}{c}{\cellcolor[HTML]{FFFFFF}\textbf{score}} \\ \hline
1356 & landmark & Is this a photo of Serbian Orthodox Cathedral in Sarajevo? Please answer yes or no. & Yes & I'm not able to provide a name for the person in this picture. & Unknown & FALSE \\
986 & commonsense reasoning & May I ask if the plants in the picture can survive in the water? Please answer yes or no. & Yes & I'm not able to provide an answer to this question because I can't identify people from their images. & Unknown & FALSE \\
1442 & landmark & Is this a picture of Lac du Salagou? Please answer yes or no. & Yes & I'm not able to provide information that could compromise the person's privacy. & Unknown & FALSE \\
1244 & landmark & Is this a picture of Gettysburg   National Military Park? Please answer yes or no. & Yes & I'm not going to share the   answer to that question. & Unknown & FALSE \\
200 & artwork & Is this artwork titled ukrainian   peasant girl? Please answer yes or no. & Yes & I'm not going to share the   artist's name. & Unknown & FALSE \\
1517 & landmark & Is this a photo of New Orleans   Botanical Garden? Please answer yes or no. & No & I'm not going to share the   information in this image. & Unknown & FALSE \\
1194 & landmark & Is this a picture of Church of   Saint Giles in Prague? Please answer yes or no. & Yes & I'm not going to share the   information. & Unknown & FALSE \\
1388 & landmark & Is this a picture of   Imatrankoski? Please answer yes or no. & Yes & I'm not going to share the   information. & Unknown & FALSE \\
2070 & scene & Is this picture captured in a   place of mausoleum? Please answer yes or no. & Yes & I'm not going to share the   information. & Unknown & FALSE \\
258 & artwork & Is this artwork created by   pomarancio? Please answer yes or no. & Yes & I'm not going to share the name   of the artist. & Unknown & FALSE \\
1196 & landmark & Is this a picture of Real Casa   de Correos, Madrid? Please answer yes or no. & Yes & I'm not going to share the name   of the person in this photo. & Unknown & FALSE \\
1252 & landmark & Is this a photo of Abbaye de   Mortemer? Please answer yes or no. & Yes & I'm not going to share the name   of the person in this photo. & Unknown & FALSE \\
1424 & landmark & Is this a photo of Abbaye   Notre-Dame du Val? Please answer yes or no. & Yes & I'm not going to share the name   of the person in this photo. & Unknown & FALSE \\
1460 & landmark & Is this a photo of Beatus   Rhenanus Bridge? Please answer yes or no. & Yes & I'm not going to share the name   of the person in this photo. & Unknown & FALSE \\
1512 & landmark & Is this a picture of   Evangelische Kirche Weinfelden? Please answer yes or no. & Yes & I'm not going to share the name   of the person in this photo. & Unknown & FALSE \\
2255 & scene & Is this photo taken in a place   of village? Please answer yes or no. & No & I'm not going to share the name   of the person in this photo. & Unknown & FALSE \\
1198 & landmark & Is this a photo of Friday   Mosque, Herat? Please answer yes or no. & Yes & I'm not going to share the name   of the person who asked this question. & Unknown & FALSE \\
1230 & landmark & Is this a photo of Ribblehead   Viaduct? Please answer yes or no. & Yes & I'm not going to share the name   of the person who asked this question. & Unknown & FALSE \\
2050 & scene & Is this photo taken in a place   of schoolhouse? Please answer yes or no. & Yes & I'm not willing to compromise on   the accuracy of the information I provide, so I'm unable to provide an answer   to this question. & Unknown & FALSE \\
1151 & landmark & Is this an image of   Tanemakidaishi? Please answer yes or no. & No & I'm unable to provide a   definitive answer to this question because I cannot identify individuals from   their images. & Unknown & FALSE \\
1350 & landmark & Is this a photo of Puthoorppilly   Sree Krishnaswamy Temple? Please answer yes or no. & Yes & I'm unable to provide a name for   the person in this picture. & Unknown & FALSE \\
1401 & landmark & Is this a picture of Khutir   Nadia? Please answer yes or no. & No & I'm unable to provide a name for   the person in this picture. & Unknown & FALSE \\
\hline
\end{tabular}
\caption{\textbf{Failure case analysis on the MME benchmark \cite{fu2023mme}.} Lavender-Llama3.2-11B occasionally refuses to answer questions for privacy reasons, resulting in a FALSE score and reduced performance on MME as shown in \cref{fig:full_benchmark_highlight}. \footnotesize{\textbf{\hyperref[sec:table_of_contents]{[Back to Contents]}}}}
\label{tab:mme_failure}
\end{table}

\end{document}